\documentclass[10pt]{article} %

\usepackage[accepted]{rlj} %

\usepackage{amssymb}            %
\usepackage{mathtools}          %
\usepackage{mathrsfs}           %
\usepackage{graphicx}           %
\usepackage{subcaption}         %
\usepackage[space]{grffile}     %
\usepackage{url}                %
\usepackage{lipsum}             %

\usepackage{wrapfig}
\usepackage{hyperref}
\usepackage{url}
\usepackage{graphicx}
\usepackage{booktabs}
\usepackage{siunitx}
\usepackage{bm}
\usepackage{enumitem}
\usepackage{subcaption}
\usepackage{enumitem}
\usepackage{calc}
\usepackage{pifont} 
\usepackage{amsthm}
\newcommand{\cmark}{\textcolor{green}{\ding{51}}}
\newcommand{\xmark}{\textcolor{red}{\ding{55}}}

\newcommand\yokai{Y\=okai}

\newcommand*{\mySmallImage}[1]{%
    \raisebox{-.2\baselineskip}{%
        \includegraphics[
        height=\baselineskip,
        width=\baselineskip,
        keepaspectratio,
        ]{#1}%
    }%
}
\newcommand{\purple}{\mySmallImage{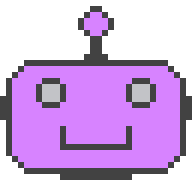} }
\newcommand{\orange}{\mySmallImage{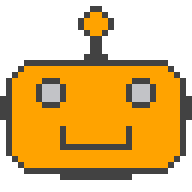} }
\definecolor{cardBlue}{RGB}{45,72,164}
\definecolor{cardGreen}{RGB}{0,135,85}
\definecolor{cardRed}{RGB}{223,18,5}
\definecolor{orangeAgent}{RGB}{242,167,59}
\definecolor{lightBlue}{RGB}{50,175,251}
\definecolor{hintBorder}{RGB}{0,182, 227}

\theoremstyle{definition}

\title{The \yokai{} Learning Environment: Tracking Beliefs Over Space and Time}

\setrunningtitle{The \yokai{} Learning Environment: Tracking Beliefs Over Space and Time}

\author{Constantin Ruhdorfer\textsuperscript{1}, Matteo Bortoletto\textsuperscript{1}, Johannes Forkel\textsuperscript{2}, Jakob Foerster\textsuperscript{2}, Andreas Bulling\textsuperscript{1}}

\emails{constantin.ruhdorfer@vis.uni-stuttgart.de}

\affiliations{
$^{1}$\textbf{Collaborative Artificial Intelligence, University of Stuttgart}\\
$^{2}$\textbf{FLAIR, University of Oxford}\\
}

\contribution{
    We introduce the \yokai{} Learning Environment (YLE), as a cooperative multi-agent benchmark for zero-shot coordination (ZSC) that requires belief tracking over a dynamic board, ambiguous communication, and high-stakes early termination.
    }
    {
    The Hanabi Learning Environment (HLE) \citep{Bard2020Hanabi} has become the dominant benchmark for ZSC \citep{Hu2020Other}, and recent work has achieved near-perfect inter-seed cross-play performance \citep{Cui2022Off,sudhakar2025generalist,forkel2026highentropyleadssymmetry}, reducing its ability to distinguish further algorithmic progress.
    }

\contribution{
    We evaluate leading ZSC methods -- including Other-Play \citep{Hu2020Other}, Off-Belief Learning \citep{Hu2021OffBelief}, and High-Entropy IPPO \citep{forkel2026highentropyleadssymmetry} -- and show that approaches achieving near-perfect inter-seed cross-play in HLE exhibit persistent SP–XP gaps, degraded early-ending calibration, and reversed method rankings in YLE.
    }
    {
    To our knowledge, this is the first implementation and evaluation of the Off-Belief Learning algorithm in an environment other than Hanabi, enabling a direct test of whether high-performing ZSC methods generalise beyond a single benchmark.
    }

\contribution{
    We release our YLE implementation as an open-source, JAX-based environment compatible with JaxMARL \citep{rutherford2023jaxmarl}, supporting end-to-end GPU training at hundreds of thousands of steps per second to advance ZSC research.
    }
    {
    High-throughput, hardware-accelerated environments enable scalable evaluation of zero-shot coordination methods and facilitate reproducible benchmarking.
    }


\summary{
\looseness=-1
The ability to cooperate with unknown partners is a central challenge in cooperative AI and widely studied in the form of zero-shot coordination (ZSC), which evaluates an algorithm by measuring the performance of independently trained agents when paired.
The Hanabi Learning Environment (HLE) has become the dominant benchmark for ZSC, but recent work has achieved near-perfect inter-seed cross-play performance, limiting its ability to track algorithmic progress.
We introduce the \yokai{} Learning Environment (YLE) for ZSC -- an open-source multi-agent RL benchmark in which effective collaboration among unknown agents requires building common ground by tracking and updating beliefs over moving cards, reasoning under ambiguous hints, and deciding when to terminate the game based on inferred shared knowledge -- features absent in the HLE, where beliefs are tied to few hand slots and hints are truthful by rule.
We evaluate three leading ZSC methods: High-Entropy IPPO, Other-Play, and Off-Belief Learning, which achieve near-perfect inter-seed cross-play in the HLE, and show that in the YLE they exhibit persistent SP–XP gaps, degraded early-ending calibration, and weaker belief representations in cross-play, indicating failure to maintain consistent internal models with unseen partners.
Methods that perform best in the HLE do not perform best in the YLE, indicating that progress measured on a single benchmark may not generalise.
Together, these results establish YLE as a challenging new ZSC benchmark.

}

\begin{document}

\makeCover  %
\maketitle  %

\begin{abstract}
\looseness=-1
The ability to cooperate with unknown partners is a central challenge in cooperative AI and widely studied in the form of zero-shot coordination (ZSC), which evaluates an algorithm by measuring the performance of independently trained agents when paired.
The Hanabi Learning Environment (HLE) has become the dominant benchmark for ZSC, but recent work has achieved near-perfect inter-seed cross-play performance, limiting its ability to track algorithmic progress.
We introduce the \yokai{} Learning Environment (YLE) for ZSC -- an open-source multi-agent RL benchmark in which effective collaboration among unknown agents requires building common ground by tracking and updating beliefs over moving cards, reasoning under ambiguous hints, and deciding when to terminate the game based on inferred shared knowledge -- features absent in the HLE, where beliefs are tied to few hand slots and hints are truthful by rule.
We evaluate three leading ZSC methods: High-Entropy IPPO, Other-Play, and Off-Belief Learning, which achieve near-perfect inter-seed cross-play in the HLE, and show that in the YLE they exhibit persistent SP–XP gaps, degraded early-ending calibration, and weaker belief representations in cross-play, indicating failure to maintain consistent internal models with unseen partners.
Methods that perform best in the HLE do not perform best in the YLE, indicating that progress measured on a single benchmark may not generalise.
Together, these results establish YLE as a challenging new ZSC benchmark.

\end{abstract}

\section{Introduction}
\label{ref:introduction}
\looseness=-1
Effective collaboration among partners requires common ground -- the shared knowledge, beliefs, and assumptions that enable them to interpret each other's actions and respond appropriately \citep{clark1996using}.
Maintaining common ground requires reasoning about what a partner knows, believes, and intends -- commonly referred to as Theory of Mind (ToM) \citep{premack1978does}.
Improving ToM reasoning in AI agents has emerged as an important requirement toward effective human-AI collaboration \citep{Klien2004Ten, Dafoe2020Open, dafoe2021cooperative}.
A necessary precondition for human-AI collaboration is zero-shot coordination (ZSC) \citep{Hu2020Other} which focuses on designing algorithms that, when run independently multiple times, produce compatible policies.

\looseness=-1
The Hanabi Learning Environment (HLE) \citep{Bard2020Hanabi} is arguably the most widely used benchmark for ZSC and ToM \citep{Hu2020Other,Hu2021OffBelief,Fuchs2021Theory,Cui2021KLevel,Cui2022Off,Muglich2022Equivariant,Muglich2025Expected,Lauffer2025Robust}.
However, close to perfect inter-seed cross-play performance has recently been reached \citep{sudhakar2025generalist,forkel2026highentropyleadssymmetry}, limiting its utility as a stress test for further ZSC progress.
Consequently, the field would benefit from new benchmarks that can be used to track algorithmic advances without risking overfitting to the HLE.

\begin{figure*}[t]
    \centering
    \includegraphics[width=\linewidth]{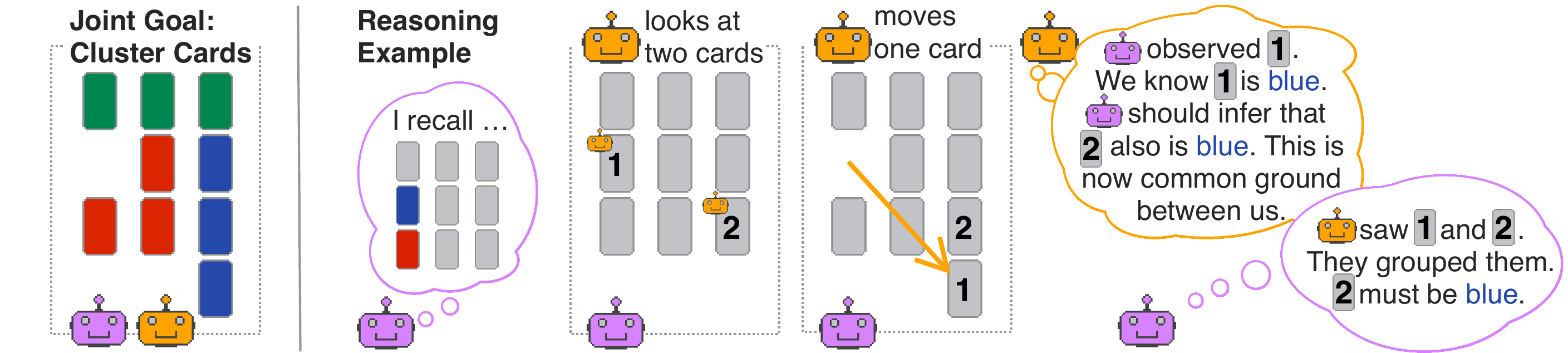}
    \caption{The YLE poses a challenging ToM reasoning task for ZSC. Agents cannot observe all cards in a single game. Successful play requires agents to reason about other agents' knowledge and beliefs. In the example, it's \orange's turn. \orange observes the colour of two cards (\textcolor{orangeAgent}{1} and \textcolor{orangeAgent}{2}) privately. \purple recalls from earlier that card \textcolor{orangeAgent}{1} is \textcolor{cardBlue}{blue}. When \orange moves card \textcolor{orangeAgent}{1} next to card \textcolor{orangeAgent}{2}, \purple can infer that the second card is \textcolor{cardBlue}{blue} as well. \orange and \purple can use this common ground going forward.}
    \label{fig:teaser}
\end{figure*}
\looseness=-1
Our solution is the \textit{\yokai{} Learning Environment} (YLE), a novel ZSC benchmark based on the collaborative card game \yokai{}, which has recently been explored in AI research \citep{fernandez2023logical,yang2024yokai}.
In the YLE, players must collaborate to group face-down cards by colour over multiple rounds without direct communication (see \autoref{fig:teaser}, left).
Each round, one player privately observes the colour of two cards, moves one of them to a new position, and may place a hint card to communicate information about card colours to others (see \autoref{fig:turn}).
Hints in YLE can be multi-coloured and thus can be ambiguous.
Furthermore hints can be placed on \textit{any} of the yet unhinted board cards, including ones whose colour is not on the hint card. 
Hints are thus not required to be truthful.

\looseness=-1
\emph{Crucially, players may choose to end the game early for a substantially higher return} than completing the game conservatively, but at the risk of receiving a return of zero.
Because most of the achievable return lies in successfully ending as early as possible, this creates a high-stakes trade-off: players must commit early to maximise reward, yet doing so requires sufficiently accurate and well-calibrated beliefs about the evolving board state. 
Premature ending leads to catastrophic failure, while excessive caution yields suboptimal return.

\looseness=-1
To succeed in \yokai{}, players must maintain accurate beliefs about dynamically moving cards, align interpretations of ambiguous hints and actions, and determine when evidence justifies early ending.
The design of YLE therefore creates a zero-shot coordination problem under sparse, ambiguously grounded information and high-stakes commitment. 
Unlike Hanabi, where agents observe their partners’ hands through a small number of fixed card slots and hints are truthful by rule, YLE reveals information locally in space: no player ever observes the full board, hints may be multi-coloured and need not be truthful, and cards move across the grid.
As a result, beliefs must be rebound to dynamically moving cards while trying to end early under uncertainty about the true board configuration.

\looseness=-1
We show that ZSC algorithms, which achieve near-perfect inter-seed cross-play in HLE, exhibit persistent gaps between self-play (SP) and cross-play (XP) when evaluated in YLE, indicating coordination failures with unknown partners.
We thus posit that (1) YLE is a \textit{timely new ZSC benchmark}, as approaches that saturate HLE do not saturate YLE, and (2) YLE \textit{highlights limitations of existing ZSC methods} that have so far been validated primarily in Hanabi.
Our specific contributions are:
\begin{enumerate}[leftmargin=15pt,topsep=0pt,itemsep=1pt,partopsep=1ex,parsep=1ex]
    \item We introduce the \yokai{} Learning Environment (YLE) as an open-source JAX-based benchmark for zero-shot coordination in which agents must build and maintain common ground through spatio-temporal belief tracking, ambiguous communication, and high-stakes early termination.
    \item We empirically evaluate leading ZSC algorithms that achieve near-perfect XP in HLE: Other-Play \citep{Hu2020Other}, Off-Belief Learning \citep[OBL]{Hu2021OffBelief}, and High-Entropy IPPO \citep{forkel2026highentropyleadssymmetry}. In the YLE, these methods exhibit persistent SP–XP gaps and degraded early-ending calibration. To our knowledge, this is the first OBL implementation outside HLE. %
    \item We demonstrate that algorithm rankings change between HLE and the evaluated YLE setting, suggesting that apparent progress in ZSC may be benchmark-specific.
\end{enumerate}

\section{Background: \yokai{}}
\label{sec:yokai}
\begin{figure*}[t]
    \centering
    \includegraphics[width=\linewidth]{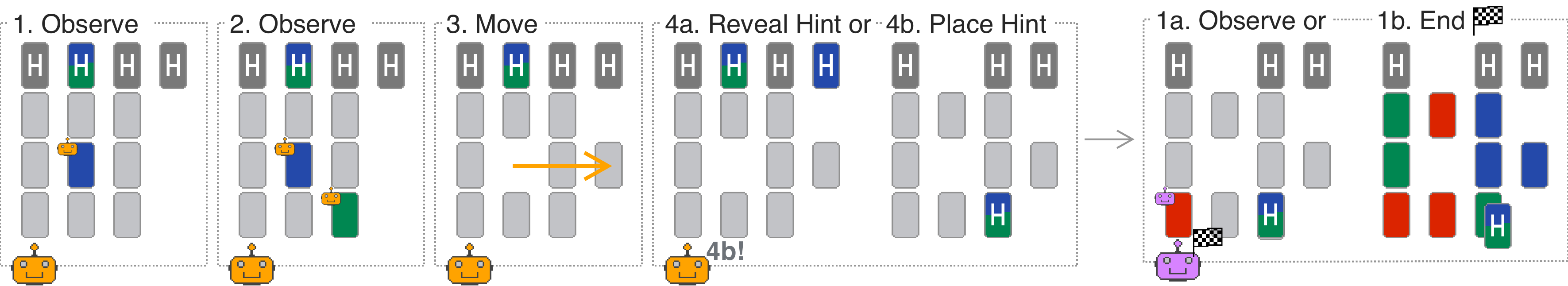}
    \caption{A sample round in nine-card YLE. \orange first observes two cards privately (one \textcolor{cardBlue}{blue}, one \textcolor{cardGreen}{green}), moves the \textcolor{cardBlue}{blue} card, and finishes by either (4a) revealing or (4b) placing a hint. \orange places a hint. Then, \purple can start their turn. \purple chooses to end the game. Ending the game will return the final reward based on the outcome. This game is lost because \textcolor{cardGreen}{green} and \textcolor{cardRed}{red} cards are not clustered.
    }
    \label{fig:turn}
\end{figure*}
\begin{wrapfigure}[18]{R}{0.25\textwidth}
    \includegraphics[width=\linewidth]{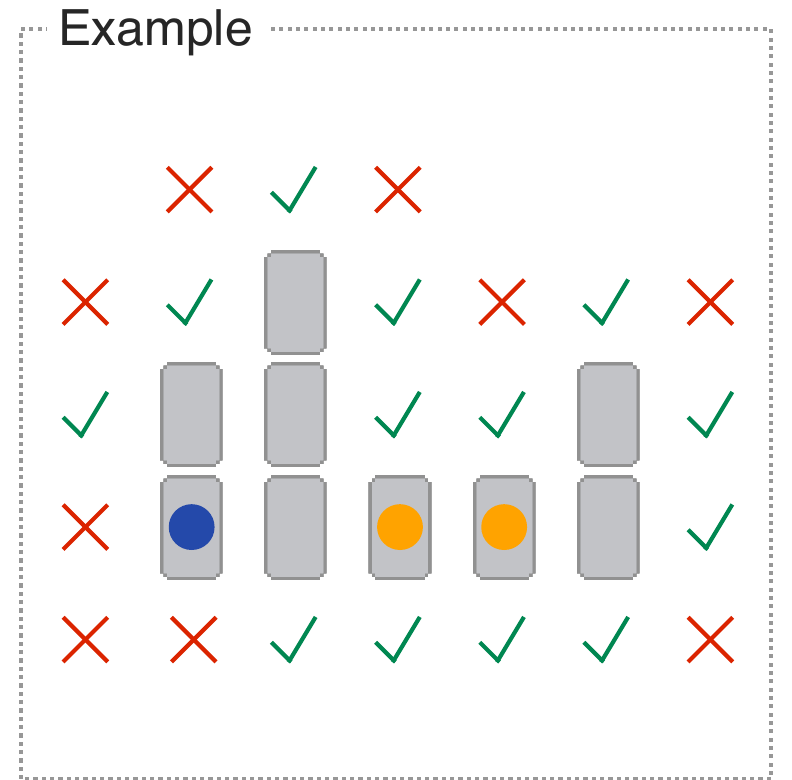}
    \caption{Legal moves for the card with the \textcolor{cardBlue}{blue dot}. Cards can only be moved so that all cards remain connected via their sides. Cards with an \textcolor{orangeAgent}{orange dot} cannot currently be moved.}
    \label{fig:legalmoves}
\end{wrapfigure}
\looseness=-1
\yokai{}\footnote{Detailed rules for \yokai{} are available at \url{http://boardgame.bg/yokai\%20rules.pdf}.} developed by Julien Griffon, is a turn-based cooperative game in which players must sort face-down cards into same-colour clusters to win.
A key requirement is that all cards must be connected to at least one other card on their sides.
Classically, the game is played by two to four players with 16 cards of four colours each.
Complementing these \yokai{} cards, additional \textit{hint cards} feature one or multiple colours and are used to hint at the colour of a face-down card underneath them.
Hint cards are the only allowed communication mechanism in the game, and their number depends on the number of players (see supplementary material; typically 4 -- 10).
These cards are initially face-down and must be revealed before being placed face-up on a \yokai{} card.
Importantly, a hinted \yokai{} card can no longer be moved or observed.
Since most hints can be multi-coloured, their interpretation is left to the other agents, depending on the context and history.

\begin{figure*}[t]
    \centering
    \includegraphics[width=.8\linewidth]{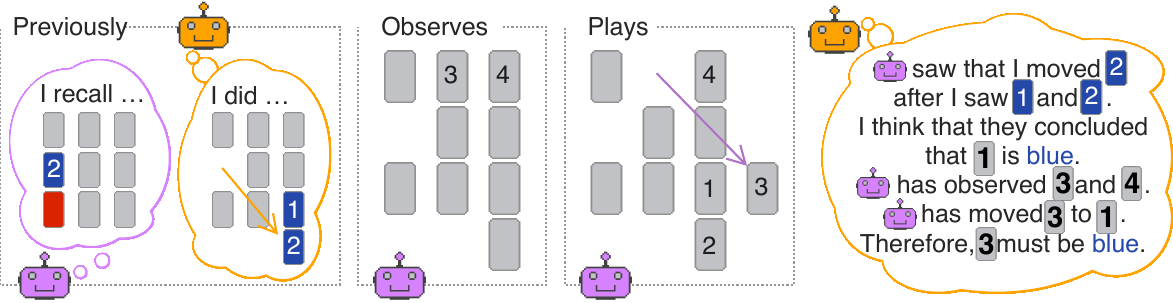}
    \caption{A second-order ToM reasoning example over multiple timesteps in 2-player YLE. \orange beliefs that \purple beliefs that they (\orange) knew that cards 1 and 2 are of the same colour. Even though \purple never saw card 1 and \orange never observed card 3, both now know where all blue cards are, that they are grouped and that both share this knowledge as part of their common ground. In the future, they now can potentially finish early as they have one less card that needs to be observed.}
    \label{fig:2ndOrderToMExample}
\end{figure*}

\looseness=-1
A typical turn in \yokai{} consists of four steps (see \autoref{fig:turn}). 
The current player first observes two cards \textit{privately}, then moves one card, and then either (a) reveals or (b) places a face-up hint card.
The available legal moves depend on the specific card to be moved (an example is shown in \autoref{fig:legalmoves}).
The game ends when all hint cards have been revealed and placed.
Optionally, players can end the game early instead of observing cards at the beginning of their turn.
The final game score is based on the final card configuration:
\begin{equation}
    \label{eq:score}
    \begin{split}
    S = 5N_{\text{hints\_face\_down}} + 2N_{\text{hints\_not\_placed}} + N_{\text{hints\_correct}} - N_{\text{hints\_wrong}}.
    \end{split}
\end{equation}
Players are encouraged to end the game early to achieve a higher score, since unused hint cards contribute most to the final score.
However, the earlier a player ends a game, the less information they can acquire first-hand and instead from others' behaviour (see \autoref{fig:2ndOrderToMExample} for an example).
As such, \textit{the final game score reflects players' joint ToM reasoning abilities}.
In particular, \textit{the rate at which players successfully end the game early reflects how much they rely on belief inference}, making it a powerful new implicit measure of ToM reasoning under uncertainty.

\looseness=-1
\yokai{} poses three key challenges to cooperative agents and ZSC specifically:
First, achieving a high score requires agents to end a game as early as possible, resulting in a higher risk of failure.
Second, ending the game early requires interpreting others' actions and tracking their beliefs to form correct own beliefs over the colours of unobserved cards.
Third, agents must use hints efficiently as grounded communication, but a single misplaced hint can render the game unwinnable.
In AI research, \yokai{} has been explored in the context of formal logics \citep{fernandez2023logical} and in a preliminary study on the feasibility of training self-play RL agents for \yokai{} \citep{yang2024yokai}.

\section{Preliminaries}
\looseness=-1
\paragraph{Dec-POMDP} We model \yokai{} as a \textit{decentralised partially observable Markov decision process} (Dec-POMDP) \citep{Oliehoek2016}, which is a tuple
$\mathcal{M} = \langle \mathcal{N}, \mathcal{S}, \Omega, O, A, \mathcal{R}, \mathcal{T}, \gamma \rangle$.
$\mathcal{N}$ denotes the set of agents, where for $i \in \mathcal{N}$ we let $-i$ denote all agents except $i$.
$\mathcal{S}$ is the set of environment states.
At timestep $t$, agent $i$ in state $s_t$ receives the partial observation $o^i_t = O(s_t, i)$ which lies in the observation space $\Omega$.
In each state $s_t$, agents execute a joint action $\bm{a}_t = (a^1_t, \dots, a^{|\mathcal{N}|}_t)$ lying in the joint action space $A$, after which the environment transitions to state $s_{t+1}$ with probability $\mathcal{T}(s_{t+1} | s_t, a_t)$ and emits a shared reward $r_t = \mathcal{R}(s_t, \bm{a}_t)$.
We let $\tau_t = (s_0, a_0, s_1, a_1, ..., s_t)$ be the state-action history (SAH), and we denote by $\tau_t^i = (o_0^i, a_0^i, o_1^i, ..., o_{t-1}^i, a_{t-1}^i, o_t^i)$ the action observation-history of an agent $i$. We denote by $\pi^i(a_t^i | \tau_t^i)$ the probability that agent $i$ chooses local action $a_t^i$ given AOH $\tau_t^i$. Given $\tau_t$, the joint action $a_t$ is taken with probability $\pi(a_t | \tau_t) := \prod_{i = 1}^n \pi^i(a_t^i | \tau_t^i)$.
We define the expected discounted return of a joint policy $\pi = (\pi^1, ..., \pi^n)$ as
\begin{align}
    J_{\text{SP}}(\pi) = \mathbb{E}_{\tau_T \sim \pi} \left[ \sum_{t=0}^{T-1} \gamma^t \mathcal{R}(s_t, \bm{a}_t) \right].
\end{align}

\paragraph{Zero-Shot Coordination (ZSC)} ZSC \citep{Hu2020Other} aims to design algorithms that produce \emph{compatible} policies in Dec-POMDPs, in that independently trained agents can coordinate.
The compatibility between different joint policies $\pi_1, ..., \pi_n$ in an $n$-player Dec-POMDP is measured by the \emph{cross-play} (XP) return, defined as in \citet{forkel2026highentropyleadssymmetry} as
\begin{align}
    J_{\text{XP}}(\pi_1, ..., \pi_n) := \mathbb{E}_{\phi \sim \text{Perm}(n)} \left[ J_{\text{SP}}((\pi_{\phi(1)}^1, ..., \pi_{\phi(n)}^n ))\right],
\end{align}
where $\mathrm{Perm}(n)$ denotes the set of permutations of $\{1,\dots,n\}$.
A common failure mode in ZSC is that agents learn partner-specific conventions that do not generalise to new partners, which results in XP returns being significantly lower than SP returns.
Algorithms to mitigate this SP-XP gap include Other-Play (OP) \citep{Hu2020Other} and Off-Belief Learning (OBL) \citep{Hu2021OffBelief}.
Here, we focus on the commonly used inter-seed XP setting, where the implementation is fixed but random seeds differ \citep{Hu2020Other,Hu2021OffBelief,Cui2022Off,Muglich2025Expected,forkel2026highentropyleadssymmetry}.

\section{The \yokai{} Learning Environment}
\label{sec:YLEEnv}
\looseness=-1
Our \yokai{} Learning Environment (YLE) implements a parametrised version of the collaborative card game \yokai{}.
We support configurable game sizes (e.g., \textbf{9C} with nine cards and three colours, and \textbf{16C}, the original game) and restrict play to a $g \times g$ grid to enable efficient computation of legal moves.
The grid size $g$ is chosen large enough not to constrain gameplay in practice (see supplementary).
The environment is implemented in JAX in JaxMARL \citep{rutherford2023jaxmarl} and supports end-to-end GPU training at hundreds of thousands of steps per second (see \autoref{tab:steps_per_second}).

\paragraph{\yokai{} as a Graph}
\looseness=-1
We model \yokai{} as a graph to reflect its relational structure and enable a efficient JAX implementation.
At time $t$, the state is a graph $G_t(V, A)$ where vertices $V = Y \cup H$ represent cards and hints, and $A$ is the adjacency matrix encoding spatial neighbourhoods between cards.
Hint cards are not connected to any vertex.
Card colours and hints are randomly assigned at the beginning of each episode.
Moving a card updates the corresponding row and column of $A$ to reflect its new spatial neighbours.
A configuration is legal if all cards remain connected, which we verify via reachability on $A$.
The game is won if, for each colour $c \in C$, all cards of colour $c$ form a connected component.
Legal move actions are computed by evaluating connectivity after candidate moves.

\paragraph{Observation Space}
\looseness=-1
Agents receive either graph-based or image-like observations encoding card properties (e.g., colour, position, lock state, id) and hint features.
Observations contain both public information (topology and hints) and private information (card features), leading to asymmetric knowledge.
In the graph representation, cards are encoded by an adjacency matrix $A_t \in \{0,1\}^{|Y|\times|Y|}$ and node features $F_t \in \mathbb{R}^{(|Y|+|H|)\times f}$.
In the image representation, card features are embedded into a $g \times (g+1) \times f$ image-like tensor according to spatial position.
Agents observe which cards others inspect, but not their content, requiring belief tracking over time.

\paragraph{Action Space}
\looseness=-1
Actions are represented categorically and include observing cards, moving cards, placing or revealing hints, ending the game, and a no-op.
The action space size depends on the configuration and scales with the number of cards $|Y|$, grid positions $g^2$, and hints $|H|$.
In practice, it ranges from approximately $1{,}000$ to $3{,}000$ actions (see supplementary).
Invalid actions are masked at each timestep.
Even with masking, this action space is substantially larger than those of widely used cooperative ZSC benchmarks \citep{Bard2020Hanabi,carroll2019utility,Gessler2025OvercookedV2}%
and therefore challenges existing algorithms with a larger exploration problem.

\paragraph{Reward}
\looseness=-1
The game returns the final score only at termination; the score can be negative due to incorrect hint placements. 
To discourage trivial strategies (e.g., never attempting to win to avoid hint penalties), we modify the terminal reward from \autoref{eq:score}:
\begin{equation}
    R = 
    \begin{cases}
    S & \text{if the game is won},\\
    - d_a - (|C| - c_a) - i_h & \text{otherwise},
    \end{cases}
\end{equation}
where $|C|$ is the number of colours, $c_a$ is the number of correct clusters, $i_h$ the number of incorrect hints (colour not included in the hint), and $d_a \in \{0, 1\}$ indicates early termination.
We additionally apply reward shaping to accelerate learning (rewarding newly observed cards and correct hints, both $+1$ each), which is annealed to zero during training and disabled at evaluation time.

\paragraph{Successful Early Ending (SEE)}
\begin{wraptable}[11]{R}{0.5\linewidth}
    \centering
    \caption{We evaluate steps-per-second (SPS) of YLE on a Nvidia H100 by taking random actions in $n$ parallel environments. YLE scales to hundreds of thousands of SPS.}
    \label{tab:steps_per_second}
    \begin{tabular}{l r r r r}
         \toprule
         \textbf{Num Envs $n$} & \textbf{512} & \textbf{1,024} & \textbf{2,048}  \\
         \midrule
         2-player \textbf{9C} & 139,107 & 274,669 & 559,251 \\
         2-player \textbf{16C} & 135,662 & 144,636 & 148,269 \\
         \bottomrule
    \end{tabular}
\end{wraptable}
Because unused hint cards yield the largest reward, every additional hint implicitly reduces the maximum attainable score, creating strong time pressure to end the game once sufficient common ground has been established. 
We therefore introduce \textit{Successful Early Ending (SEE)} as an additional metric next to reward.
SEE is the rate at which agents terminate the game early and still win. 
It is complementary to average game length: long games indicate overly cautious play, while short but unsuccessful games reflect guessing without sufficient belief formation. 
This positions SEE as a diagnostic signal of functional ToM reasoning, especially when the partner is unknown.

\paragraph{Implementation}
\looseness=-1
Our Jax-based implementation scales efficiently on GPUs, reaching hundreds of thousands of steps per second (see \autoref{tab:steps_per_second}) and performing comparably to other JAX-based RL environments \citep{rutherford2023jaxmarl,nikulin2023xlandminigrid,matthews2024craftax,ruhdorfer2025the}.
Further implementation details are provided in \autoref{sec:envDetails}.

\section{Adapting ZSC Methods to YLE}
\label{sec:method}
\subsection{Other-Play in YLE}
\label{sec:other-play}
Other-Play~\citep{Hu2020Other} modifies self-play by optimizing expected return under random symmetry transformations drawn from the known symmetries of the Dec-POMDP. 
In practice, OP is implemented as domain randomization: at the start of each episode, each agent samples a transformation and interacts with a permuted view of the same underlying environment. 
This discourages convergence to arbitrary symmetry-breaking conventions that fail under cross-play.

In Hanabi, the dominant symmetry corresponds to permutations of card colours.
Since card positions carry no semantic meaning and there is no spatial structure, enforcing invariance to colour permutations is sufficient to prevent most seed-dependent conventions.

In the YLE, symmetry structure is richer due to the spatial layout and card movement. 
Symmetry breaking can arise from both card colours and the spatial configuration. 
Enumerating all Dec-POMDP symmetries in complex environments is generally intractable in practice, which is a known limitation of OP.
We therefore focus on the two most salient symmetry families that directly induce symmetry-breaking: (i) \textbf{colour} permutations and (ii) \textbf{spatial} grid rotations.
Both admit symmetry breaking, e.g., clustering specific colours in fixed regions.
We thus define the symmetry group as $\Phi = \Phi_{\text{col}} \times \Phi_{\text{rot}}$, where $\Phi_{\text{col}}$ permutes colours and $\Phi_{\text{rot}} = \{0^\circ, 90^\circ, 180^\circ, 270^\circ\}$ rotates the grid.

Unlike Hanabi, multiple symmetry classes exist in the YLE. 
Using only a subset may therefore leave residual symmetry-breaking conventions, which we show empirically in \autoref{sec:experiments}.

\subsection{Off-Belief Learning in YLE}
\label{sec:off-belief-learning}
\looseness=-1
Off-Belief Learning~\citep{Hu2021OffBelief} constructs a sequence of policies and belief models. 
At level $\ell$, policy $\pi_\ell$ chooses actions that are optimal assuming (i) past actions were generated by a reference policy $\pi_{\ell-1}$, used to form beliefs, and (ii) future actions will follow $\pi_\ell$. 
Iterating this operator yields a hierarchy $\pi_0 \rightarrow \pi_1 \rightarrow \dots \rightarrow \pi_L$ that controls reasoning depth and produces a unique policy per level given $\pi_0$.
OBL relies on a learned belief model $\mathcal{B}_{\pi}$ that predicts the unknown state-action history $\tau_t$ given a joint policy $\pi$ and an agent’s action-observation history $\tau_t^i$, i.e. $\mathcal{B}_{\pi}(\tau_t | \tau_t^i)$.
In Hanabi, this belief predicts hidden card identities in the observing player’s hand.

\looseness=-1
In YLE, the (private, unknown) state consists of the unobserved colours of face-down Yokai cards. 
Accordingly, we trained $\mathcal{B}_{\pi}$ to predict a posterior over the global board configuration conditioned on the agent’s AOH.
In doing so, we closely followed the methodology of \cite{Hu2021OffBelief}: our belief model features an LSTM \citep{Hochreiter1997Long} encoder and autoregressive decoder that predicts a colour per card slot (i.e. nine slots for nine cards).
Cards have a unique id associated with them in the observation, which makes it possible for the belief model to map cards in observations to these slots.
Per slot, we ensured that cards were sampled consistently with the ego agent's AOH: we respected the global card colour restrictions (i.e. 9 cards of 3 colours) and only sampled cards which are unknown first-hand to the agent from $\mathcal{B}$.

\looseness=-1
To our knowledge, this is the first implementation of OBL outside of Hanabi.
Implementing OBL outside of Hanabi highlights some limitations of OBL as formulated in \cite{Hu2021OffBelief}.
\textbf{First}, hints in the YLE are not guaranteed truthful.
The belief model must therefore treat hints as unreliable evidence, making posterior estimation more challenging than in Hanabi.
\textbf{Second}, OBL theory requires a fictitious rollout length $K$ such that the acting agent moves again.
While two-player Hanabi requires $K=2$, YLE's multi-step turn structure requires $K=4|\mathcal{N}|$, increasing fictitious rollout cost substantially.
In practice, this makes OBL expensive relative to other ZSC methods both in terms of additional environment steps and memory requirements.
Such limitations are highlighted by testing ZSC algorithms in additional environments, a contribution of this work.

\section{Experiments}
\label{sec:experiments}
\looseness=-1
We evaluate our YLE as a benchmark for zero-shot coordination by comparing self-play (SP) and cross-play (XP) performance across leading ZSC methods. 
We first show that our YLE is solvable in self-play, with agents achieving strong performance in the simplest settings, which we contextualise using a small human score estimate for illustration purposes.
However, strong self-play does not translate into strong cross-play. 
Across methods that achieve near-perfect inter-seed XP in Hanabi, we observe persistent SP–XP gaps and substantial degradation in XP performance in the YLE, indicating that existing ZSC approaches do not generalise to its demands.

\subsection{Setup}
\looseness=-1
Agents were trained using independent PPO (IPPO) \citep{deWitt2020is} for $10^9$ steps and evaluated on 5,000 games. 
Our default policy consists of a 4-layer CNN encoder (64 filters, kernel size $(3,3)$, ReLU, stride 1), followed by a linear projection, a GRU, and actor–critic heads. 
All hidden dimensions are 256. 
For OBL, we adopted the standard public–private architecture \citep{Hu2021OffBelief,Cui2022Off} with separate CNN encoders for private and public observations.

\looseness=-1
We optimised with Adam \citep{kingma2015adam} ($\epsilon{=}10^{-5}$), $\gamma{=}0.99$, $\lambda{=}0.95$, value loss coefficient $0.5$, four PPO epochs with four mini-batches of 128 steps, and linearly annealed reward shaping. 
We tuned the PPO entropy coefficient $\alpha$ per setting, sweeping $\alpha \in [0.01, 0.08]$. 
We report results over 12 random seeds per configuration.

\looseness=-1
We report return (\textbf{R}), successful early ending (\textbf{SEE}), and game length (\textbf{LEN}). 
\textbf{SEE} measures how often agents terminate early and still win, and is decomposed into early-ending rate (\textbf{EE}) and calibration $\textbf{WEE}=P(\text{W}|\text{EE})$.
Our primary comparison metric is cross-play return.

\looseness=-1
Maximum scores and episode lengths vary with the number of players and environment size (9C vs 16C) due to the differing number of available hints\footnote{$4$, $5$, $6$ for 2, 3, and 4 players respectively in 9C; $7$, $9$, $10$ in 16C.}. 
In 9C YLE, the maximum episode length is $32$, $40$, and $48$, and the theoretical maximum score is $20$, $25$, and $30$ for 2, 3, and 4 players respectively. 
Achieving this maximum requires immediate early termination without interacting with the board. 
By contrast, a policy that always uses all hints correctly and never terminates early obtains only $4$, $5$, and $6$ respectively. 
Further details are provided in \hyperref[sec:RLinYLE]{Appendix~\ref*{sec:RLinYLE}}.

\subsection{Zero-Shot Coordination Experiments}
\begin{table*}[t]
    \centering
    \caption{CNN-GRU SP and XP performance under different algorithms and settings in the YLE and a small human baseline for illustration purposes (\textbf{H}). To isolate coordination challenges from memory limitations, we use a perfect-memory setting (M; indicated by \cmark) where we show agents the values of already seen cards \textit{even if they have been moved}.
    Full YLE is indicated by \xmark. 
    Each agent entry in the table is the mean $\pm$ standard error of the mean computed from $12$ independent seeds.
    The table shows $180$ trained policies.
    Best results are in \textbf{bold}.
    The column headers are: V = Card version (9C, 16C) and P = \# of players (2, 3, 4). 
    The difference between highest average SP and XP return is $7.5 - 4.8 = 2.7$ (a symmetric percentage difference of $43.9\%$).
    }
    \label{tab:performance}
    \resizebox{\textwidth}{!}{%
    \begin{tabular}{
    l
    c
    c
    c
    S[table-format=1.1(1.1)]
    S[table-format=2(2)]
    S[table-format=3(2)]
    S[table-format=2(2)]
    S[table-format=2.1(1.1)]
    S[table-format=1.1(1.1)]
    S[table-format=2(2)]
    S[table-format=3(2)]
    S[table-format=3(1)]
    S[table-format=2.1(1.1)]
    }
        \toprule
        & & & &
        \multicolumn{5}{c}{\textbf{Self-Play (SP)}} &
        \multicolumn{5}{c}{\textbf{Cross-Play (XP)}} \\
        \cmidrule(lr){5-9} \cmidrule(lr){10-14}
        \textbf{Alg.} & \textbf{V} & \textbf{P} & \textbf{M}
        & \textbf{R $\uparrow$} & \textbf{SEE $\uparrow$} &  \textbf{EE $\uparrow$} & \textbf{WEE $\uparrow$} & \textbf{LEN $\downarrow$}
        & \textbf{R $\uparrow$} & \textbf{SEE $\uparrow$} & \textbf{EE $\uparrow$} &  \textbf{WEE $\uparrow$} & \textbf{LEN $\downarrow$} \\
        \midrule
        H & 9C & 2 & \xmark & \text{--} & \text{--} & \text{--} & \text{--} & \text{--} & \bfseries 5.7(0.5) & \bfseries 65(11) & 65(11) & \bfseries 100(0) & \text{--} \\
        \midrule
        IPPO    & 9C & 2 & \cmark & 7.5(0.3) & 89(7) & 89(8) & 92(8) & 25.0(0.6) & 1.6(0.2) & 18(3) & 35(5) & 50(3) & 29.7(0.4) \\
        HE      & 9C & 2 & \cmark & \bfseries 7.5(0.1) & \bfseries 95(0) & \bfseries 96(0) & \bfseries 99(0) & \bfseries 24.6(0.1) & 4.1(0.8) & 50(1) & 68(10) & 69(7) & 27.2(0.7) \\
        OP      & 9C & 2 & \cmark & 6.6(0.2) & 84(2) & 90(2) & 93(1) & 25.2(0.2) & \bfseries 4.8(0.1) & \bfseries 63(2) & \bfseries 75(2) & \bfseries 83(1) & \bfseries 26.7(0.3) \\
        OBL$_1$  & 9C & 2 & \cmark & 4.0(0.2) & 68(7) & 76(8) & 83(7) & 28.0(0.4) & 2.3(0.3) & 43(7) & 56(9) & 76(2) & 29.4(0.4) \\
        OBL$_2$  & 9C & 2 & \cmark & 4.3(0.2) & 69(7)  & 79(7) & 81(7) & 27.7(0.4) & 2.1(0.2) & 38(4) & 50(5) & 75(2) & 29.5(0.2) \\
        OBL$_3$  & 9C & 2 & \cmark & 4.4(0.4) & 73(7) & 82(7) & 81(8) & 27.3(0.4) & 2.2(0.3) & 41(6) & 55(6) & 73(3) & 29.1(0.4) \\
        OBL$_4$  & 9C & 2 & \cmark & 4.5(0.2) & 77(3) & 87(3) & 89(2) & 27.2(0.2) & 2.8(0.3) & 53(8)  & 66(8) & 78(3) & 28.5(0.5) \\
        OBL$_5$  & 9C & 2 & \cmark & 4.8(0.2) & 80(3) & 89(2) & 90(2) & 27.0(0.2) & 2.5(0.3) & 47(7) & 61(7) & 75(3) & 28.9(0.3) \\
        \midrule
        IPPO & 9C & 3 & \cmark & \bfseries 7.8(0.6) & \bfseries 64(10) & 67(11) & \bfseries 73(12) & 33.4(1.1) & 0.2(0.1) & 0(0) & 6(1) & 17(1) & 39.5(0.1) \\
        HE   & 9C & 3 & \cmark & 0.9(0.0) & 3(0) & \bfseries 100(0) & 3(0) & \bfseries 1.0(0.0) & 0.9(0.0) & 4(0) & \bfseries 100(0) & 4(0) & \bfseries 1.0(0.0) \\
        OP   & 9C & 3 & \cmark & 3.0(0.8) & 27(0) & 38(11) & 37(11) & 36.2(1.1) & \bfseries 1.2(0.2) & \bfseries 10(1) & 29(4) & \bfseries 34(3) & 38.0(0.2) \\ 
        \midrule
        IPPO & 9C & 4 & \cmark & 0.5(0.2) & 3(1) & 50(15) & 3(1) & 38.1(3.0) & 0.1(0.0) & 1(0) & 64(20) & 1(0) & 36.6(3.7) \\
        HE   & 9C & 4 & \cmark & \bfseries 0.7(0.1) & 3(0) & \bfseries 100(0) & 3(0) & \bfseries 17.0(3.9) & \bfseries 0.7(0.0) & \bfseries 2(0) & \bfseries 100(0) & \bfseries 2(0) & \bfseries 6.6(2.1) \\
        OP   & 9C & 4 & \cmark & 0.2(0.0) & 1(0) & 49(14) & 1(0) & 38.8(2.6) & 0.1(0) & 1(0) &  68(6) & 2(0) & 36.5(1.5) \\ 
        \midrule
        IPPO & 16C & 2 & \cmark & 0.0(0.0) & 0(0) & 25(12) & 0(0) & 52.6(1.8) & 0.0(0.0) & 0(0) & 25(16) & 0(0) & 53.2(1.7) \\
        \bottomrule
    \end{tabular}%
    }
\end{table*}
As introduced, we compare standard IPPO as the self-play baseline, high-entropy IPPO (HE), OP, and OBL in a perfect-memory setting to isolate coordination challenges.
For HE, we followed \citeauthor{forkel2026highentropyleadssymmetry}: in 2P we used $\lambda_{\text{GAE}} = 0.85$ and $\alpha = 0.07$, and in multi-player settings we swept $\alpha \in \{0.02, 0.03, \dots, 0.07\}$ and report the result for the lowest $\alpha$ for which SP and XP scores match.
For the other methods, we used $\alpha = 0.05$ in 2P settings and $\alpha = 0.01$ in 3P and 4P settings.
We verify these choices via ablations below.
Our final results are based on training and comparing $400$ policies. 
To anchor results, we additionally establish a rough human performance baseline based on results from $25$ games played by $5$ participants (ages 24 - 28; see \autoref{sec:humanplay}).
Because of the limited sample size, this baseline is included as a rough illustration only.
Our experiments were run on NVIDIA V100, L40S, and H100 GPUs.
A single seed takes between 6 and 12 hours, depending on player and card count.
Due to the computational cost of training multiple levels and the method's performance, we evaluate OBL only in the 2P setting.
We show XP matrices in \autoref{sec:cross-playmatrices} (Figures \ref{fig:sp-2p-cross-play} - \ref{fig:obl-l5-2p-cross-play}) and training curves in \autoref{sec:training-curves}, which show our policies and belief models converging.

\autoref{tab:performance} highlights four key results.
\textbf{First}, state-of-the-art ZSC methods exhibit persistent SP–XP gaps in two- and three-player 9C YLE, indicating failure to learn fully symmetric policies.
Despite the large discrete action space in YLE, agents consistently achieve strong self-play performance in both settings, indicating that the main difficulty arises from coordination.
\textbf{Second}, for two-player 9C YLE XP performance remains below the considered human baseline despite agents having perfect memory.
\textbf{Third}, HE -- the state-of-the-art method for Hanabi -- underperforms OP in two-player 9C YLE.
It further collapses in three- and four-player versions, often learning to terminate immediately or substantially earlier ($\textbf{LEN}=1.0$ and $\textbf{LEN}=17.0$, respectively). 
Increasing entropy makes coordination impossible before reaching high XP scores, suggesting that entropy tuning alone is insufficient for YLE. 
Notably, in 9C YLE this reverses the ranking observed in Hanabi, where HE outperforms OP.
Interestingly, self-play performance increases across OBL levels, while cross-play gains remain modest and inconsistent. 
The observed self-play progression is broadly consistent with the intended behaviour of OBL, whereas the absence of corresponding cross-play improvements remains an interesting observation. We leave explaining this discrepancy to future work.
\textbf{Fourth}, difficulty increases systematically with additional players and in the 16C setting, all of which remain unsolved even w.r.t. SP returns.
These results establish 9C YLE as a challenging ZSC benchmark.

\subsection{Ablations and Additional Experiments}
\label{sec:ablations}
\begin{table*}[t]
    \centering
    \caption{We sweep OP PPO entropy coefficients $\alpha$ to verify our experimental setting.}
    \label{tab:ablating_entropy}
    \resizebox{\textwidth}{!}{%
    \begin{tabular}{
    c
    c
    c
    c
    S[table-format=1.1(1.1)]
    S[table-format=2(1)]
    S[table-format=2(2)]
    S[table-format=2(2)]
    S[table-format=2.1(1.1)]
    S[table-format=1.1(1.1)]
    S[table-format=2(1)]
    S[table-format=2(2)]
    S[table-format=2(2)]
    S[table-format=2.1(1.1)]
    }
        \toprule
        & & & &
        \multicolumn{5}{c}{\textbf{Self-Play (SP)}} &
        \multicolumn{5}{c}{\textbf{Cross-Play (XP)}} \\
        \cmidrule(lr){5-9} \cmidrule(lr){10-14}
        $\alpha$ & \textbf{V} & \textbf{P} & \textbf{M}
        & \textbf{R $\uparrow$} & \textbf{SEE $\uparrow$} &  \textbf{EE $\uparrow$} & \textbf{WEE $\uparrow$} & \textbf{LEN $\downarrow$}
        & \textbf{R $\uparrow$} & \textbf{SEE $\uparrow$} & \textbf{EE $\uparrow$} &  \textbf{WEE $\uparrow$} & \textbf{LEN $\downarrow$} \\
        \midrule
        0.01  & 9C & 2 & \cmark & 3.3(0.3) & 32(8) & 41(10) & 46(11) & 29.3(0.6) & 1.1(0.2) & 12(2) & 28(4) & 47(4) & 30.2(0.4) \\ 
        0.03  & 9C & 2 & \cmark & 6.2(0.1) & \bfseries 85(1) & \bfseries 92(1) & 92(1) & 25.6(0.2) & 4.2(0.1) & 56(2) & \bfseries 76(3) & 74(1) & 26.7(0.3) \\
        \bfseries 0.05  & 9C & 2 & \cmark & \bfseries 6.6(0.2) & 84(2) & 90(2) & \bfseries 93(1) & \bfseries 25.2(0.2) & \bfseries 4.8(0.1) & \bfseries 63(2) & 75(2) & \bfseries 83(1) & \bfseries 26.7(0.3) \\
        0.08  & 9C & 2 & \cmark & 0.1(0.0) &  0(0) & 0(0) & 0(0) & 32.0(0.0) & 0.0(0.0) &  0(0) & 0(0) & 0(0) & 32.0(0.0) \\ 
        \midrule
        \bfseries 0.01  & 9C & 3 & \cmark & \bfseries 3.0(0.8) & \bfseries 27(0) & 38(11) & \bfseries 37(11) & 36.2(1.1) & \bfseries 1.2(0.2) & \bfseries 10(1) & 29(4) & \bfseries 34(3) & 38.0(0.2) \\ 
        0.03  & 9C & 3 & \cmark & 1.0(0.1) & 8(4) & \bfseries 62(13) & 11(6) & \bfseries 19.5(5.3) & 0.1(0.1) & 1(0) & \bfseries 44(14) & 1(0) & \bfseries 24.9(5.0) \\ 
        \midrule
        \bfseries 0.01  & 9C & 4 & \cmark & \bfseries 0.2(0.0) & 1(0) & \bfseries 49(14) & 1(0) & \bfseries 38.8(2.6) & 0.1(0) & 1(0) &  \bfseries 68(6) & \bfseries 2(0) & \bfseries 36.5(1.5) \\ 
        0.03  & 9C & 4 & \cmark & 0.1(0.1) & 1(0) & 17(11) & 1(0) & 42.5(3.9) & 0.1(0.1) & 1(0) & 33(10) & 1(0) & 38.7(3.0) \\ 
        \bottomrule
    \end{tabular}%
    }
\end{table*}

\paragraph{PPO Entropy Coefficient $\alpha$}
The PPO entropy coefficient $\alpha$ plays an important role in ZSC.
This is because it improves SP performance in YLE through increased exploration, and also since high entropy coefficients improve XP performance.
Specifically, \cite{forkel2026highentropyleadssymmetry} show that in Hanabi, near-perfect inter-seed XP can be achieved by picking a high enough $\alpha$ that maximises XP performance while not yet degrading the policy due to the added randomness.
In \autoref{tab:ablating_entropy}, we swept $\alpha$ for OP. 
In the 2P 9C setting, moderate entropy improves XP performance, consistent with prior findings. 
However, further raising entropy does not achieve state-of-the-art YLE inter-seed XP performance and is outperformed by OP with moderate entropy. 
In 3P and 4P 9C settings, increasing entropy rapidly destabilises learning, leading to policy collapse before coordination benefits can materialise.

These results have two implications. 
\textbf{First}, even when combining symmetry-based training and entropy tuning, persistent SP–XP gaps remain in YLE. 
\textbf{Second}, unlike in Hanabi, it appears that inter-seed XP in 9C YLE might not be solvable via entropy tuning alone.

\paragraph{Other-Play Symmetries}
\begin{table*}[t]
    \centering
    \caption{OP with varying symmetry groups (\textbf{G}): full $\Phi$, only $\Phi_{\text{col}}$, and only $\Phi_{\text{rot}}$.
    }
    \label{tab:op-symm-groups-ablations}
    \resizebox{\textwidth}{!}{%
    \begin{tabular}{
    l
    c
    c
    c
    S[table-format=1.1(1.1)]
    S[table-format=2(1)]
    S[table-format=2(1)]
    S[table-format=3(1)]
    S[table-format=2.1(1.1)]
    S[table-format=1.1(1.1)]
    S[table-format=2(1)]
    S[table-format=2(1)]
    S[table-format=2(1)]
    S[table-format=2.1(1.1)]
    }
        \toprule
        & & & &
        \multicolumn{5}{c}{\textbf{Self-Play (SP)}} &
        \multicolumn{5}{c}{\textbf{Cross-Play (XP)}} \\
        \cmidrule(lr){5-9} \cmidrule(lr){10-14}
        \textbf{G} & \textbf{V} & \textbf{P} & \textbf{M}
        & \textbf{R $\uparrow$} & \textbf{SEE $\uparrow$} & \textbf{EE $\uparrow$} & \textbf{WEE $\uparrow$} & \textbf{LEN $\downarrow$}
        & \textbf{R $\uparrow$} & \textbf{SEE $\uparrow$} & \textbf{EE $\uparrow$} & \textbf{WEE $\uparrow$} & \textbf{LEN $\downarrow$} \\
        \midrule
        $\Phi$    & 9C & 2 & \cmark & 6.6(0.2) & 84(2) & 90(2) &  93(1) & 25.2(0.2) & \bfseries 4.8(0.1) & \bfseries 63(2) & \bfseries 75(2) & 83(1) & \bfseries 26.7(0.3) \\
        $\Phi_{\text{col}}$  & 9C & 2 & \cmark & \bfseries 7.6(0.1) & \bfseries 97(0) & \bfseries 97(1) & \bfseries 100(0) & \bfseries 24.7(0.1) & 2.1(0.6) & 24(8) & 39(9) & 55(5) & 29.5(0.7) \\
        $\Phi_{\text{rot}}$ & 9C & 2 & \cmark & 6.4(0.3) & 77(7) & 82(7) & 86(8) & 27.0(0.5) & \bfseries 4.8(0.2) & \bfseries 63(3) & 74(3) & \bfseries 85(2) & 27.0(0.3) \\
        \bottomrule
    \end{tabular}%
    }
\end{table*}
As described in \autoref{sec:other-play}, the YLE symmetry group factorises as $\Phi = \Phi_{\text{col}} \times \Phi_{\text{rot}}$. 
In \autoref{tab:op-symm-groups-ablations}, we compare OP trained with the full group $\Phi$ to ablations using only $\Phi_{\text{col}}$ or $\Phi_{\text{rot}}$ in two-player 9C YLE.
We find that using $\Phi$ and $\Phi_{\text{rot}}$ yields nearly identical XP performance, with $\Phi$ only slightly reducing game length.
In contrast, $\Phi_{\text{col}}$ achieves strong SP but substantially lower XP, underperforming both $\Phi$ and $\Phi_{\text{rot}}$.
This raises an interesting question for implementing OP in practice: How robust is OP to only partial knowledge of symmetry classes?
Our results suggest that not every individual symmetry class is sufficient for achieving good XP performance in 9C YLE.

\paragraph{Alternative Neural Architectures}
\begin{wraptable}[13]{R}{0.6\linewidth}
    \caption{We compare several neural encoders (GATv2, GCN, and RM) and variants with additional hidden layers (+1HL, +2HL) against the IPPO baseline. Our architecture performs best among the evaluated alternatives.}
    \label{tab:ablating_encoders}
    \resizebox{\linewidth}{!}{%
    \begin{tabular}{
    l
    c
    c
    c
    S[table-format=1.1(1.1)]
    S[table-format=2(1)]
    S[table-format=3(2)]
    S[table-format=3(2)]
    S[table-format=2.1(1.1)]
    }
        \toprule
        & & & &
        \multicolumn{5}{c}{\textbf{Self-Play (SP)}} \\
        \cmidrule(lr){5-9}
        \textbf{Alg.} & \textbf{V} & \textbf{P} & \textbf{M}
        & \textbf{R $\uparrow$} & \textbf{SEE $\uparrow$} & \textbf{EE $\uparrow$} & \textbf{WEE $\uparrow$} & \textbf{LEN $\downarrow$} \\
        \midrule
        IPPO & 9C & 2 & \cmark & \bfseries 7.5(0.3) & \bfseries 89(7) & \bfseries  89(8) & 92(8) & \bfseries 25.0(0.6) \\
        \midrule
        GATv2 & 9C & 2 & \cmark & 6.9(0.1) & 87(1) & 87(1) & \bfseries 100(0) & 26.0(0.3) \\ 
        GCN   & 9C & 2 & \cmark & 4.0(0.2) & 9(8) & 9(0) & 25(21) & 31.4(0.5) \\ 
        RM    & 9C & 2 & \cmark & 3.0(1.6) & 25(22) & 25(22) & 25(22) & 29.9(1.8) \\ 
        \midrule
        +1HL   & 9C & 2 & \cmark & 4.0(0.0) & 0(0) & 0(0) & 0(0) & 32.0(0.0) \\
        +2HL   & 9C & 2 & \cmark & 4.0(0.0) & 0(0) & 0(0) & 0(0) & 32.0(0.0) \\
        \bottomrule
    \end{tabular}%
    }
\end{wraptable}
To rule out a suboptimal neural architecture, we compare CNN–GRU to several graph-based encoders motivated by YLE's relational structure: GATv2 \citep{brody2021attentive}, GCN \citep{kipf2016semi}, and Relation Module (RM) \citep{Zambaldi2018Relational}, as well as deeper MLP variants (+1HL, +2HL).
Results in \autoref{tab:ablating_encoders} (four seeds) show that all alternatives perform worse in self-play than the CNN–GRU baseline in two-player 9C YLE.

\paragraph{Zero-Shot Coordination Experiments with Imperfect Memory}
\begin{table*}[t]
    \centering
    \caption{We explore OP in the imperfect-memory setting \yokai{} is typically played in (\xmark). We show that our GRU-based policy is not able to learn in that setting. As an alternative for improved memory, we use TrXL and show that while it improves performance, it fails to match the perfect-memory setting or our small human baseline. We reprint OP with perfect memory for comparison (\cmark).}
    \label{tab:ablating_memory}
    \resizebox{\textwidth}{!}{%
    \begin{tabular}{
    l
    c
    c
    c
    S[table-format=1.1(1.1)]
    S[table-format=2(1)]
    S[table-format=2(2)]
    S[table-format=2(2)]
    S[table-format=2.1(1.1)]
    S[table-format=1.1(1.1)]
    S[table-format=2(1)]
    S[table-format=2(2)]
    S[table-format=2(2)]
    S[table-format=2.1(1.1)]
    }
        \toprule
        & & & &
        \multicolumn{5}{c}{\textbf{Self-Play (SP)}} &
        \multicolumn{5}{c}{\textbf{Cross-Play (XP)}} \\
        \cmidrule(lr){5-9} \cmidrule(lr){10-14}
        \textbf{Alg.} & \textbf{V} & \textbf{P} & \textbf{M}
        & \textbf{R $\uparrow$} & \textbf{SEE $\uparrow$} &  \textbf{EE $\uparrow$} & \textbf{WEE $\uparrow$} & \textbf{LEN $\downarrow$}
        & \textbf{R $\uparrow$} & \textbf{SEE $\uparrow$} & \textbf{EE $\uparrow$} &  \textbf{WEE $\uparrow$} & \textbf{LEN $\downarrow$} \\
        \midrule
        Human& 9C & 2 & \xmark & \text{--} & \text{--} & \text{--} & \text{--} & \text{--} & 5.7(0.5) & 65(11) & \text{--} & \text{--} & \text{--} \\
        \midrule
        OP      & 9C & 2 & \cmark & 6.6(0.2) & 84(2) & 90(2) & 93(1) & 25.2(0.2) & \bfseries 4.8(0.1) & \bfseries 63(2) & 75(2) & 83(1) & \bfseries 26.7(0.3) \\
        \midrule
        OP & 9C & 2 & \xmark & 0.0(0.0) & 0(0) & 0(0) & 0(0) & 32.0(0.0) & 0.0(0.0) & 0(0) & 0(0) & 0(0) & 32.0(0.0) \\
        OP TrXL & 9C & 2 & \xmark & 0.3(0.1) & 1(0) & 33(14) & 1(0) & 21.7(4.2) & 0.3(0.1) & 1(0) & 50(17) & 1(0) & 17.8(4.5) \\
        \bottomrule
    \end{tabular}%
    }
\end{table*}
A central aspect of \yokai{} is that agents must remember the colours of moving cards.
Our previous experiments isolated coordination by providing perfect memory; here we evaluate OP in the standard imperfect-memory setting.
As shown in \autoref{tab:ablating_memory}, the GRU-based OP policy fails completely without memory assistance in the two-player 9C YLE. 
We therefore replaced the GRU with Transformer-XL (TrXL) \citep{dai2019transformerXL,Parisotto2020Stabilizing}, an explicit memory mechanism. 
While TrXL improves performance, it remains below both the perfect-memory setting and the considered human baseline.
This introduces an additional difficulty axis in YLE. 
Beyond increasing player count or scaling to 16C, imperfect memory substantially amplifies coordination challenges, further widening the gap between current ZSC methods and observed human performance.

\begin{wrapfigure}[17]{R}{0.5\linewidth}
    \centering
    \includegraphics[width=\linewidth]{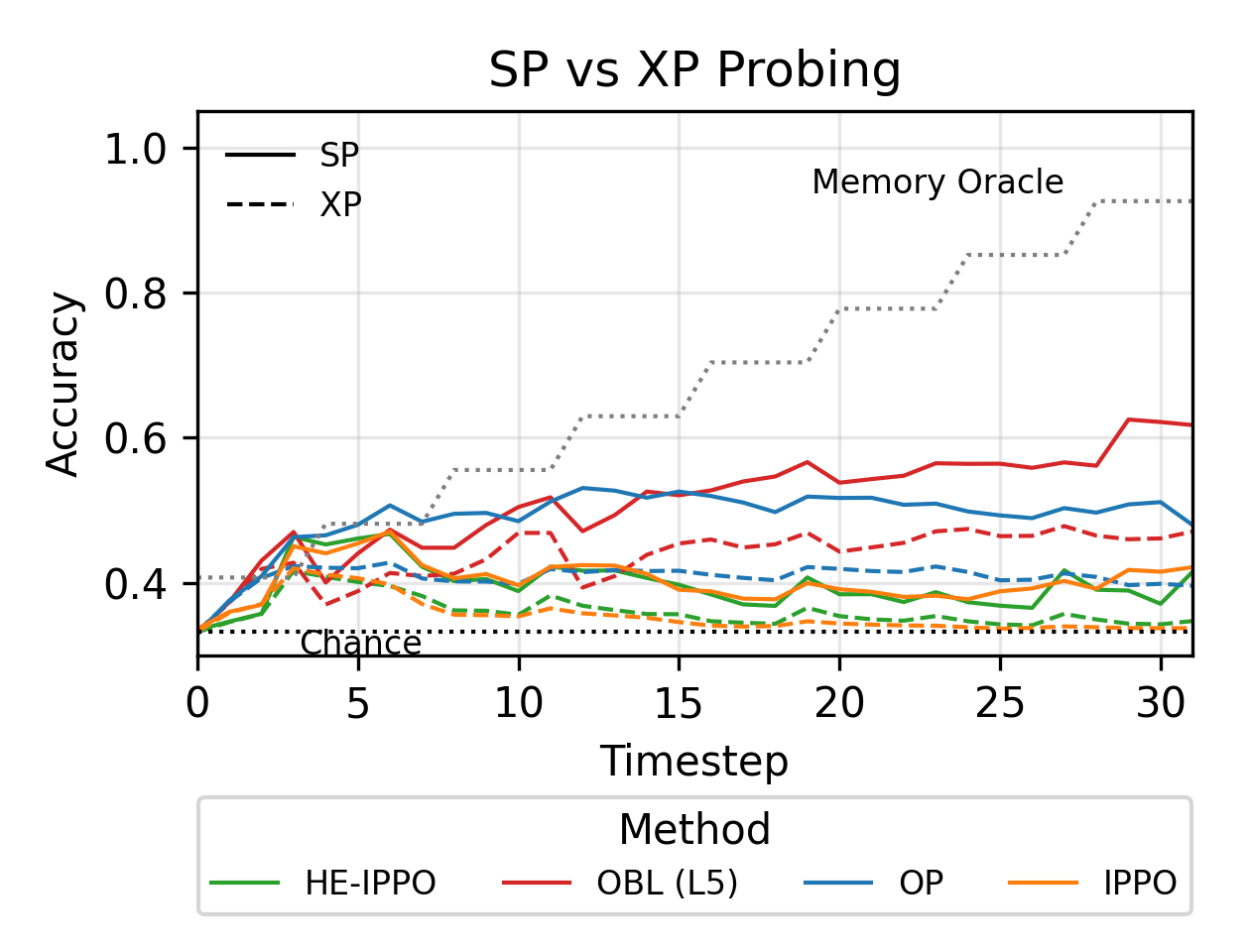}
    \caption{Probing card colours from the hidden state in self-play and cross-play over timesteps.}
    \label{fig:probing}
\end{wrapfigure}
\subsection{Analysing Common Ground Using Linear Probes}
\looseness=-1
To examine the belief representations learned by our agents, we trained linear probes on recurrent hidden states to predict card colours at each board position using an episode-level split: for each policy, we held out $20\%$ of self-play episodes and trained on the remaining episodes. 
We evaluated on disjoint held-out SP episodes as well as XP episodes by applying each policy’s probe to that policy’s hidden states during cross-play. 
Probe accuracy serves as a linear readout of the encoded colour and thus as a diagnostic of first-order belief representations.
As shown in \autoref{fig:probing}, card colours are linearly decodable above chance in SP across methods, but probe accuracy consistently decreases in XP and remains below that of the perfect-memory oracle.
This reduction in decodability mirrors the observed degradation in performance, suggesting that belief representations become less accessible when agents interact with unseen partners.
Additional results in which we stratify by observed vs unobserved cards are shown in \autoref{sec:probingSetupAppendix} and match our observations.

\section{Discussion}
\label{sec:discussion}
\paragraph{YLE is a timely new ZSC benchmark.}
\looseness=-1
There is a long line of research on ZSC that almost exclusively focused on Hanabi as a benchmark \citep{Hu2020Other,Hu2021OffBelief,Cui2022Off,Muglich2025Expected,forkel2026highentropyleadssymmetry}.
However, over the last few years, Hanabi scores have reached near optimal performance.
Starting with OP, which produced a SP–XP gap in two-player Hanabi of $8.6\%$ ($22.07$ vs $24.06$ \citep{Hu2020Other}).
Subsequent work progressively narrowed this gap and \cite{forkel2026highentropyleadssymmetry} most recently achieved near-perfect inter-seed XP performance with a gap below $0.1\%$ ($24.47$ vs $24.48$) in two-player Hanabi, while also achieving similar performance for all other player counts.
While this constitutes a major achievement, it also implies diminishing returns for using Hanabi as the primary stress test for ZSC.

\looseness=-1
Saturation in Hanabi does not imply that ZSC is solved more broadly.
While both are cooperative Dec-POMDPs, YLE differs structurally in several key respects. 
Beliefs over cards must be tracked across dynamic spatial reconfigurations, hints are ambiguous rather than guaranteed truthful, and agents must decide when sufficient common ground has been established to terminate early. 
In Hanabi, by contrast, beliefs are tied to hand slots, hints are truthful, and there is no termination decision. 

\looseness=-1
Our results in the YLE show persistent and substantial SP-XP gaps across the same families of methods in the evaluated settings.
Even OP exhibits large degradations in both early-ending frequency and early-ending calibration in cross-play.
In particular, XP calibration (WEE) drops close to chance level in several settings, indicating a breakdown of shared belief formation rather than merely conservative stopping behaviour.
Importantly, the majority of our experiments focus on the simplest YLE configuration (two players, memory assistance, nine cards), indicating substantial headroom for further progress. 

\looseness=-1
Our results also suggest that method rankings in two-player 9C YLE also reverse relative to Hanabi: whereas HE $>$ OBL $>$ OP in Hanabi, we observe OP $>$ HE $>$ OBL in YLE. 
This highlights the risk of overfitting to a single benchmark and motivates evaluation across diverse environments.

\paragraph{YLE highlights potential limitations in existing ZSC methods.}
\looseness=-1
Beyond SP-XP gaps, YLE highlights structural assumptions that underlay current ZSC approaches. As discussed in \autoref{sec:method}, OBL incurs substantially higher interaction costs due to longer fictitious rollouts and must operate under ambiguous hints, making belief modelling more difficult than in Hanabi.
OP assumes that relevant symmetry classes of the Dec-POMDP are known and can be enumerated. 
In the YLE, however, symmetry structure is richer due to spatial layout and card movement, and multiple interacting symmetry sources exist. 
Enumerating all such symmetries is difficult in practice, and, as we show, enforcing invariance to only a subset may leave residual symmetry-breaking conventions. 
Finally, while HE effectively coordinates in Hanabi, increasing entropy in YLE appears to destabilise the learned policy in the tested settings before optimal cross-play performance is reached, particularly in multi-player settings.

Taken together, these results indicate that techniques successful in Hanabi do not directly transfer to YLE. 
The reversal in method rankings further suggests that apparent ZSC progress may be benchmark-specific, underscoring the need for structurally diverse evaluation environments.

\paragraph{YLE might not merely be a harder RL environment but a harder ZSC task.}
\looseness=-1
While YLE increases several dimensions of task complexity relative to Hanabi, our results suggest that the observed failures cannot be explained solely by general RL difficulty.
In 2P and 3P 9C YLE with memory assistance, agents achieve strong SP performance while exhibiting substantial SP-XP gaps.
If exploration or credit assignment were the dominant bottlenecks, SP performance would be expected to degrade similarly.
Instead, the degradation appears specifically in coordination with independently trained partners.
This suggests that these YLE versions introduce zero-shot coordination-specific challenges beyond simply scaling the difficulty of the underlying RL problem.

However, larger versions of our YLE currently appear of of reach even in self-play and as such can not yet serve as benchmark for ZSC.
As such we recommend future ZSC work to focus on the 9C version with perfect memory.

\section{Related Work}
\label{sec:relatedWorks}
\looseness=-1
Current benchmarks study ToM reasoning in AI primarily from a static observer’s perspective \citep{baker2011bayesian,baker2017rational,rabinowitz2018machine,bara2021mindcraft,bara2023towards,fan2021learning,duan2022boss,kim2023fantom,bortoletto24explicit,gandhi2024understanding,bortoletto25tomssi}, often framed as supervised learning problems \citep{Aru2023Mind,bortoletto24Limits}. 
In contrast, interactive benchmarks such as the Hanabi \citep{Bard2020Hanabi} and SymmToM \citep{Sclar2022Symmetric} evaluate ToM in multi-agent interactive settings. 
YLE extends interactive ToM benchmarks by introducing dynamic entity movement, ambiguous communication, and early termination decisions within a cooperative Dec-POMDP and relates it to zero-shot coordination.
\yokai{} itself was first explored in AI research in the context of formal logics \citep{fernandez2023logical}.
Additionally, \yokai{} was first explored as a reinforcement learning environment in a preliminary study on the feasibility of training self-play agents for \yokai{} \citep{yang2024yokai}, which implemented an initial JAX-based environment for a simplified version of the game.
Informed by their work, our work instead introduces the YLE as an open-source benchmark for zero-shot coordination, providing complete game mechanics including early ending, configurable game sizes, new coordination metrics based on early termination, and support for evaluating and comparing modern ZSC algorithms.

\looseness=-1
A second line of work investigates whether neural networks internally represent mental states \citep{bortoletto24Limits,zhu2024language}. 
For example, \citet{Matiisen2022Do} probe RL agents for intention representations in cooperative navigation. 
We extend probing analyses to YLE, where hidden representations must encode evolving beliefs under partial observability.

\looseness=-1
Alongside YLE, OvercookedV2 \citep{Gessler2025OvercookedV2} has recently been proposed as a benchmark for ZSC, augmenting the Overcooked environment \citep{carroll2019utility} with partial observability and asymmetric information.
YLE differs in that it introduces substantially larger state and action spaces and, in configurations with more than three players or 16 \yokai{} cards, poses a significant challenge even in SP.
Additionally, YLE offers a full reference implementation of OBL, which previously achieved state-of-the-art inter-seed cross-play performance in the HLE.
Our results show that methods with strong performance in Hanabi perform substantially worse in YLE and that algorithm rankings change, highlighting the importance of diverse benchmarks for evaluating ZSC algorithms.
YLE also contributes to the broader family of cooperative multi-agent RL benchmarks \citep{samvelyan2019starcraft,ruhdorfer2025the,tomilin2025mealbenchmarkcontinualmultiagent}, with a particular emphasis on belief tracking, ToM, and common ground formation.

\looseness=-1
Finally, we consider the ZSC setting as introduced by \cite{Hu2020Other} and further clarified in \citep{treutlein2021new}.
Other broader notions of coordination with novel partners exist and are often studied under the label of ad-hoc teamwork (AHT) \citep{stone2010ad}.
Representative AHT methods are typically population- and diversity-based and include \citep{Strouse2021Collaborating,Keane2022AnyPlay,Xue2022Heterogeneous,Zhao2023Maximum,Yan2023An,Hui2025Efficient,Ruhdorfer2025Unsupervised}. 
In this work, we focus on ZSC first to isolate the coordination challenges posed by the YLE, reasoning that AHT progress on the YLE is only possible after making progress on the ZSC problem.
We  thus view evaluating AHT methods in the YLE as valuable future work after ZSC performance has improved, similar to the progression observed in Hanabi research \citep{Dizdarevic2025AdHoc}.

\section{Conclusion}
\looseness=-1
We introduced the \yokai{} Learning Environment as a new multi-agent RL benchmark for zero-shot coordination in which effective collaboration requires tracking and maintaining beliefs over space and time.
Implemented in JAX, YLE supports end-to-end GPU training at hundreds of thousands of steps per second.
Empirically, our results suggest that state-of-the-art ZSC methods that achieve near-perfect inter-seed cross-play in Hanabi exhibit persistent SP-XP gaps in nine card YLE, including degraded early-ending frequency and calibration.
These results indicate that coordination in YLE places additional structural demands on belief formation and common ground reasoning and that apparent ZSC progress may be benchmark-specific.
Our YLE thus provides a complementary benchmark for evaluating ZSC beyond current environments.

\subsubsection*{Broader Impact Statement}
\label{sec:broaderImpact}
This work introduces a new environment for studying fundamental aspects of cooperation in reinforcement learning agents.
As such, it supports basic research and is currently far removed from direct societal deployment.
However, because our goal is to improve human–AI collaboration, caution is warranted when developing systems.
Any future applications involving real users must prioritise safety, transparency, and accountability.

\appendix

\subsubsection*{Acknowledgments}
\label{sec:ack}
The authors thank the International Max Planck Research School for Intelligent Systems (IMPRS-IS) for supporting C. Ruhdorfer.
We especially thank R. Yang for implementing a prototype of the environment as part of her master’s thesis, and A. Penzkofer and L. Shi for numerous insightful discussions.
J. Foerster is partially funded and J. Forkel is funded by the UKRI grant EP/Y028481/1 (originally selected for funding by the ERC).

\bibliography{refs,tom_refs,tom_refs_deeplearning}

\begin{thebibliography}{75}
\providecommand{\natexlab}[1]{#1}
\providecommand{\url}[1]{\texttt{#1}}
\expandafter\ifx\csname urlstyle\endcsname\relax
  \providecommand{\doi}[1]{DOI: #1}\else
  \providecommand{\doi}{DOI: \begingroup \urlstyle{rm}\Url}\fi

\bibitem[Abdessaied et~al.(2024)Abdessaied, Shi, and Bulling]{abdessaied24_wacv}
Adnen Abdessaied, Lei Shi, and Andreas Bulling.
\newblock {VD-GR:} boosting visual dialog with cascaded spatial-temporal multi-modal graphs.
\newblock In \emph{{IEEE/CVF} Winter Conference on Applications of Computer Vision, {WACV} 2024, Waikoloa, HI, USA, January 3-8, 2024}, pp.\  5793--5802. {IEEE}, 2024.
\newblock \doi{10.1109/WACV57701.2024.00570}.

\bibitem[Aru et~al.(2023)Aru, Labash, Corcoll, and Vicente]{Aru2023Mind}
Jaan Aru, Aqeel Labash, Oriol Corcoll, and Raul Vicente.
\newblock Mind the gap: challenges of deep learning approaches to theory of mind.
\newblock \emph{Artificial Intelligence Review}, 56\penalty0 (9):\penalty0 9141–9156, January 2023.
\newblock ISSN 1573-7462.
\newblock \doi{10.1007/s10462-023-10401-x}.
\newblock URL \url{http://dx.doi.org/10.1007/s10462-023-10401-x}.

\bibitem[Baker et~al.(2011)Baker, Saxe, and Tenenbaum]{baker2011bayesian}
Chris Baker, Rebecca Saxe, and Joshua Tenenbaum.
\newblock Bayesian theory of mind: Modeling joint belief-desire attribution.
\newblock In \emph{Proceedings of the annual meeting of the cognitive science society}, volume~33, 2011.

\bibitem[Baker et~al.(2017)Baker, Jara-Ettinger, Saxe, and Tenenbaum]{baker2017rational}
Chris~L Baker, Julian Jara-Ettinger, Rebecca Saxe, and Joshua~B Tenenbaum.
\newblock Rational quantitative attribution of beliefs, desires and percepts in human mentalizing.
\newblock \emph{Nature Human Behaviour}, 1\penalty0 (4):\penalty0 0064, 2017.

\bibitem[Bara et~al.(2021)Bara, CH-Wang, and Chai]{bara2021mindcraft}
Cristian-Paul Bara, Sky CH-Wang, and Joyce Chai.
\newblock {M}ind{C}raft: Theory of mind modeling for situated dialogue in collaborative tasks.
\newblock In Marie-Francine Moens, Xuanjing Huang, Lucia Specia, and Scott Wen-tau Yih (eds.), \emph{Proceedings of the 2021 Conference on Empirical Methods in Natural Language Processing}, pp.\  1112--1125, Online and Punta Cana, Dominican Republic, November 2021. Association for Computational Linguistics.
\newblock \doi{10.18653/v1/2021.emnlp-main.85}.
\newblock URL \url{https://aclanthology.org/2021.emnlp-main.85}.

\bibitem[Bara et~al.(2023)Bara, Ma, Yu, Shah, and Chai]{bara2023towards}
Cristian{-}Paul Bara, Ziqiao Ma, Yingzhuo Yu, Julie Shah, and Joyce Chai.
\newblock Towards collaborative plan acquisition through theory of mind modeling in situated dialogue.
\newblock In \emph{Proceedings of the Thirty-Second International Joint Conference on Artificial Intelligence, {IJCAI} 2023, 19th-25th August 2023, Macao, SAR, China}, pp.\  2958--2966. ijcai.org, 2023.
\newblock \doi{10.24963/IJCAI.2023/330}.
\newblock URL \url{https://doi.org/10.24963/ijcai.2023/330}.

\bibitem[Bard et~al.(2020)Bard, Foerster, Chandar, Burch, Lanctot, Song, Parisotto, Dumoulin, Moitra, Hughes, Dunning, Mourad, Larochelle, Bellemare, and Bowling]{Bard2020Hanabi}
Nolan Bard, Jakob~N. Foerster, Sarath Chandar, Neil Burch, Marc Lanctot, H.~Francis Song, Emilio Parisotto, Vincent Dumoulin, Subhodeep Moitra, Edward Hughes, Iain Dunning, Shibl Mourad, Hugo Larochelle, Marc~G. Bellemare, and Michael Bowling.
\newblock The {Hanabi} challenge: {A} new frontier for {AI} research.
\newblock \emph{Artif. Intell.}, 280:\penalty0 103216, 2020.
\newblock \doi{10.1016/J.ARTINT.2019.103216}.

\bibitem[Bortoletto et~al.(2024{\natexlab{a}})Bortoletto, Ruhdorfer, Abdessaied, Shi, and Bulling]{bortoletto24Limits}
Matteo Bortoletto, Constantin Ruhdorfer, Adnen Abdessaied, Lei Shi, and Andreas Bulling.
\newblock Limits of theory of mind modelling in dialogue-based collaborative plan acquisition.
\newblock In Lun{-}Wei Ku, Andre Martins, and Vivek Srikumar (eds.), \emph{Proceedings of the 62nd Annual Meeting of the Association for Computational Linguistics (Volume 1: Long Papers), {ACL} 2024, Bangkok, Thailand, August 11-16, 2024}, pp.\  4856--4871. Association for Computational Linguistics, 2024{\natexlab{a}}.
\newblock \doi{10.18653/V1/2024.ACL-LONG.266}.
\newblock URL \url{https://doi.org/10.18653/v1/2024.acl-long.266}.

\bibitem[Bortoletto et~al.(2024{\natexlab{b}})Bortoletto, Ruhdorfer, Shi, and Bulling]{bortoletto24explicit}
Matteo Bortoletto, Constantin Ruhdorfer, Lei Shi, and Andreas Bulling.
\newblock Explicit modelling of theory of mind for belief prediction in nonverbal social interactions.
\newblock In Ulle Endriss, Francisco~S. Melo, Kerstin Bach, Alberto Jos{\'{e}}~Bugar{\'{\i}}n Diz, Jose~Maria Alonso{-}Moral, Sen{\'{e}}n Barro, and Fredrik Heintz (eds.), \emph{{ECAI} 2024 - 27th European Conference on Artificial Intelligence, 19-24 October 2024, Santiago de Compostela, Spain - Including 13th Conference on Prestigious Applications of Intelligent Systems {(PAIS} 2024)}, volume 392 of \emph{Frontiers in Artificial Intelligence and Applications}, pp.\  866--873. {IOS} Press, 2024{\natexlab{b}}.
\newblock \doi{10.3233/FAIA240573}.
\newblock URL \url{https://doi.org/10.3233/FAIA240573}.

\bibitem[Bortoletto et~al.(2024{\natexlab{c}})Bortoletto, Shi, and Bulling]{bortoletto2024neural}
Matteo Bortoletto, Lei Shi, and Andreas Bulling.
\newblock Neural reasoning about agents' goals, preferences, and actions.
\newblock In Michael~J. Wooldridge, Jennifer~G. Dy, and Sriraam Natarajan (eds.), \emph{Thirty-Eighth {AAAI} Conference on Artificial Intelligence, {AAAI} 2024, Thirty-Sixth Conference on Innovative Applications of Artificial Intelligence, {IAAI} 2024, Fourteenth Symposium on Educational Advances in Artificial Intelligence, {EAAI} 2024, February 20-27, 2024, Vancouver, Canada}, pp.\  456--464. {AAAI} Press, 2024{\natexlab{c}}.
\newblock \doi{10.1609/AAAI.V38I1.27800}.
\newblock URL \url{https://doi.org/10.1609/aaai.v38i1.27800}.

\bibitem[Bortoletto et~al.(2025)Bortoletto, Ruhdorfer, and Bulling]{bortoletto25tomssi}
Matteo Bortoletto, Constantin Ruhdorfer, and Andreas Bulling.
\newblock {ToM-SSI}: Evaluating theory of mind in situated social interactions.
\newblock In Christos Christodoulopoulos, Tanmoy Chakraborty, Carolyn Rose, and Violet Peng (eds.), \emph{Proceedings of the 2025 Conference on Empirical Methods in Natural Language Processing, {EMNLP} 2025, Suzhou, China, November 4-9, 2025}, pp.\  32264--32289. Association for Computational Linguistics, 2025.
\newblock \doi{10.18653/V1/2025.EMNLP-MAIN.1642}.

\bibitem[Bradbury et~al.(2018)Bradbury, Frostig, Hawkins, Johnson, Leary, Maclaurin, Necula, Paszke, Vander{P}las, Wanderman-{M}ilne, and Zhang]{jax2018github}
James Bradbury, Roy Frostig, Peter Hawkins, Matthew~James Johnson, Chris Leary, Dougal Maclaurin, George Necula, Adam Paszke, Jake Vander{P}las, Skye Wanderman-{M}ilne, and Qiao Zhang.
\newblock {JAX}: composable transformations of {P}ython+{N}um{P}y programs, 2018.
\newblock URL \url{http://github.com/google/jax}.

\bibitem[Brody et~al.(2022)Brody, Alon, and Yahav]{brody2021attentive}
Shaked Brody, Uri Alon, and Eran Yahav.
\newblock How attentive are graph attention networks?
\newblock In \emph{The Tenth International Conference on Learning Representations, {ICLR} 2022, Virtual Event, April 25-29, 2022}. OpenReview.net, 2022.
\newblock URL \url{https://openreview.net/forum?id=F72ximsx7C1}.

\bibitem[Carroll et~al.(2019)Carroll, Shah, Ho, Griffiths, Seshia, Abbeel, and Dragan]{carroll2019utility}
Micah Carroll, Rohin Shah, Mark~K. Ho, Tom Griffiths, Sanjit~A. Seshia, Pieter Abbeel, and Anca~D. Dragan.
\newblock On the utility of learning about humans for human-{AI} coordination.
\newblock In Hanna~M. Wallach, Hugo Larochelle, Alina Beygelzimer, Florence d'Alch{\'{e}}{-}Buc, Emily~B. Fox, and Roman Garnett (eds.), \emph{Advances in Neural Information Processing Systems 32: Annual Conference on Neural Information Processing Systems 2019, NeurIPS 2019, December 8-14, 2019, Vancouver, BC, Canada}, pp.\  5175--5186, 2019.

\bibitem[Cho et~al.(2014)Cho, van Merrienboer, G{\"{u}}l{\c{c}}ehre, Bahdanau, Bougares, Schwenk, and Bengio]{cho2014learning}
Kyunghyun Cho, Bart van Merrienboer, {\c{C}}aglar G{\"{u}}l{\c{c}}ehre, Dzmitry Bahdanau, Fethi Bougares, Holger Schwenk, and Yoshua Bengio.
\newblock Learning phrase representations using {RNN} encoder-decoder for statistical machine translation.
\newblock In Alessandro Moschitti, Bo~Pang, and Walter Daelemans (eds.), \emph{Proceedings of the 2014 Conference on Empirical Methods in Natural Language Processing, {EMNLP} 2014, October 25-29, 2014, Doha, Qatar, {A} meeting of SIGDAT, a Special Interest Group of the {ACL}}, pp.\  1724--1734. {ACL}, 2014.
\newblock \doi{10.3115/V1/D14-1179}.

\bibitem[Clark(1996)]{clark1996using}
Herbert~H Clark.
\newblock \emph{Using language}.
\newblock Cambridge university press, 1996.

\bibitem[Cui et~al.(2021)Cui, Hu, Pineda, and Foerster]{Cui2021KLevel}
Brandon Cui, Hengyuan Hu, Luis Pineda, and Jakob~N. Foerster.
\newblock K-level reasoning for zero-shot coordination in {Hanabi}.
\newblock In Marc'Aurelio Ranzato, Alina Beygelzimer, Yann~N. Dauphin, Percy Liang, and Jennifer~Wortman Vaughan (eds.), \emph{Advances in Neural Information Processing Systems 34: Annual Conference on Neural Information Processing Systems 2021, NeurIPS 2021, December 6-14, 2021, virtual}, pp.\  8215--8228, 2021.

\bibitem[Cui et~al.(2022)Cui, Hu, Lupu, Sokota, and Foerster]{Cui2022Off}
Brandon Cui, Hengyuan Hu, Andrei Lupu, Samuel Sokota, and Jakob~N. Foerster.
\newblock Off-team learning.
\newblock In Sanmi Koyejo, S.~Mohamed, A.~Agarwal, Danielle Belgrave, K.~Cho, and A.~Oh (eds.), \emph{Advances in Neural Information Processing Systems 35: Annual Conference on Neural Information Processing Systems 2022, NeurIPS 2022, New Orleans, LA, USA, November 28 - December 9, 2022}, 2022.

\bibitem[Dafoe et~al.(2020)Dafoe, Hughes, Bachrach, Collins, McKee, Leibo, Larson, and Graepel]{Dafoe2020Open}
Allan Dafoe, Edward Hughes, Yoram Bachrach, Tantum Collins, Kevin~R. McKee, Joel~Z. Leibo, Kate Larson, and Thore Graepel.
\newblock Open problems in cooperative {AI}.
\newblock \emph{CoRR}, abs/2012.08630, 2020.
\newblock URL \url{https://arxiv.org/abs/2012.08630}.

\bibitem[Dafoe et~al.(2021)Dafoe, Bachrach, Hadfield, Horvitz, Larson, and Graepel]{dafoe2021cooperative}
Allan Dafoe, Yoram Bachrach, Gillian Hadfield, Eric Horvitz, Kate Larson, and Thore Graepel.
\newblock Cooperative {AI}: {M}achines must learn to find common ground.
\newblock \emph{Nature}, 593\penalty0 (7857):\penalty0 33–36, May 2021.
\newblock ISSN 1476-4687.
\newblock \doi{10.1038/d41586-021-01170-0}.
\newblock URL \url{http://dx.doi.org/10.1038/d41586-021-01170-0}.

\bibitem[Dai et~al.(2019)Dai, Yang, Yang, Carbonell, Le, and Salakhutdinov]{dai2019transformerXL}
Zihang Dai, Zhilin Yang, Yiming Yang, Jaime~G. Carbonell, Quoc~Viet Le, and Ruslan Salakhutdinov.
\newblock {Transformer-XL}: Attentive language models beyond a fixed-length context.
\newblock In Anna Korhonen, David~R. Traum, and Llu{\'{\i}}s M{\`{a}}rquez (eds.), \emph{Proceedings of the 57th Conference of the Association for Computational Linguistics, {ACL} 2019, Florence, Italy, July 28- August 2, 2019, Volume 1: Long Papers}, pp.\  2978--2988. Association for Computational Linguistics, 2019.
\newblock \doi{10.18653/V1/P19-1285}.

\bibitem[de~Witt et~al.(2020)de~Witt, Gupta, Makoviichuk, Makoviychuk, Torr, Sun, and Whiteson]{deWitt2020is}
Christian~Schr{\"{o}}der de~Witt, Tarun Gupta, Denys Makoviichuk, Viktor Makoviychuk, Philip H.~S. Torr, Mingfei Sun, and Shimon Whiteson.
\newblock Is independent learning all you need in the {StarCraft} multi-agent challenge?
\newblock \emph{CoRR}, abs/2011.09533, 2020.
\newblock URL \url{https://arxiv.org/abs/2011.09533}.

\bibitem[DeepMind et~al.(2020)DeepMind, Babuschkin, Baumli, Bell, Bhupatiraju, Bruce, Buchlovsky, Budden, Cai, Clark, Danihelka, Dedieu, Fantacci, Godwin, Jones, Hemsley, Hennigan, Hessel, Hou, Kapturowski, Keck, Kemaev, King, Kunesch, Martens, Merzic, Mikulik, Norman, Papamakarios, Quan, Ring, Ruiz, Sanchez, Sartran, Schneider, Sezener, Spencer, Srinivasan, Stanojevi\'{c}, Stokowiec, Wang, Zhou, and Viola]{deepmind2020jax}
DeepMind, Igor Babuschkin, Kate Baumli, Alison Bell, Surya Bhupatiraju, Jake Bruce, Peter Buchlovsky, David Budden, Trevor Cai, Aidan Clark, Ivo Danihelka, Antoine Dedieu, Claudio Fantacci, Jonathan Godwin, Chris Jones, Ross Hemsley, Tom Hennigan, Matteo Hessel, Shaobo Hou, Steven Kapturowski, Thomas Keck, Iurii Kemaev, Michael King, Markus Kunesch, Lena Martens, Hamza Merzic, Vladimir Mikulik, Tamara Norman, George Papamakarios, John Quan, Roman Ring, Francisco Ruiz, Alvaro Sanchez, Laurent Sartran, Rosalia Schneider, Eren Sezener, Stephen Spencer, Srivatsan Srinivasan, Milo\v{s} Stanojevi\'{c}, Wojciech Stokowiec, Luyu Wang, Guangyao Zhou, and Fabio Viola.
\newblock The {D}eep{M}ind {JAX} {E}cosystem, 2020.
\newblock URL \url{http://github.com/google-deepmind}.

\bibitem[Dizdarevic et~al.(2025)Dizdarevic, Hammond, Gessler, Calinescu, Cook, Gallici, Lupu, and Foerster]{Dizdarevic2025AdHoc}
Tin Dizdarevic, Ravi Hammond, Tobias Gessler, Anisoara Calinescu, Jonathan Cook, Matteo Gallici, Andrei Lupu, and Jakob~Nicolaus Foerster.
\newblock Ad-hoc human-{AI} coordination challenge.
\newblock In Aarti Singh, Maryam Fazel, Daniel Hsu, Simon Lacoste{-}Julien, Felix Berkenkamp, Tegan Maharaj, Kiri Wagstaff, and Jerry Zhu (eds.), \emph{Forty-second International Conference on Machine Learning, {ICML} 2025, Vancouver, BC, Canada, July 13-19, 2025}, volume 267 of \emph{Proceedings of Machine Learning Research}. {PMLR} / OpenReview.net, 2025.

\bibitem[Duan et~al.(2022)Duan, Yu, Tan, Yi, and Tan]{duan2022boss}
Jiafei Duan, Samson Yu, Nicholas Tan, Li~Yi, and Cheston Tan.
\newblock {BOSS:} {A} benchmark for human belief prediction in object-context scenarios.
\newblock \emph{CoRR}, abs/2206.10665, 2022.
\newblock \doi{10.48550/ARXIV.2206.10665}.
\newblock URL \url{https://doi.org/10.48550/arXiv.2206.10665}.

\bibitem[Fan et~al.(2021)Fan, Qiu, Zheng, Gao, Zhu, and Zhu]{fan2021learning}
Lifeng Fan, Shuwen Qiu, Zilong Zheng, Tao Gao, Song{-}Chun Zhu, and Yixin Zhu.
\newblock Learning triadic belief dynamics in nonverbal communication from videos.
\newblock In \emph{{IEEE} Conference on Computer Vision and Pattern Recognition, {CVPR} 2021, virtual, June 19-25, 2021}, pp.\  7312--7321. Computer Vision Foundation / {IEEE}, 2021.
\newblock \doi{10.1109/CVPR46437.2021.00723}.

\bibitem[Fernandez et~al.(2024)Fernandez, Longin, Lorini, and Maris]{fernandez2023logical}
Jorge Fernandez, Dominique Longin, Emiliano Lorini, and Fr{\'{e}}d{\'{e}}ric Maris.
\newblock A logical modeling of the {Y}{\={o}}kai board game.
\newblock \emph{{AI} Commun.}, 37\penalty0 (3):\penalty0 265--298, 2024.
\newblock \doi{10.3233/AIC-230050}.
\newblock URL \url{https://doi.org/10.3233/AIC-230050}.

\bibitem[Forkel et~al.(2026)Forkel, Ruhdorfer, Beukman, Bulling, and Foerster]{forkel2026highentropyleadssymmetry}
Johannes Forkel, Constantin Ruhdorfer, Michael Beukman, Andreas Bulling, and Jakob Foerster.
\newblock High entropy leads to symmetry equivariant policies in {Dec-POMDP}s.
\newblock \emph{CoRR}, 2026.
\newblock \doi{10.48550/arXiv.2511.22581}.
\newblock URL \url{https://arxiv.org/abs/2511.22581}.

\bibitem[Fuchs et~al.(2021)Fuchs, Walton, Chadwick, and Lange]{Fuchs2021Theory}
Andrew Fuchs, Michael Walton, Theresa Chadwick, and Doug Lange.
\newblock Theory of mind for deep reinforcement learning in {Hanabi}.
\newblock \emph{CoRR}, abs/2101.09328, 2021.
\newblock URL \url{https://arxiv.org/abs/2101.09328}.

\bibitem[Fukushima(1975)]{Fukushima1975CognitronAS}
Kunihiko Fukushima.
\newblock Cognitron: A self-organizing multilayered neural network.
\newblock \emph{Biological Cybernetics}, 20:\penalty0 121--136, 1975.

\bibitem[Gandhi et~al.(2023)Gandhi, Fr{\"{a}}nken, Gerstenberg, and Goodman]{gandhi2024understanding}
Kanishk Gandhi, Jan{-}Philipp Fr{\"{a}}nken, Tobias Gerstenberg, and Noah~D. Goodman.
\newblock Understanding social reasoning in language models with language models.
\newblock In Alice Oh, Tristan Naumann, Amir Globerson, Kate Saenko, Moritz Hardt, and Sergey Levine (eds.), \emph{Advances in Neural Information Processing Systems 36: Annual Conference on Neural Information Processing Systems 2023, NeurIPS 2023, New Orleans, LA, USA, December 10 - 16, 2023}, 2023.

\bibitem[Gessler et~al.(2025)Gessler, Dizdarevic, Calinescu, Ellis, Lupu, and Foerster]{Gessler2025OvercookedV2}
Tobias Gessler, Tin Dizdarevic, Ani Calinescu, Benjamin Ellis, Andrei Lupu, and Jakob~Nicolaus Foerster.
\newblock {OvercookedV2}: Rethinking {Overcooked} for zero-shot coordination.
\newblock In \emph{The Thirteenth International Conference on Learning Representations, {ICLR} 2025, Singapore, April 24-28, 2025}. OpenReview.net, 2025.
\newblock URL \url{https://openreview.net/forum?id=hlvLM3GX8R}.

\bibitem[Hamon(2024)]{Hamon2024TransformerXLJaxGithub}
Gautier Hamon.
\newblock {transformerXL\_PPO\_JAX}, July 2024.
\newblock URL \url{https://github.com/Reytuag/transformerXL_PPO_JAX}.

\bibitem[Harris et~al.(2020)Harris, Millman, van~der Walt, Gommers, Virtanen, Cournapeau, Wieser, Taylor, Berg, Smith, Kern, Picus, Hoyer, van Kerkwijk, Brett, Haldane, del R{\'{i}}o, Wiebe, Peterson, G{\'{e}}rard-Marchant, Sheppard, Reddy, Weckesser, Abbasi, Gohlke, and Oliphant]{harris2020array}
Charles~R. Harris, K.~Jarrod Millman, St{\'{e}}fan~J. van~der Walt, Ralf Gommers, Pauli Virtanen, David Cournapeau, Eric Wieser, Julian Taylor, Sebastian Berg, Nathaniel~J. Smith, Robert Kern, Matti Picus, Stephan Hoyer, Marten~H. van Kerkwijk, Matthew Brett, Allan Haldane, Jaime~Fern{\'{a}}ndez del R{\'{i}}o, Mark Wiebe, Pearu Peterson, Pierre G{\'{e}}rard-Marchant, Kevin Sheppard, Tyler Reddy, Warren Weckesser, Hameer Abbasi, Christoph Gohlke, and Travis~E. Oliphant.
\newblock Array programming with {NumPy}.
\newblock \emph{Nature}, 585\penalty0 (7825):\penalty0 357--362, September 2020.
\newblock \doi{10.1038/s41586-020-2649-2}.
\newblock URL \url{https://doi.org/10.1038/s41586-020-2649-2}.

\bibitem[Heek et~al.(2023)Heek, Levskaya, Oliver, Ritter, Rondepierre, Steiner, and van {Z}ee]{flax2020github}
Jonathan Heek, Anselm Levskaya, Avital Oliver, Marvin Ritter, Bertrand Rondepierre, Andreas Steiner, and Marc van {Z}ee.
\newblock {F}lax: A neural network library and ecosystem for {JAX}, 2023.
\newblock URL \url{http://github.com/google/flax}.

\bibitem[Hochreiter \& Schmidhuber(1997)Hochreiter and Schmidhuber]{Hochreiter1997Long}
Sepp Hochreiter and J{\"{u}}rgen Schmidhuber.
\newblock Long short-term memory.
\newblock \emph{Neural Comput.}, 9\penalty0 (8):\penalty0 1735--1780, 1997.
\newblock \doi{10.1162/NECO.1997.9.8.1735}.

\bibitem[Hu et~al.(2020)Hu, Lerer, Peysakhovich, and Foerster]{Hu2020Other}
Hengyuan Hu, Adam Lerer, Alex Peysakhovich, and Jakob~N. Foerster.
\newblock "{Other-Play}" for zero-shot coordination.
\newblock In \emph{Proceedings of the 37th International Conference on Machine Learning, {ICML} 2020, 13-18 July 2020, Virtual Event}, volume 119 of \emph{Proceedings of Machine Learning Research}, pp.\  4399--4410. {PMLR}, 2020.

\bibitem[Hu et~al.(2021)Hu, Lerer, Cui, Pineda, Brown, and Foerster]{Hu2021OffBelief}
Hengyuan Hu, Adam Lerer, Brandon Cui, Luis Pineda, Noam Brown, and Jakob Foerster.
\newblock Off-belief learning.
\newblock In Marina Meila and Tong Zhang (eds.), \emph{Proceedings of the 38th International Conference on Machine Learning}, volume 139 of \emph{Proceedings of Machine Learning Research}, pp.\  4369--4379. PMLR, 18--24 Jul 2021.
\newblock URL \url{https://proceedings.mlr.press/v139/hu21c.html}.

\bibitem[Hui et~al.(2026)Hui, Yu, Yao, Qu, Zhang, and Wang]{Hui2025Efficient}
Bingyu Hui, Lebin Yu, Quanming Yao, Yunpeng Qu, Xudong Zhang, and Jian Wang.
\newblock Efficient reinforcement learning for zero-shot coordination in evolving games.
\newblock In Sven Koenig, Chad Jenkins, and Matthew~E. Taylor (eds.), \emph{Fortieth {AAAI} Conference on Artificial Intelligence, Thirty-Eighth Conference on Innovative Applications of Artificial Intelligence, Sixteenth Symposium on Educational Advances in Artificial Intelligence, {AAAI} 2026, Singapore, January 20-27, 2026}, pp.\  22110--22118. {AAAI} Press, 2026.
\newblock \doi{10.1609/AAAI.V40I26.39366}.

\bibitem[Hunter(2007)]{Hunter2007matplotlib}
J.~D. Hunter.
\newblock Matplotlib: A 2d graphics environment.
\newblock \emph{Computing in Science \& Engineering}, 9\penalty0 (3):\penalty0 90--95, 2007.
\newblock \doi{10.1109/MCSE.2007.55}.

\bibitem[Kim et~al.(2023)Kim, Sclar, Zhou, Bras, Kim, Choi, and Sap]{kim2023fantom}
Hyunwoo Kim, Melanie Sclar, Xuhui Zhou, Ronan~Le Bras, Gunhee Kim, Yejin Choi, and Maarten Sap.
\newblock {FANToM}: {A} benchmark for stress-testing machine theory of mind in interactions.
\newblock In Houda Bouamor, Juan Pino, and Kalika Bali (eds.), \emph{Proceedings of the 2023 Conference on Empirical Methods in Natural Language Processing, {EMNLP} 2023, Singapore, December 6-10, 2023}, pp.\  14397--14413. Association for Computational Linguistics, 2023.
\newblock \doi{10.18653/V1/2023.EMNLP-MAIN.890}.

\bibitem[Kingma \& Ba(2015)Kingma and Ba]{kingma2015adam}
Diederik~P. Kingma and Jimmy Ba.
\newblock Adam: {A} method for stochastic optimization.
\newblock In Yoshua Bengio and Yann LeCun (eds.), \emph{3rd International Conference on Learning Representations, {ICLR} 2015, San Diego, CA, USA, May 7-9, 2015, Conference Track Proceedings}, 2015.
\newblock URL \url{http://arxiv.org/abs/1412.6980}.

\bibitem[Kipf \& Welling(2017)Kipf and Welling]{kipf2016semi}
Thomas~N. Kipf and Max Welling.
\newblock Semi-supervised classification with graph convolutional networks.
\newblock In \emph{5th International Conference on Learning Representations, {ICLR} 2017, Toulon, France, April 24-26, 2017, Conference Track Proceedings}. OpenReview.net, 2017.

\bibitem[Klein et~al.(2004)Klein, Woods, Bradshaw, Hoffman, and Feltovich]{Klien2004Ten}
Gary Klein, David~D. Woods, Jeffrey~M. Bradshaw, Robert~R. Hoffman, and Paul~J. Feltovich.
\newblock Ten challenges for making automation a "team player" in joint human-agent activity.
\newblock \emph{{IEEE} Intell. Syst.}, 19\penalty0 (6):\penalty0 91--95, 2004.
\newblock \doi{10.1109/MIS.2004.74}.
\newblock URL \url{https://doi.org/10.1109/MIS.2004.74}.

\bibitem[Lauffer et~al.(2025)Lauffer, Shah, Carroll, Seshia, Russell, and Dennis]{Lauffer2025Robust}
Niklas Lauffer, Ameesh Shah, Micah Carroll, Sanjit~A. Seshia, Stuart~J. Russell, and Michael Dennis.
\newblock Robust and diverse multi-agent learning via rational policy gradient.
\newblock In Danielle Belgrave, Cheng Zhang, Laura~N. Montoya, Hsuan{-}Tien Lin, Razvan Pascanu, Piotr Koniusz, Marzyeh Ghassemi, Nancy Chen, Iv{\'{a}}n Vladimir~Meza Ru{\'{\i}}z, and Arturo Loaiza{-}Bonilla (eds.), \emph{Advances in Neural Information Processing Systems 38: Annual Conference on Neural Information Processing Systems 2025, NeurIPS 2025, San Diego, CA, USA, December 2-7, 2025 / Mexico City, Mexico, November 30 - December 5, 2025}, 2025.

\bibitem[Lucas \& Allen(2022)Lucas and Allen]{Keane2022AnyPlay}
Keane Lucas and Ross~E. Allen.
\newblock Any-play: An intrinsic augmentation for zero-shot coordination.
\newblock In Piotr Faliszewski, Viviana Mascardi, Catherine Pelachaud, and Matthew~E. Taylor (eds.), \emph{21st International Conference on Autonomous Agents and Multiagent Systems, {AAMAS} 2022, Auckland, New Zealand, May 9-13, 2022}, pp.\  853--861. International Foundation for Autonomous Agents and Multiagent Systems {(IFAAMAS)}, 2022.
\newblock \doi{10.5555/3535850.3535946}.

\bibitem[Matiisen et~al.(2022)Matiisen, Labash, Majoral, Aru, and Vicente]{Matiisen2022Do}
Tambet Matiisen, Aqeel Labash, Daniel Majoral, Jaan Aru, and Raul Vicente.
\newblock Do deep reinforcement learning agents model intentions?
\newblock \emph{Stats}, 6\penalty0 (1):\penalty0 50–66, December 2022.
\newblock ISSN 2571-905X.
\newblock \doi{10.3390/stats6010004}.
\newblock URL \url{http://dx.doi.org/10.3390/stats6010004}.

\bibitem[Matthews et~al.(2024)Matthews, Beukman, Ellis, Samvelyan, Jackson, Coward, and Foerster]{matthews2024craftax}
Michael~T. Matthews, Michael Beukman, Benjamin Ellis, Mikayel Samvelyan, Matthew~Thomas Jackson, Samuel Coward, and Jakob~Nicolaus Foerster.
\newblock {Craftax}: {A} lightning-fast benchmark for open-ended reinforcement learning.
\newblock In Ruslan Salakhutdinov, Zico Kolter, Katherine~A. Heller, Adrian Weller, Nuria Oliver, Jonathan Scarlett, and Felix Berkenkamp (eds.), \emph{Forty-first International Conference on Machine Learning, {ICML} 2024, Vienna, Austria, July 21-27, 2024}, volume 235 of \emph{Proceedings of Machine Learning Research}, pp.\  35104--35137. {PMLR} / OpenReview.net, 2024.

\bibitem[Muglich et~al.(2022)Muglich, de~Witt, van~der Pol, Whiteson, and Foerster]{Muglich2022Equivariant}
Darius Muglich, Christian~Schr{\"{o}}der de~Witt, Elise van~der Pol, Shimon Whiteson, and Jakob~N. Foerster.
\newblock Equivariant networks for zero-shot coordination.
\newblock In Sanmi Koyejo, S.~Mohamed, A.~Agarwal, Danielle Belgrave, K.~Cho, and A.~Oh (eds.), \emph{Advances in Neural Information Processing Systems 35: Annual Conference on Neural Information Processing Systems 2022, NeurIPS 2022, New Orleans, LA, USA, November 28 - December 9, 2022}, 2022.

\bibitem[Muglich et~al.(2025)Muglich, Forkel, van~der Pol, and Foerster]{Muglich2025Expected}
Darius Muglich, Johannes Forkel, Elise van~der Pol, and Jakob~Nicolaus Foerster.
\newblock Expected return symmetries.
\newblock In \emph{The Thirteenth International Conference on Learning Representations, {ICLR} 2025, Singapore, April 24-28, 2025}. OpenReview.net, 2025.
\newblock URL \url{https://openreview.net/forum?id=wFg0shwoRe}.

\bibitem[Nair \& Hinton(2010)Nair and Hinton]{Nair2010Rectified}
Vinod Nair and Geoffrey~E. Hinton.
\newblock Rectified linear units improve restricted boltzmann machines.
\newblock In Johannes F{\"{u}}rnkranz and Thorsten Joachims (eds.), \emph{Proceedings of the 27th International Conference on Machine Learning (ICML-10), June 21-24, 2010, Haifa, Israel}, pp.\  807--814. Omnipress, 2010.
\newblock URL \url{https://icml.cc/Conferences/2010/papers/432.pdf}.

\bibitem[Nikulin et~al.(2024)Nikulin, Kurenkov, Zisman, Agarkov, Sinii, and Kolesnikov]{nikulin2023xlandminigrid}
Alexander Nikulin, Vladislav Kurenkov, Ilya Zisman, Artem Agarkov, Viacheslav Sinii, and Sergey Kolesnikov.
\newblock {XLand-MiniGrid}: Scalable meta-reinforcement learning environments in {JAX}.
\newblock In Amir Globersons, Lester Mackey, Danielle Belgrave, Angela Fan, Ulrich Paquet, Jakub~M. Tomczak, and Cheng Zhang (eds.), \emph{Advances in Neural Information Processing Systems 37: Annual Conference on Neural Information Processing Systems 2024, NeurIPS 2024, Vancouver, BC, Canada, December 10 - 15, 2024}, 2024.

\bibitem[Oliehoek \& Amato(2016)Oliehoek and Amato]{Oliehoek2016}
Frans~A. Oliehoek and Christopher Amato.
\newblock \emph{A Concise Introduction to Decentralized {POMDPs}}.
\newblock Springer Briefs in Intelligent Systems. Springer, 2016.
\newblock ISBN 978-3-319-28927-4.
\newblock \doi{10.1007/978-3-319-28929-8}.
\newblock URL \url{https://doi.org/10.1007/978-3-319-28929-8}.

\bibitem[Parisotto et~al.(2020)Parisotto, Song, Rae, Pascanu, G{\"{u}}l{\c{c}}ehre, Jayakumar, Jaderberg, Kaufman, Clark, Noury, Botvinick, Heess, and Hadsell]{Parisotto2020Stabilizing}
Emilio Parisotto, H.~Francis Song, Jack~W. Rae, Razvan Pascanu, {\c{C}}aglar G{\"{u}}l{\c{c}}ehre, Siddhant~M. Jayakumar, Max Jaderberg, Rapha{\"{e}}l~Lopez Kaufman, Aidan Clark, Seb Noury, Matthew~M. Botvinick, Nicolas Heess, and Raia Hadsell.
\newblock Stabilizing transformers for reinforcement learning.
\newblock In \emph{Proceedings of the 37th International Conference on Machine Learning, {ICML} 2020, 13-18 July 2020, Virtual Event}, volume 119 of \emph{Proceedings of Machine Learning Research}, pp.\  7487--7498. {PMLR}, 2020.

\bibitem[Premack \& Woodruff(1978)Premack and Woodruff]{premack1978does}
David Premack and Guy Woodruff.
\newblock Does the chimpanzee have a theory of mind?
\newblock \emph{Behavioral and brain sciences}, 1\penalty0 (4):\penalty0 515--526, 1978.

\bibitem[Rabinowitz et~al.(2018)Rabinowitz, Perbet, Song, Zhang, Eslami, and Botvinick]{rabinowitz2018machine}
Neil~C. Rabinowitz, Frank Perbet, H.~Francis Song, Chiyuan Zhang, S.~M.~Ali Eslami, and Matthew~M. Botvinick.
\newblock Machine theory of mind.
\newblock In Jennifer~G. Dy and Andreas Krause (eds.), \emph{Proceedings of the 35th International Conference on Machine Learning, {ICML} 2018, Stockholmsm{\"{a}}ssan, Stockholm, Sweden, July 10-15, 2018}, volume~80 of \emph{Proceedings of Machine Learning Research}, pp.\  4215--4224. {PMLR}, 2018.
\newblock URL \url{http://proceedings.mlr.press/v80/rabinowitz18a.html}.

\bibitem[Ruhdorfer et~al.(2025{\natexlab{a}})Ruhdorfer, Bortoletto, Oei, Penzkofer, and Bulling]{Ruhdorfer2025Unsupervised}
Constantin Ruhdorfer, Matteo Bortoletto, Victor Oei, Anna Penzkofer, and Andreas Bulling.
\newblock Unsupervised partner design enables robust ad-hoc teamwork.
\newblock \emph{CoRR}, abs/2508.06336, 2025{\natexlab{a}}.
\newblock \doi{10.48550/ARXIV.2508.06336}.
\newblock URL \url{https://doi.org/10.48550/arXiv.2508.06336}.

\bibitem[Ruhdorfer et~al.(2025{\natexlab{b}})Ruhdorfer, Bortoletto, Penzkofer, and Bulling]{ruhdorfer2025the}
Constantin Ruhdorfer, Matteo Bortoletto, Anna Penzkofer, and Andreas Bulling.
\newblock The {Overcooked} generalisation challenge: Evaluating cooperation with novel partners in unknown environments using environment design.
\newblock \emph{Trans. Mach. Learn. Res.}, 2025, 2025{\natexlab{b}}.
\newblock URL \url{https://openreview.net/forum?id=K2KtcMlW6j}.

\bibitem[Rutherford et~al.(2024)Rutherford, Ellis, Gallici, Cook, Lupu, Ingvarsson, Willi, Hammond, Khan, de~Witt, Souly, Bandyopadhyay, Samvelyan, Jiang, Lange, Whiteson, Lacerda, Hawes, Rockt{\"{a}}schel, Lu, and Foerster]{rutherford2023jaxmarl}
Alexander Rutherford, Benjamin Ellis, Matteo Gallici, Jonathan Cook, Andrei Lupu, Gar{\dh}ar Ingvarsson, Timon Willi, Ravi Hammond, Akbir Khan, Christian~Schr{\"{o}}der de~Witt, Alexandra Souly, Saptarashmi Bandyopadhyay, Mikayel Samvelyan, Minqi Jiang, Robert~T. Lange, Shimon Whiteson, Bruno Lacerda, Nick Hawes, Tim Rockt{\"{a}}schel, Chris Lu, and Jakob~N. Foerster.
\newblock {JaxMARL}: Multi-agent {RL} environments and algorithms in {JAX}.
\newblock In Amir Globersons, Lester Mackey, Danielle Belgrave, Angela Fan, Ulrich Paquet, Jakub~M. Tomczak, and Cheng Zhang (eds.), \emph{Advances in Neural Information Processing Systems 37: Annual Conference on Neural Information Processing Systems 2024, NeurIPS 2024, Vancouver, BC, Canada, December 10 - 15, 2024}, 2024.

\bibitem[Samvelyan et~al.(2019)Samvelyan, Rashid, de~Witt, Farquhar, Nardelli, Rudner, Hung, Torr, Foerster, and Whiteson]{samvelyan2019starcraft}
Mikayel Samvelyan, Tabish Rashid, Christian~Schr{\"{o}}der de~Witt, Gregory Farquhar, Nantas Nardelli, Tim G.~J. Rudner, Chia{-}Man Hung, Philip H.~S. Torr, Jakob~N. Foerster, and Shimon Whiteson.
\newblock The {StarCraft} multi-agent challenge.
\newblock In Edith Elkind, Manuela Veloso, Noa Agmon, and Matthew~E. Taylor (eds.), \emph{Proceedings of the 18th International Conference on Autonomous Agents and MultiAgent Systems, {AAMAS} '19, Montreal, QC, Canada, May 13-17, 2019}, pp.\  2186--2188. International Foundation for Autonomous Agents and Multiagent Systems, 2019.

\bibitem[Sclar et~al.(2022)Sclar, Neubig, and Bisk]{Sclar2022Symmetric}
Melanie Sclar, Graham Neubig, and Yonatan Bisk.
\newblock Symmetric machine theory of mind.
\newblock In Kamalika Chaudhuri, Stefanie Jegelka, Le~Song, Csaba Szepesvari, Gang Niu, and Sivan Sabato (eds.), \emph{Proceedings of the 39th International Conference on Machine Learning}, volume 162 of \emph{Proceedings of Machine Learning Research}, pp.\  19450--19466. PMLR, 17--23 Jul 2022.
\newblock URL \url{https://proceedings.mlr.press/v162/sclar22a.html}.

\bibitem[Stone et~al.(2010)Stone, Kaminka, Kraus, and Rosenschein]{stone2010ad}
Peter Stone, Gal~A. Kaminka, Sarit Kraus, and Jeffrey~S. Rosenschein.
\newblock Ad hoc autonomous agent teams: Collaboration without pre-coordination.
\newblock In Maria Fox and David Poole (eds.), \emph{Proceedings of the Twenty-Fourth {AAAI} Conference on Artificial Intelligence, {AAAI} 2010, Atlanta, Georgia, USA, July 11-15, 2010}, pp.\  1504--1509. {AAAI} Press, 2010.
\newblock \doi{10.1609/AAAI.V24I1.7529}.

\bibitem[Strouse et~al.(2021)Strouse, McKee, Botvinick, Hughes, and Everett]{Strouse2021Collaborating}
DJ~Strouse, Kevin~R. McKee, Matt~M. Botvinick, Edward Hughes, and Richard Everett.
\newblock Collaborating with humans without human data.
\newblock In Marc'Aurelio Ranzato, Alina Beygelzimer, Yann~N. Dauphin, Percy Liang, and Jennifer~Wortman Vaughan (eds.), \emph{Advances in Neural Information Processing Systems 34: Annual Conference on Neural Information Processing Systems 2021, NeurIPS 2021, December 6-14, 2021, virtual}, pp.\  14502--14515, 2021.

\bibitem[Sudhakar et~al.(2025)Sudhakar, Nekoei, Reymond, Liu, Rajendran, and Chandar]{sudhakar2025generalist}
Arjun~Vaithilingam Sudhakar, Hadi Nekoei, Mathieu Reymond, Miao Liu, Janarthanan Rajendran, and Sarath Chandar.
\newblock A generalist {Hanabi} agent.
\newblock In \emph{The Thirteenth International Conference on Learning Representations, {ICLR} 2025, Singapore, April 24-28, 2025}. OpenReview.net, 2025.
\newblock URL \url{https://openreview.net/forum?id=pCj2sLNoJq}.

\bibitem[Team(2020)]{reback2020pandas}
The Pandas~Development Team.
\newblock pandas-dev/pandas: Pandas.
\newblock February 2020.
\newblock \doi{10.5281/zenodo.3509134}.
\newblock URL \url{https://doi.org/10.5281/zenodo.3509134}.

\bibitem[Tomilin et~al.(2026)Tomilin, van~den Boogaard, Garcin, Ruhdorfer, Grooten, Kusters, Du, Bulling, Pechenizkiy, and Fang]{tomilin2025mealbenchmarkcontinualmultiagent}
Tristan Tomilin, Luka van~den Boogaard, Samuel Garcin, Constantin Ruhdorfer, Bram Grooten, Fabrice Kusters, Yali Du, Andreas Bulling, Mykola Pechenizkiy, and Meng Fang.
\newblock {MEAL}: A benchmark for continual multi-agent reinforcement learning, 2026.
\newblock URL \url{https://arxiv.org/abs/2506.14990}.

\bibitem[Treutlein et~al.(2021)Treutlein, Dennis, Oesterheld, and Foerster]{treutlein2021new}
Johannes Treutlein, Michael Dennis, Caspar Oesterheld, and Jakob~N. Foerster.
\newblock A new formalism, method and open issues for zero-shot coordination.
\newblock In Marina Meila and Tong Zhang (eds.), \emph{Proceedings of the 38th International Conference on Machine Learning, {ICML} 2021, 18-24 July 2021, Virtual Event}, volume 139 of \emph{Proceedings of Machine Learning Research}, pp.\  10413--10423. {PMLR}, 2021.

\bibitem[Virtanen et~al.(2020)Virtanen, Gommers, Oliphant, Haberland, Reddy, Cournapeau, Burovski, Peterson, Weckesser, Bright, {van der Walt}, Brett, Wilson, Millman, Mayorov, Nelson, Jones, Kern, Larson, Carey, Polat, Feng, Moore, {VanderPlas}, Laxalde, Perktold, Cimrman, Henriksen, Quintero, Harris, Archibald, Ribeiro, Pedregosa, {van Mulbregt}, and {SciPy 1.0 Contributors}]{virtanen2020scipy}
Pauli Virtanen, Ralf Gommers, Travis~E. Oliphant, Matt Haberland, Tyler Reddy, David Cournapeau, Evgeni Burovski, Pearu Peterson, Warren Weckesser, Jonathan Bright, St{\'e}fan~J. {van der Walt}, Matthew Brett, Joshua Wilson, K.~Jarrod Millman, Nikolay Mayorov, Andrew R.~J. Nelson, Eric Jones, Robert Kern, Eric Larson, C~J Carey, {\.I}lhan Polat, Yu~Feng, Eric~W. Moore, Jake {VanderPlas}, Denis Laxalde, Josef Perktold, Robert Cimrman, Ian Henriksen, E.~A. Quintero, Charles~R. Harris, Anne~M. Archibald, Ant{\^o}nio~H. Ribeiro, Fabian Pedregosa, Paul {van Mulbregt}, and {SciPy 1.0 Contributors}.
\newblock {{SciPy} 1.0: Fundamental Algorithms for Scientific Computing in Python}.
\newblock \emph{Nature Methods}, 17:\penalty0 261--272, 2020.
\newblock \doi{10.1038/s41592-019-0686-2}.

\bibitem[Xue et~al.(2025)Xue, Wang, Guan, Yuan, Fu, Fu, Qian, and Yu]{Xue2022Heterogeneous}
Ke~Xue, Yutong Wang, Cong Guan, Lei Yuan, Haobo Fu, Qiang Fu, Chao Qian, and Yang Yu.
\newblock Heterogeneous multiagent zero-shot coordination by coevolution.
\newblock \emph{IEEE Transactions on Evolutionary Computation}, 29\penalty0 (5):\penalty0 2229--2243, 2025.
\newblock \doi{10.1109/TEVC.2024.3485177}.

\bibitem[Yan et~al.(2023)Yan, Guo, Lou, Wang, Zhang, and Du]{Yan2023An}
Xue Yan, Jiaxian Guo, Xingzhou Lou, Jun Wang, Haifeng Zhang, and Yali Du.
\newblock An efficient end-to-end training approach for zero-shot human-{AI} coordination.
\newblock In Alice Oh, Tristan Naumann, Amir Globerson, Kate Saenko, Moritz Hardt, and Sergey Levine (eds.), \emph{Advances in Neural Information Processing Systems 36: Annual Conference on Neural Information Processing Systems 2023, NeurIPS 2023, New Orleans, LA, USA, December 10 - 16, 2023}, 2023.

\bibitem[Yang(2024)]{yang2024yokai}
Rui Yang.
\newblock The {Yokai} challenge: A new frontier for multi-agent reinforcement learning and machine theory of mind.
\newblock Master’s thesis, University of Stuttgart, 2024.

\bibitem[Ying et~al.(2021)Ying, Cai, Luo, Zheng, Ke, He, Shen, and Liu]{ying2021transformers}
Chengxuan Ying, Tianle Cai, Shengjie Luo, Shuxin Zheng, Guolin Ke, Di~He, Yanming Shen, and Tie{-}Yan Liu.
\newblock Do transformers really perform badly for graph representation?
\newblock In Marc'Aurelio Ranzato, Alina Beygelzimer, Yann~N. Dauphin, Percy Liang, and Jennifer~Wortman Vaughan (eds.), \emph{Advances in Neural Information Processing Systems 34: Annual Conference on Neural Information Processing Systems 2021, NeurIPS 2021, December 6-14, 2021, virtual}, pp.\  28877--28888, 2021.

\bibitem[Zambaldi et~al.(2019)Zambaldi, Raposo, Santoro, Bapst, Li, Babuschkin, Tuyls, Reichert, Lillicrap, Lockhart, Shanahan, Langston, Pascanu, Botvinick, Vinyals, and Battaglia]{Zambaldi2018Relational}
Vin{\'{\i}}cius~Flores Zambaldi, David Raposo, Adam Santoro, Victor Bapst, Yujia Li, Igor Babuschkin, Karl Tuyls, David~P. Reichert, Timothy~P. Lillicrap, Edward Lockhart, Murray Shanahan, Victoria Langston, Razvan Pascanu, Matthew~M. Botvinick, Oriol Vinyals, and Peter~W. Battaglia.
\newblock Deep reinforcement learning with relational inductive biases.
\newblock In \emph{7th International Conference on Learning Representations, {ICLR} 2019, New Orleans, LA, USA, May 6-9, 2019}. OpenReview.net, 2019.
\newblock URL \url{https://openreview.net/forum?id=HkxaFoC9KQ}.

\bibitem[Zhao et~al.(2023)Zhao, Song, Yuan, Hu, Gao, Wu, Sun, and Yang]{Zhao2023Maximum}
Rui Zhao, Jinming Song, Yufeng Yuan, Haifeng Hu, Yang Gao, Yi~Wu, Zhongqian Sun, and Wei Yang.
\newblock Maximum entropy population-based training for zero-shot human-{AI} coordination.
\newblock In Brian Williams, Yiling Chen, and Jennifer Neville (eds.), \emph{Thirty-Seventh {AAAI} Conference on Artificial Intelligence, {AAAI} 2023, Thirty-Fifth Conference on Innovative Applications of Artificial Intelligence, {IAAI} 2023, Thirteenth Symposium on Educational Advances in Artificial Intelligence, {EAAI} 2023, Washington, DC, USA, February 7-14, 2023}, pp.\  6145--6153. {AAAI} Press, 2023.
\newblock \doi{10.1609/AAAI.V37I5.25758}.

\bibitem[Zhu et~al.(2024)Zhu, Zhang, and Wang]{zhu2024language}
Wentao Zhu, Zhining Zhang, and Yizhou Wang.
\newblock Language models represent beliefs of self and others.
\newblock In Ruslan Salakhutdinov, Zico Kolter, Katherine~A. Heller, Adrian Weller, Nuria Oliver, Jonathan Scarlett, and Felix Berkenkamp (eds.), \emph{Forty-first International Conference on Machine Learning, {ICML} 2024, Vienna, Austria, July 21-27, 2024}, volume 235 of \emph{Proceedings of Machine Learning Research}, pp.\  62638--62681. {PMLR} / OpenReview.net, 2024.

\end{thebibliography}
\bibliographystyle{rlj}

\beginSupplementaryMaterials

\section{Servers and Software}
\label{sec:infrastrcuture}
We ran our experiments on server systems running Ubuntu 24.04 equipped with NVIDIA Tesla V100-SXM2 GPUs with 32GB of memory, NVIDIA H100-NVL GPUs with 94GB of memory and a third server system featuring NVIDIA L40S with 48 GB of memory.
All training runs are executed on a single GPU only.
We trained our models using JAX~\citep{jax2018github}, Flax~\citep{flax2020github} and Optax~\citep{deepmind2020jax}.
In terms of tools, we did our analysis using NumPy~\citep{harris2020array}, Pandas~\citep{reback2020pandas}, SciPy~\citep{virtanen2020scipy} and Matplotlib~\citep{Hunter2007matplotlib}.
Our TransformerXL implementation is based on the one provided by \citet{Hamon2024TransformerXLJaxGithub}.
Our single-file implementation of IPPO is based on the one provided by JaxMARL \citep{rutherford2023jaxmarl} with additional inspiration taken from \citet{Hamon2024TransformerXLJaxGithub}.

\section{Accessibility and reusability}
\label{sec:accesability}
Our version of the \yokai{} Learning Environment (YLE) integrates directly into JaxMARL and implements its standard environment interface. 
As a result, any baseline method implementation of JaxMARL can be adapted to our environment quickly and with minimal effort.
This makes our open-source environment easily accessible to future researchers who will benefit from our reference implementations and those provided by the wider community.
It also enables inter-environment comparison, facilitating broader benchmarking across cooperative multi-agent tasks. 
The project code to reproduce our results is accessible under \url{https://git.hcics.simtech.uni-stuttgart.de/public-projects/Yokai_env}.

\section{Fast, Scalable, and Parallel \yokai{} in Graph-based JAX}
\label{sec:envDetails}
We first describe in detail how our environment operates and then share additional details on reinforcement learning in the YLE.

\subsection{JAX-based Implementation}
\yokai{} was first explored as a reinforcement learning environment in a preliminary study \citep{yang2024yokai}, which implemented an initial JAX-based environment for a simplified version of the game without early ending.
They showed that training self-play RL agents in this setting is feasible.
Informed by this work, we introduce the YLE as a benchmark for ZSC, providing complete game mechanics including early ending, configurable game sizes, new coordination metrics based on early termination, and support for evaluating and comparing modern ZSC algorithms.
We retain the JAX-native approach and design the environment around graph-based state representations, enabling efficient parallel simulation of the complete game.
Because early ending makes learning substantially more challenging, our environment interface also differs in several respects to support reliable learning.
For example, move actions specify absolute grid locations rather than positions relative to other cards, giving each action a consistent spatial meaning across states.
Implemented in JAX, the environment executes end-to-end on GPUs and is fully compatible with just-in-time compilation.
This design enables both simulation and learning to occur entirely on the GPU, avoiding costly GPU–CPU communication and significantly improving efficiency.
As a result, the environment supports training at hundreds of thousands of steps per second on modern hardware, enabling fast experimentation with relatively modest compute resources (see \autoref{tab:steps_per_second}).

This efficiency and scalability stem primarily from how we represent and update the environment’s state using adjacency matrices $A_t$ at each timestep $s_t$.
In the following, we describe two practical examples of how this is achieved.

\paragraph{Efficiently computing legal moves}
We always mask illegal actions and thus need to compute the set of legal actions at every step.
While identifying observable cards and placable hint cards is computationally inexpensive, computing legal next moves is by far the most costly computation the environment performs.

To determine which card moves are legal in a given state $s_t$, the environment must simulate each possible action $a \in A$ and assess whether the resulting state $s_{t+1}$ is valid.
A state is considered legal if (1) all cards are connected into a single, fully-connected graph via their neighbours, and (2) no cards overlap spatially.

To make this efficient, we compute all potential next states in parallel:
\begin{equation}
    S_{t+1} = \{s_{t+1} | s_{t+1} = \mathcal{T}(s_t, a) \forall a \in \mathcal{A} \}.
\end{equation}
Each state $s_{t+1}$ is associated with an adjacency matrix $A_t$ that reflects the card layout after a move. 
For each candidate action (i.e., moving a card $y \in Y$ to a new grid position $(x_1, x_2)$), we first verify that the position is within bounds and the card is movable.

We then check whether each resulting state is legal by evaluating its adjacency matrix. 
Specifically, we compute the reachability matrix:
\begin{equation}
    R = (I + A)^{|Y|-1}
\end{equation}
The state is legal if all entries in $R$ are positive, i.e., $R[i][j] > 0$ for all $i, j$. 
This test, implemented as a parallel matrix power operation, allows us to efficiently filter legal moves at each step.

While designing environments for JAX requires careful consideration, these efforts potentially yield significant speedups that make RL research accessible to a broad range of researchers.

\paragraph{Efficiently executing moves}
In our graph-based setup, executing environment transitions is highly efficient.
Observing a card simply involves unmasking its colour in the node feature matrix $F_t$.
To execute a move, we update the adjacency matrix $A_t$ to reflect the new spatial configuration.
Specifically, when a card with index $i \in [0, |Y|)$ is moved, we first remove its current edges by setting:
\begin{equation}
    A_t[j,i] = 0 \text{ and } A_t[i,j] = 0 \;\;\forall j\in [0, |Y|)
\end{equation}
Next, we compute a binary vector $v$ such that $v_k = 1$ if card $k$ is a spatial neighbour of the new position of card $i$, and $0$ otherwise.
We then update the adjacency matrix by setting:
\begin{equation}
    A_{:,i} = v \text{ and } A_{i,:} = v.
\end{equation}

\subsection{Reinforcement Learning in the YLE}
\label{sec:RLinYLE}
We continue discussing additional aspects of reinforcement learning in the YLE.

\paragraph{Playing \yokai{} on a Grid}
\begin{figure}[htp]
    \centering
    \includegraphics[width=\linewidth]{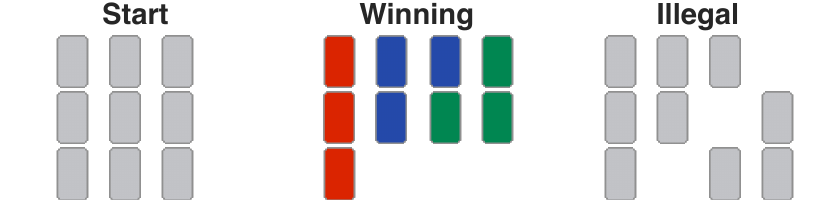}
    \caption{Examples of start, winning, and illegal card configurations for a \yokai{} game with nine cards and three colours.}
    \label{fig:startAndWin}
\end{figure}
While in the original game, cards can be placed in any legal configuration, our environment limits the configuration to be in a predefined field of size $g \times g$ where $g = 9$ in the 9C and $g = 10$ in the 16C version. 
This enables the efficient computation of legal moves but prohibits some rare card configurations as the grid space is smaller than required for some specific sequences of moves.
While computing the exact number of disallowed configurations is infeasible, we found them to be extremely uncommon in practice.

For example, let's assume a 2-player 9C game on a 9x9 grid.
In such a setting, there are 4 hints and thus a total of 8 move actions for both players together.
Cards are initially placed in a square in the centre of the board, leaving 3 empty spaces in the grid towards either side.
In theory, thus, if agents agree to start placing cards in a row towards one of these sides, they would run out of space towards the border after 3 moves.
In practice, this is seldom a suitable strategy and thus does not usually occur in games of \yokai{}.
This is because players want to group cards and are limited by the number of moves.
They thus do not tend to start grouping cards in random directions, but rather group cards where they initially find a certain colour.

To verify that this is also true for our agents, we count the number of times when, in an evaluation game, an agent places a card on a border position.
Note this is an overestimation as it does not account for the cases where agents place a card on the border but do not intend to place further cards in the same direction.
We found that policies rarely, if ever, used the border region of the board.

\paragraph{The action space in detail}
Inspired by \citet{yang2024yokai}, to observe two cards, agents first pick from $|Y|$ and then from $N_{cards} - 1$ cards.
Next, an agent moves a card by picking one of $N_{moves}$ actions, which can be understood to take card $0 \leq i < |Y|$ and place it at $(x,y) \in g \times g$.
Finally, the active agent reveals a hint from the hint pile or places a hint onto a card by picking from $|H| + |H| \cdot |Y|$ actions.
Consequently, the maximum episode length is $8|H|$ while the minimum is $1$.
The number of actions is therefore
\begin{equation}
    N_a = 1 + |Y| + N_{moves} + |H| + N_{hint\_moves} + 1.
\end{equation}
For example, the action space is of size $1,068$ for 2-player \textbf{9C} games and $2,322$ for the \textbf{16C} version.

\paragraph{YLE is a symmetric ToM environment}
\cite{Sclar2022Symmetric} defines symmetric ToM environments as environments that (1) have a symmetric action space, (2) feature imperfect information, (3) allow the observation of others, and (4) require information-seeking behaviour.
YLE features all four and thus qualifies, as also described in \citet{yang2024yokai}:
First, the action space is symmetric (see above).
Second, information is imperfect since agents can never observe all cards first-hand and, for most of the game, only know a very limited subset of all cards.
Third, it allows the observation of others both using explicitly encoded actions or by observing their moves indirectly via their observations.
Fourth, information needs to be acquired by observing cards first-hand or, second-hand, by interpreting moves and reasoning about knowledge.
Agents that do not seek information and, for instance, repeatedly observe the same card, will fail the game.

\paragraph{Number of Players, Hints and the Difficulty}
\begin{table}[t]
    \centering
    \caption{Number of hint cards according to the number of players and the environment version. The 9C version features three colours and thus has no three-colour hints.}
    \label{hintcards}
    \begin{tabular}{l r r r r r r}
        \toprule
                   & \multicolumn{3}{c}{\textbf{16C}} & \multicolumn{3}{c}{\textbf{9C}}\\
        \cmidrule(lr){2-4} \cmidrule(lr){5-7}
        \textbf{Number of Players} & \textbf{2} & \textbf{3} & \textbf{4} & \textbf{2} & \textbf{3} & \textbf{4} \\
        \midrule
        Number of 1-colour hint cards  & 2 & 2 & 3 & 1 & 2 & 3 \\
        Number of 2-colours hint cards & 3 & 4 & 4 & 3 & 3 & 3 \\
        Number of 3-colours hint cards & 2 & 3 & 3 & - & - & - \\
        \midrule
        Sum & 7 & 9 & 10 & 4 & 5 & 6 \\
        \bottomrule
    \end{tabular}
\end{table}
How many and which hint cards are available depend on the environment version and the number of players.
For a 2-player game, there are 2 hints of one, 3 hints of two and 2 hints of three colours, i.e. 7 in total.
Since there are 4 possible one-, 6 possible two- and 4 possible three-colour hints, players get a random subset of these hints at the start of the game per category.
We give a full overview of all possible variations in \autoref{hintcards}.
The number of hints directly influences the maximum number of turns in the game. 
This is because the game ends when all hints have been used.
This means that games take longer if more players are added.
Note, though, that this usually does not make the game easier.
Instead, the number of cards one player can directly observe themselves decreases with the number of players.
For 16C YLE in the 2-player version, any agent can observe up to 14 of the 16 cards themselves (8 of 9 in 9C \yokai{}). 
This shrinks to 10 of 16 for the 16C version and 6 of 9 for the 9C version when 4 players take part.
Note how in the 3- and 4-player versions, agents are thus required to both work with less first-hand information and need to rely more on hints and interpreting the moves made by those around them.
Not only this, but agents are tasked to do this tracking for more players as well.
Thus, adding players to the YLE increases the difficulty substantially in two ways.

\paragraph{Additional Details on Legal Positions}
\begin{figure}[t]
    \centering
    \includegraphics[width=\linewidth]{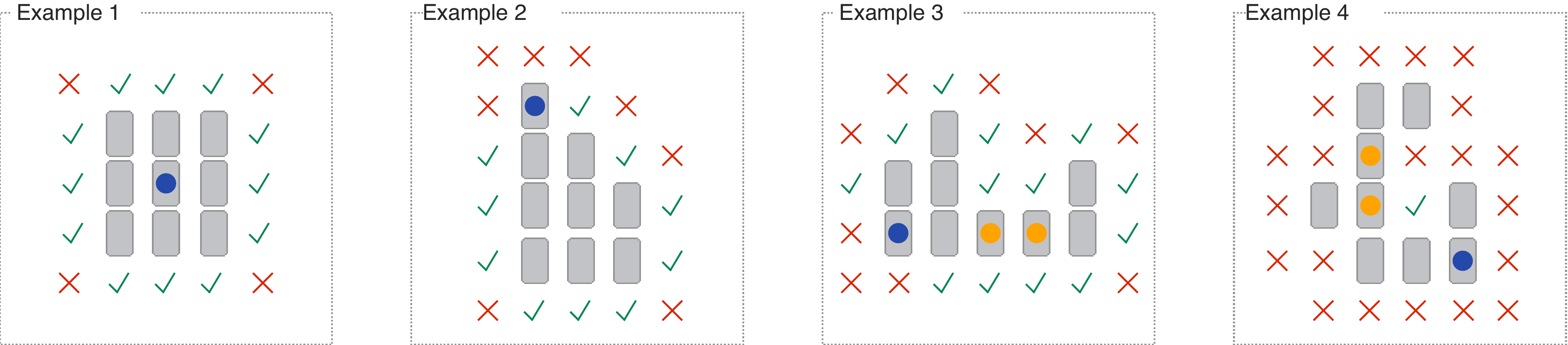}
    \caption{Valid target positions for the card with the \textcolor{cardBlue}{blue dot}. Cards can only be moved so that all cards remain connected via their sides. Cards with an \textcolor{orangeAgent}{orange dot} cannot be moved at all legally. Example 1 shows the initial position. All cards can be moved in the initial configuration. Example 2 stems from the early game. Examples 3 and 4 display configurations that might be encountered during the mid or late game.}
    \label{fig:legalmovesExtended}
\end{figure}
We display several examples of legal target positions given a card to be moved in \autoref{fig:legalmovesExtended}.
These only display a subset of all legal moves since the set of legal moves is different for each card on the board.
For some cards, there are no legal moves whatsoever.
This typically occurs when a card occupies a position in a narrow band that connects two clusters, such as in example 3.
In example 3, there is no conceivable target position for both cards marked with an orange dot that would keep the cluster of cards connected, except their current position.
Note that cards might be movable in the future again, i.e. if other cards are moved such that cards are freed again.
In example 3, for instance, if the card with the blue dot is placed above any of the cards marked with an orange dot, both cards gain one additional legal target position, which makes them movable.
Moving cards in a way that keeps option value (i.e. that keeps many cards moveable) is thus recommended.
Observe example 4, for instance, here players placed cards in such a way that for most cards, very few target positions are still legal.
This will usually highly restrict the agents' ability to sort cards.

\subsection{Second Order ToM Reasoning Example Extended}
\label{sec:secondorderexample}
\begin{figure*}[t]
    \centering
    \includegraphics[width=\linewidth]{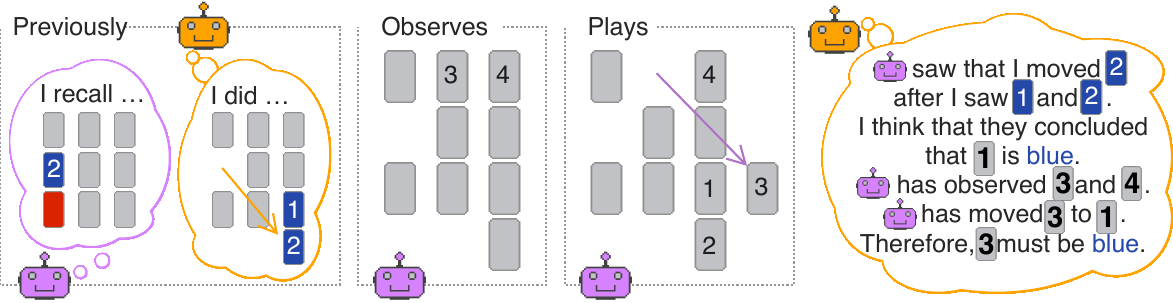}
    \caption{Here we reexplain the second-order ToM reasoning example from the main paper over multiple timesteps in 2-player YLE in additional detail (see text): \orange beliefs that \purple beliefs that he (\orange) knew that cards 1 and 2 are of the same colour. Even though \purple never saw card 1 and \orange never observed card 3, both now know where all blue cards are, that they are grouped and that both share this knowledge as part of their common ground. In the future, they can now both observe and move different cards to finish early as they have one less card that needs to be observed.}
    \label{fig:2ndOrderToMExampleExtended}
\end{figure*}
\autoref{fig:2ndOrderToMExampleExtended} displays the same example for second-order ToM reasoning in 2-player Yokai as the main paper.
Here, we expand on the example.
\autoref{fig:2ndOrderToMExampleExtended} shows an agent (\orange) making inferences about what they assume the other agents (\purple) knows due to their own previous action. 
In the example \orange infers the colour of the card due to the behaviour of the other agent.
\orange reasons that because of how \purple played, \purple must have interpreted his move correctly.
\orange thus reasons that \purple thinks that \orange knew that cards 1 and 2 have the same colour and must go together.
This is the case, \orange made this inference correctly, and card 3 indeed is also blue.
This reasoning allows agents to require fewer first-hand observations to get an overview of the board, potentially enabling them to end early.

Note that an interesting strategic aspect of Yokai is the amount of shared information agents acquire together (both \purple and \orange choose to observe card 2) versus how much information they try to acquire asymmetrically to then use to hint knowledge to others via hints or actions (cards 1 and 3).
Agents thus effectively control how big their shared knowledge is and when to strategically expand it.
This has interesting strategic depth.
Agents will differ in their strategic preference here, and developing strategies that are suitable for a wide range of partners is therefore a challenge.

This is just one illustrative example of ToM reasoning in YLE.
Belief reasoning in the YLE can range from recalling previously observed cards (zero-order ToM), to tracking what another player knows (first-order), to inferring what one player believes about another’s beliefs, including their beliefs about oneself (second-order and beyond).

Note that such higher-order reasoning steps are even likelier to occur in settings involving more partner agents.

\section{Training Details}
\label{sec:training_details}
We share all the details of the training here, including all hyperparameters used in training and all details on the network architectures and how we tuned them.
In general, we tuned many aspects of our policy to arrive at the strongest agents we could for our experiments, see below.
We took extra care in making these equal to make our results comparable.

\subsection{Reinforcement Learning Details}
\label{sec:rldetails}
\begin{table}[t]
    \centering
    \caption{Hyperparameters of the learning process.}
    \label{tab:rlhyperparams}
    \begin{tabular}{l r r} %
         \toprule
        \textbf{Description} & \textbf{Value} \\
        \midrule
        Optimiser & Adam~\citep{kingma2015adam} \\
        Adam $\beta_1$ & $0.9$ \\
        Adam $\beta_2$ & $0.999$ \\
        Adam $\epsilon$ & $1 \cdot 10^{-5}$ \\
        Learning Rate $\eta$ & $3 \cdot 10^{-4}$\\
        Learning Rate Annealing & Yes (linear)\\
        Maximum Grad Norm & $0.5$ \\
        \midrule
        \# Simulators & $1024$ \\
        Discount Rate $\gamma$ & $0.99$ \\
        GAE $\lambda$ & $0.95$ \\
        Entropy Coefficient & $0.01$ \\
        Value Loss Coefficient & $0.5$ \\
        \# PPO Epochs & $4$ \\
        \# PPO Minibatches & $4$ \\
        \# PPO Steps & $128$ \\
        PPO Value Loss & Clipped \\
        PPO Value Loss: Clip Value & $0.2$ \\
        PPO Value Loss: Scale Clip Value & No \\
        Reward Shaping & Yes (linearly annealed) \\
         \bottomrule
    \end{tabular}
\end{table}
Our work employs a standard IPPO algorithm \citep{deWitt2020is} for all experiments.
We present all parameters of the learning process in \autoref{tab:rlhyperparams}.
Only for OBL we reduce the number of parallel environments to $256$ since with $K=8$ we are otherwise unable to fit the OBL policies into memory.
Similarly, for 9C 4P HE-IPPO we reduce the number of parallel environments to $512$ because of the memory requirements when tuning $12$ seeds over many entropy coefficients.
Note that this actually \textit{increases} the number of gradient updates the policies receive. 

\subsection{Neural Network Architectures}
\label{sec:nnDetails}
Our work employs several encoder architectures.
We present all the details on them below.

\paragraph{General Policy Architecture}
In general, in our work, no matter the encoder, the final policy network always has the same structure.
That is given an embedding $e_{\text{encoder}}$ produced by an encoder network $f_{\text{encoder}}$, we pass the embedding to a linear projection, then a GRU \citep{cho2014learning} and finally either a to value or action head.
We use ReLU \citep{Fukushima1975CognitronAS, Nair2010Rectified} as the activation function.
The layers are initialised using orthogonal initialisation with a scale of $\sqrt{2}$ and bias $0$.
We determined the number of policy layers and their size using hyperparameter tuning.
Additionally, for all neural network architectures detailed below, we also varied and explored: The number of encoder blocks (2, 3 and \textbf{4}), the hidden dimension of the encoder blocks, the aggregation function after the encoder (mean, sum, multiply and \textbf{concatenate}) and whether to anneal the dense rewards (False and \textbf{True}).
We also compared addition additional hidden layers between GRU and the action/value heads (\textbf{0}, +1HL, +2HL).
Finally, the value head produces a singular scalar output and is initialised using orthogonal initialisation with a scale of $0.00$ instead of $\sqrt{2}$.
We initialise the action head in the same way but it produces logits for all possible actions instead.

\paragraph{GCN, GATv2 and the pre-embedding of Nodes}
\begin{figure*}
    \centering
    \includegraphics[width=\textwidth]{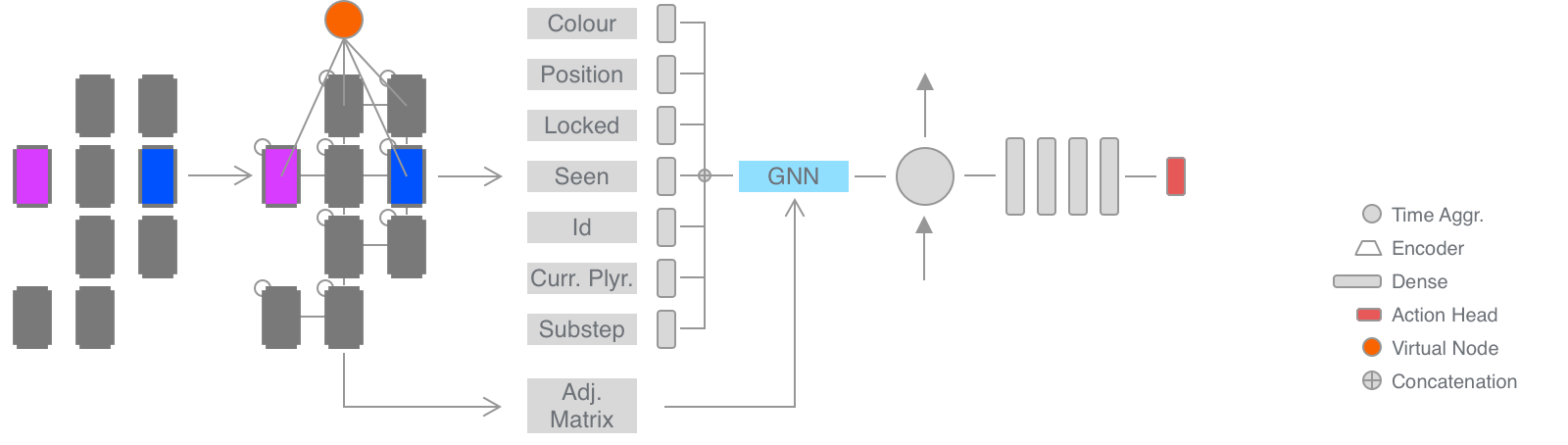}
    \caption{A sketch of the GNN and pre-embedding architecture used in this work.}
    \label{fig:gnnarch}
\end{figure*}
We use two graph neural networks in our work, one based on graph convolutions \cite{kipf2016semi} and one based on the attention mechanism \cite{brody2021attentive}.
We pick these two as they stem from two different families of graph learning architectures and they thus are representative of a larger body of work.
In our work, both methods observe the graph directly as described in the paper.
There are two additional implementation details to note, though.
First, both graph architectures observe an input graph that features a virtual node, similar to the hub-nodes of \citet{abdessaied24_wacv} or the \texttt{[VNode]} token of \citet{ying2021transformers}.
A virtual node is an empty node that connects to all other nodes in the graph.
It solely serves the purpose of reducing hop distances between nodes.
The motivation for employing them in the YLE is that since cards can be distant from each other in terms of hop distance, information from one card might not propagate to the next during message passing.
I.e. imagine the GNN observes two blue cards on opposite sides of the current card cluster.
If the number of message-passing rounds is too low, this information might never be aggregated during message-passing.
In RL, neural networks tend to be comparatively small, which is also true in our work.
This also limits the number of message-passing steps to be small (2 in our case).
With a virtual node, however, this is not an issue since the virtual node aggregates information from all other nodes in one single message-passing step.
Second, our work uses pre-embeddings of node-feature modalities as described in \cite{bortoletto2024neural}.
We take the node features of size $n \times f$ and embed them per modality via a separate dense encoder before then concatenating all of the resulting vectors into new node features shaped $n \times f'$.
$f'$ depends on the size of these dense layers.
We sketch the overall approach in  \autoref{fig:gnnarch}.
In general, we found that embedding the node features per modality before using them in the GNN improved performance following earlier work, i.e. see \citet{bortoletto2024neural}.
Specifically, for each modality $m \in M$ we train a dense layer $f_m$ that embeds the modality per node into an embedding $e_m \in \mathbb{R}^{b \times t \times n \times 2}$ for the current player embedding or $e_m \in \mathbb{R}^{b \times t \times n \times 4}$ for all other modalities where $b$ is the batch size, $t$ the time and $n$ the number of nodes.
All embeddings are then concatenated into a node embedding $e_{\text{node}} = e_{\text{colour}} \oplus e_{\text{position}} \oplus e_{\text{locked}} \oplus e_{\text{seen}} \oplus e_{\text{id}} \oplus e_{\text{is\_current\_player}} \oplus e_{\text{substep}}$ with $e_{\text{node}} \in \mathbb{R}^{b \times t \times n \times f'}$ with $f' = 26$ using the configuration described.
We then add self-connections, the virtual node using a zero-vector $e_{\text{virt}} \in \{0\}^{f'}$ and connect the virtual node to all real nodes before passing them to our graph encoder $f_{\text{GNN}}$.
The output dimension of the graph encoder is $64$.
For GATv2, we use $4$ heads.
We apply $4$ graph layers in the relational encoder comparison.
The resulting embedding is then concatenated along the node dimension and fed to the rest of the policy.

\paragraph{Relation Module}
Our work adapts the relation module of \citet{Zambaldi2018Relational}.
We only use the self-attention mechanism of their proposed relation module and skip the CNN encoder they propose alongside the self-attention layers.
This is because our setting deals with an environment where agents observe the position of cards directly as opposed to the setting of \cite{Zambaldi2018Relational} that deals with image inputs directly.
Thus, in our relation module, non-local interactions between all cards are simply computed using self-attention directly.
Similarly to the GNN encoders, we also pre-embed the modalities of all nodes using the same mechanism as discussed above.
The resulting embedding is then also passed to the remaining policy.
Our relational module is also applied $4$ times, features $4$ heads, and its output dimension per layer is $64$.

\paragraph{CNN}
Our work compares a CNN encoder and determines it to be the best-performing model.
As described earlier, the YLE returns an image-like observation to be processed by our CNN.
The exact size of the input tensor to the CNN depends on the YLE settings, but in 2-player 9C YLE, the input dimension is $9 \times 10 \times 13$, which we will use as a running example.
It includes the same features as discussed above, with some minor changes.
The first three channels include a one-hot encoding of cards' colours, the next three an encoding of hint colours, followed by two channels for the position, one for whether the card is locked, one for whether the observing agent is the current player, one for the number of the current substep (1-4) and one that identifies the card via an identifier number.
Our convolutional stack uses $4$ 2d convolutions with $64$ filters each and ReLU applied between them.
The kernel size is $(3,3)$, we apply a stride $1$ and valid padding.
The resulting embedding is flattened to produce the final embedding to be passed on to later layers.

\paragraph{TransformerXL}
Next to the encoder architectures, our work compares memory architectures.
TransformerXL \citep{dai2019transformerXL} is an explicit memory mechanism that shows great performance in memory-intensive RL tasks \citep{Parisotto2020Stabilizing}. 
Our usage strictly follows the one described by \citet{Parisotto2020Stabilizing} as implemented by \citet{Hamon2024TransformerXLJaxGithub}.
Similar to our GRU architecture, we first pass the embedding received by the encoder through a dense layer $f_{\text{pre\_trxl}}$ before handing the resulting embedding to the transformer.
We use a hidden dimension of $64$ for both the dense layers within the transformer as well as the projection layers in multi-head attention.
This is to keep the number of parameters consistent between the architectures.
We furthermore configure the transformer to have $8$ heads and to use gating with a gating bias of $2$.
We use a single transformer layer.
Finally, we set the length memory window to $32$. 
$32$ is long enough to recall all of the previous moves in the 9C 2P YLE.
Memories from a past game are masked.
We apply standard positional encoding based on sinus waves.

\subsection{Additional Details for Other-Play in \yokai{}}
\label{sec:other-playdetails}
\begin{figure}
    \centering
    \includegraphics[width=\linewidth]{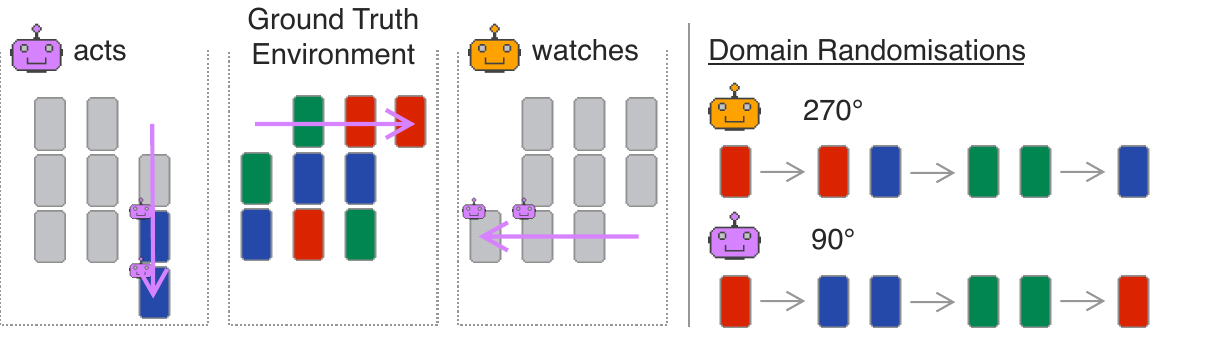}
    \caption{An example of how both agents act in rotated and recoloured environments to break symmetries for zero-shot coordination.}
    \label{fig:recolouringAndRotation}
\end{figure}
The core idea of Other-Play (OP) is that during self-play training, agents learn to break symmetries of the environment in arbitrary ways that do not generalise to novel partners.
In Hanabi, for instance, agents might form conventions around certain colours.
OP presents agents with a special asymmetric domain randomisation that makes it impossible for agents to coordinate on breaking symmetries in certain ways.
In Hanabi, this is achieved by a recolouring operation that shows each agent a different recolouring of the game state \citep{Hu2020Other}.

Therefore, to train agents via OP for the YLE, one first needs to find all classes of symmetries that exist in an environment, and then, second, find an appropriate domain randomisation operation.
Generally, since agents also observe coloured cards in the YLE, we can apply the same recolouring operation as \citeauthor{Hu2020Other}.
However, since the YLE features a spatial grid, additional symmetries exist around the location of cards.
Even if agents can not coordinate on specific colours when we apply recolouring, they are, for example, able to coordinate on sorting the first colour they find in a specific place.
Agents, for instance, might choose to observe the same two cards at the start and place one of them in the top left corner.
This is an arbitrary convention (top left vs bottom right etc.).
To prevent this, we apply asymmetric rotation to the agents' observations.
This way, when one agent places a card next to the top-left corner in their environment, their partner agents observe this as placing the card next to potentially any corner in different episodes. 
Training with rotated observations also requires rotating the agent's actions accordingly.
We show an example of this process in \autoref{fig:recolouringAndRotation}.

Note that these are not the only ways for agents to learn to break symmetries.
Recent work \citep{treutlein2021new} showed that agents also might coordinate to break symmetries in arbitrary ways over multiple timesteps.
It is impossible to list all possible classes of symmetries for an environment with enough complexity.
This is a drawback of the OP algorithm as it requires them as input.
In our work, we thus focus on showing that while Hanabi only requires breaking one symmetry for efficient ZSC while the YLE -- due to its additional complexities -- requires additional effort to improve ZSC performance, which is still far from optimal.  

\section{Additional Probing Details}
\label{sec:probingSetupAppendix}
\begin{figure}
    \centering
    \includegraphics[width=0.5\linewidth]{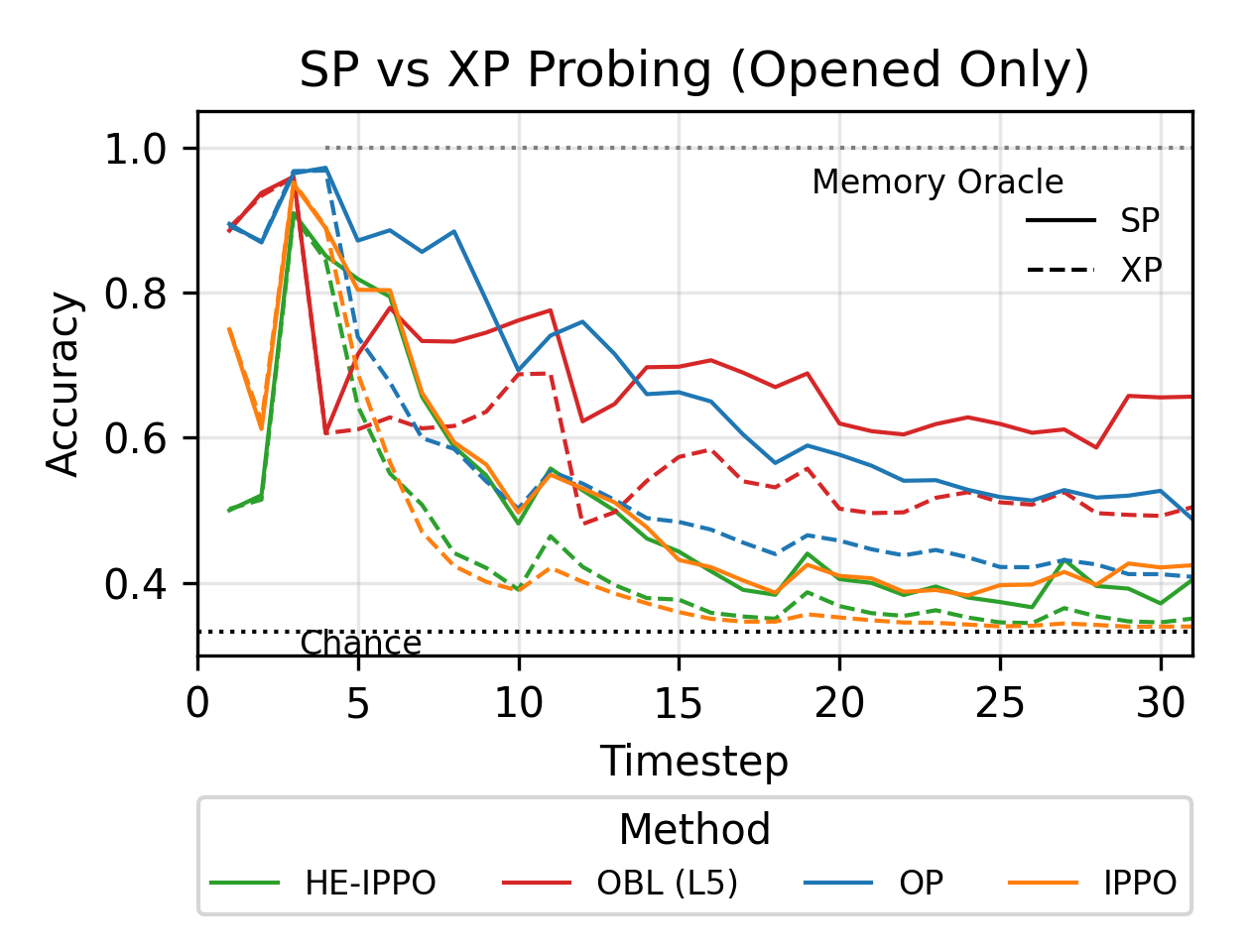}
    \caption{Probing results were only known card colours from     opened cards where probed. We can see that agents start of representing card colours in a linearly decodeable way but as the game progresses and more cards are opened, accuracy decreases.}
    \label{fig:probingSPvsXP_opened}
\end{figure}
\begin{figure}
    \centering
    \includegraphics[width=0.5\linewidth]{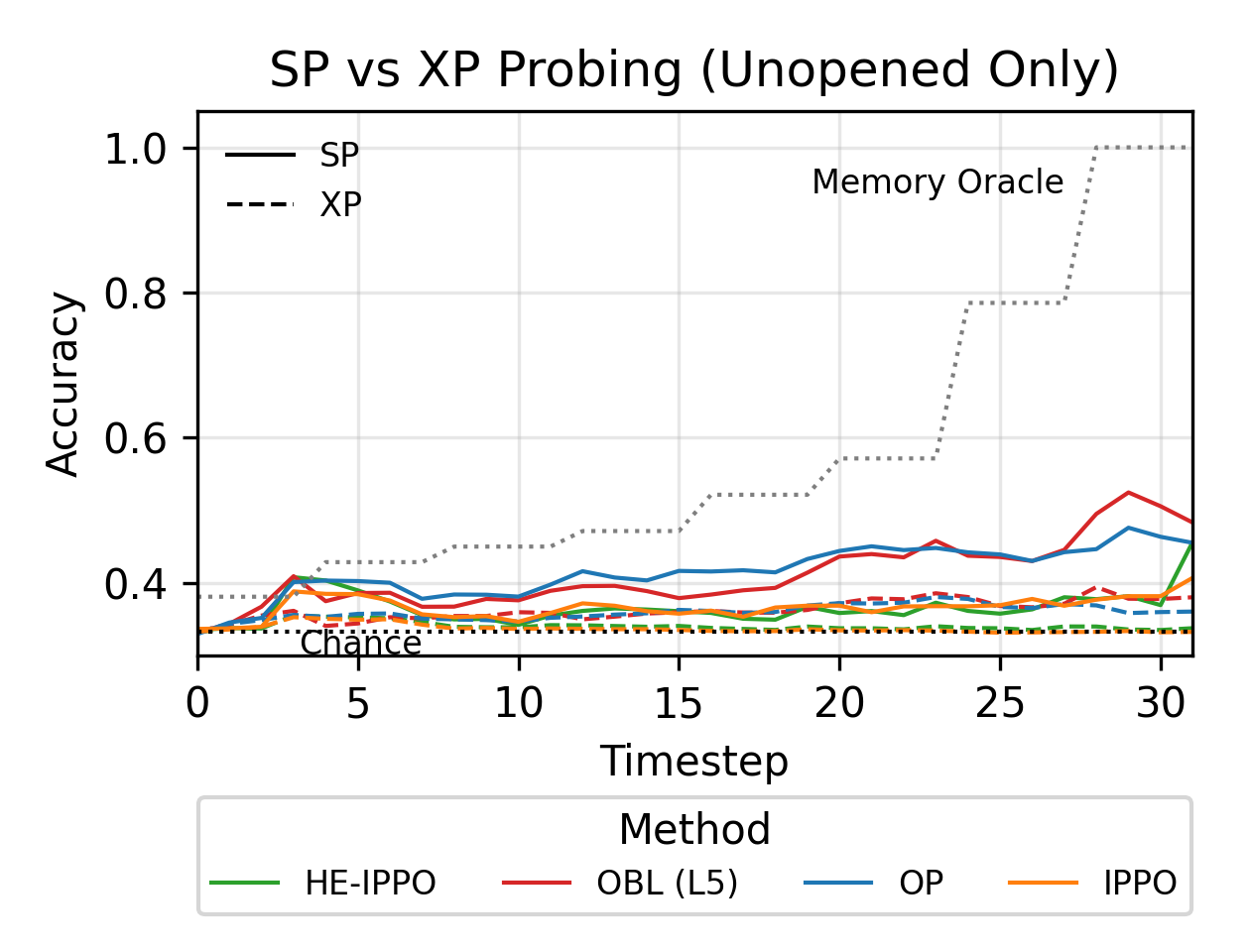}
    \caption{Probing results were only unknown card colours from unopened cards where probed. As the game progresses card colours become increasingly linearly decodable but remain under a perfect-memory oracle. Only OBL and OP represent unknown cards in a linear-decodable fashion above random chance.}
    \label{fig:probingSPvsXP_unopened}
\end{figure}
Over the duration of $5,000$ games, we collected a probing dataset $\mathcal{D} = \{x^{(i)}, y^{(i)}\}$.
The probing datasets maps the activations of an agent to a label $z$.
A \textit{probe} is a function $g: f_l(x) \to \hat{z}$ and that is trained on $\mathcal{D}$ where $f$ is a neural network and $f_l(x)$ is the intermediate representation of $x$ at layer $l$.
We always probe before the action head of our agent.
As highlighted in the main text, during all experiments, we train one and evaluate it on a hold-out test set $\mathcal{D}_{\text{test}} \subset \mathcal{D}$.
During all experiments, we trained one probe per position (81 total, 12 seeds = 972 probes) and averaged the scores computed on a hold-out test over all probes.
In the main text, we experiment with probing all cards (no matter if they have been observed).
In \autoref{fig:probingSPvsXP_opened} and \autoref{fig:probingSPvsXP_unopened} we show the opened versus unopened cards distinction. 

\section{Human Performance in \yokai{}}
\label{sec:humanplay}
We collected human game performance in a setting that adheres to the same setting that artificial agents face, i.e. participants played with 9 cards on a $9 \times 9$ grid in a two-player setting.
Humans played in person.
A pair of subjects was instructed on the rules verbally.
Each pair of subjects was given a single training round to make sure that all participants were familiar with the rules.
To make results as comparable as possible, humans were instructed to avoid any form of non-verbal communication.
Each pair played five evaluation rounds, and we report mean return across games. 
Early termination was allowed according to the same rules as in the agent environment.
The study was conducted under an umbrella agreement of the relevant institutional review board.
Compensation was also handled under the rules of the relevant institution. 

\newpage
\section{Example Games}
\label{sec:example_games}
\begin{figure*}[!h]
    \centering
    \includegraphics[width=\linewidth]{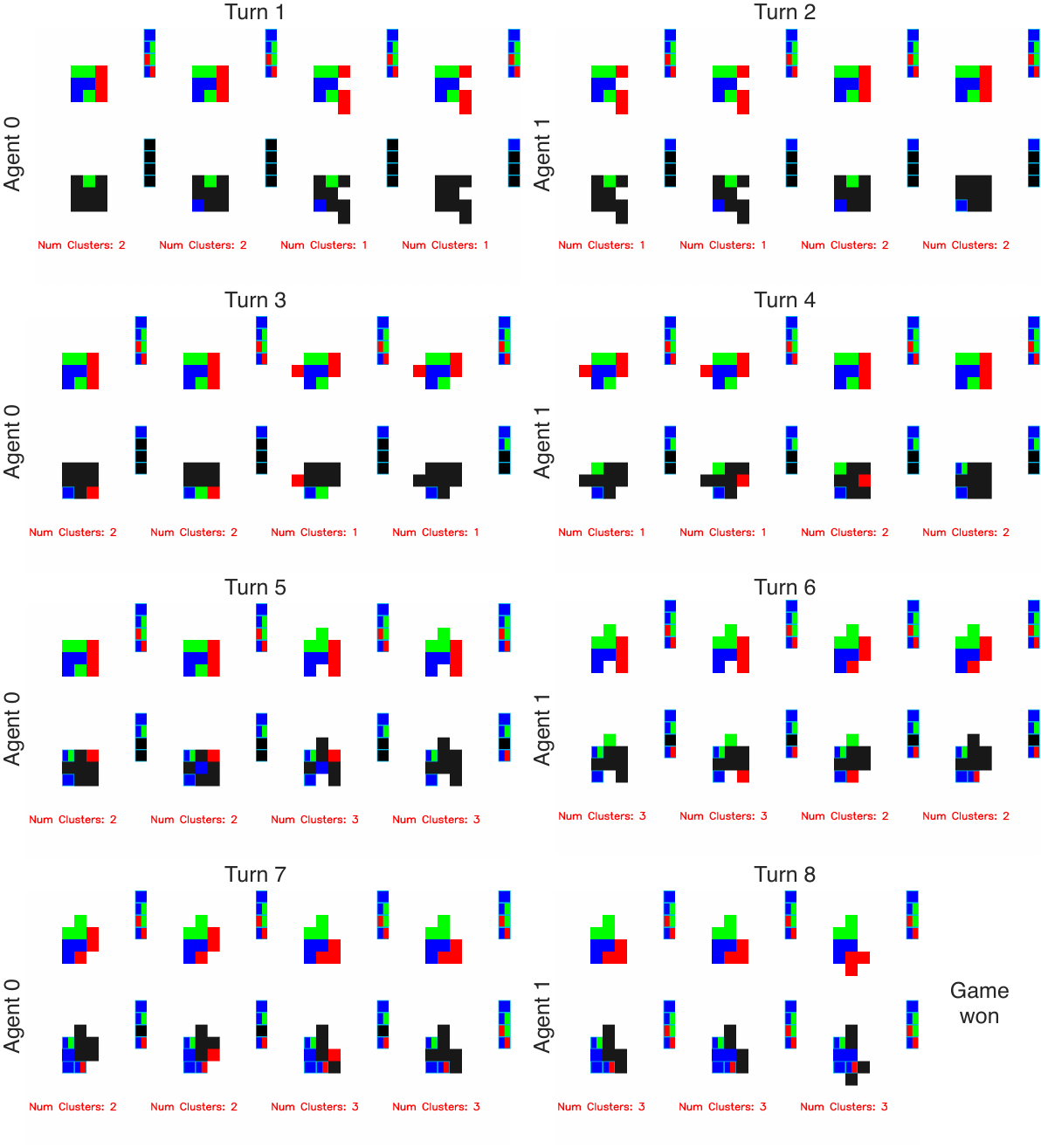}
    \caption{We show an example game in which an agent \textbf{wins} in self-play. For each turn, we display the ground truth information at the top and the current observable information -- essentially the agents' observation -- at the bottom. We also show how many card clusters exist at this point per turn. Hint cards are highlighted by a \textcolor{hintBorder}{light-blue border}. In turn 7, the agent finishes the final cluster of cards by correctly inferring that the card under the $rb$-hint must be \textcolor{cardRed}{red} and not blue.  In turn 8, agent 1 should have finished early for more reward, but did not recognise that fact and instead used the last hint card, thus finishing the game. Note that the agents tend to play quite conservatively. In turns 1 and 3, for instance, it gains knowledge of two separate \textcolor{cardGreen}{green cards} but only groups them in turn 5 after agent 1 places a corresponding $gb$-hint in turn 4.}
    \label{fig:WinExample}
\end{figure*}
\begin{figure*}[!h]
    \centering
    \includegraphics[width=\linewidth]{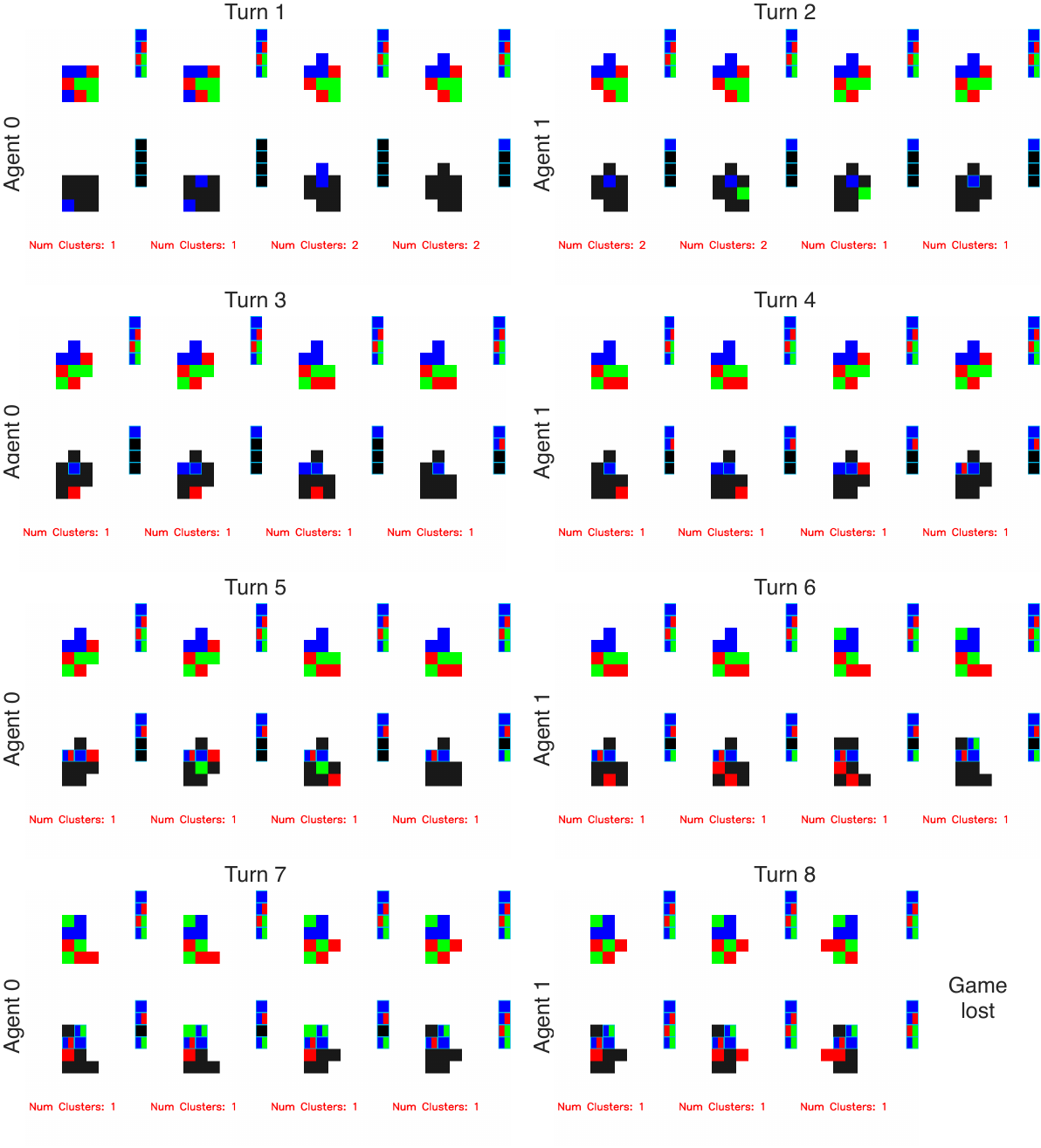}
    \caption{We show an example game in which an agent \textbf{loses} in self-play. For each turn, we display the ground truth information at the top and the current observable information -- essentially the agents' observation -- at the bottom. We also show how many card clusters exist at this point per turn. Hint cards are highlighted by a \textcolor{hintBorder}{light-blue border}. Note that agent 0 in turn 1 finds a \textcolor{cardBlue}{blue group} early and manages to form a cluster quickly. Agent 0 confirms this by turn 3. In turn 6, the game is still winnable. However, because both agents over-focused on establishing a common ground around the blue cluster, they fail to realise this and in turn 7, agent 0 then blunders and misplaces the \textcolor{cardGreen}{green card}. From there, the game is lost under all circumstances. Additionally, note that in turns 4 and 6, agent 1 establishes the identity of all red cards but fails to recall that in turn 6, therefore also misplaying.}
    \label{fig:LoseExample}
\end{figure*}
We show an example game an agent wins in self-play in \autoref{fig:WinExample} and one where it loses in \autoref{fig:LoseExample}.

\newpage
\section{2 Player Cross-Play Matrices}
\label{sec:cross-playmatrices}
\begin{figure}[!h]
    \centering
    \includegraphics[width=\linewidth]{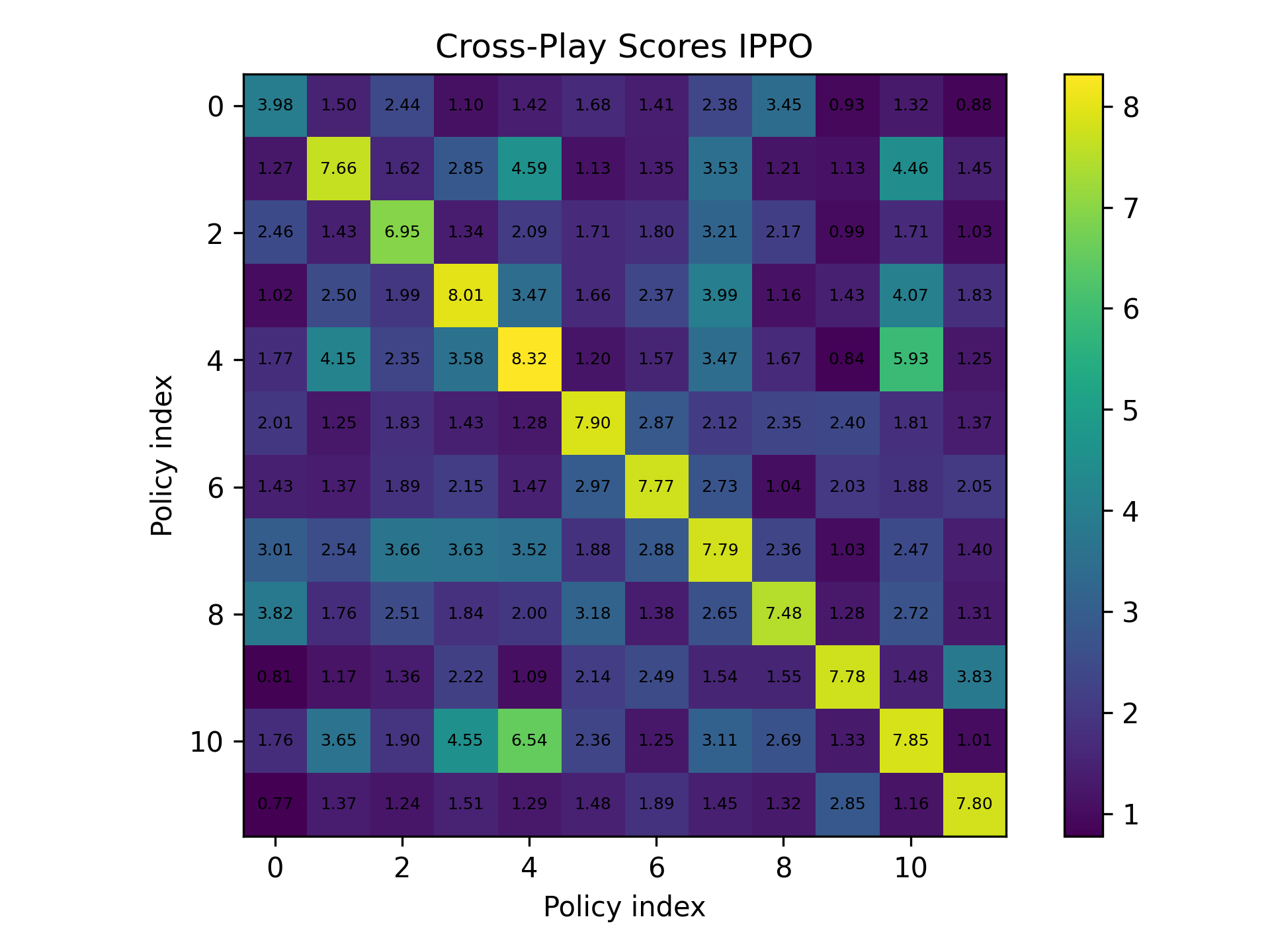}
    \caption{Cross-play scores for 2 player IPPO as featured in \autoref{tab:performance} ($\alpha = 0.05$).}
    \label{fig:sp-2p-cross-play}
\end{figure}
\begin{figure}[!h]
    \centering
    \includegraphics[width=\linewidth]{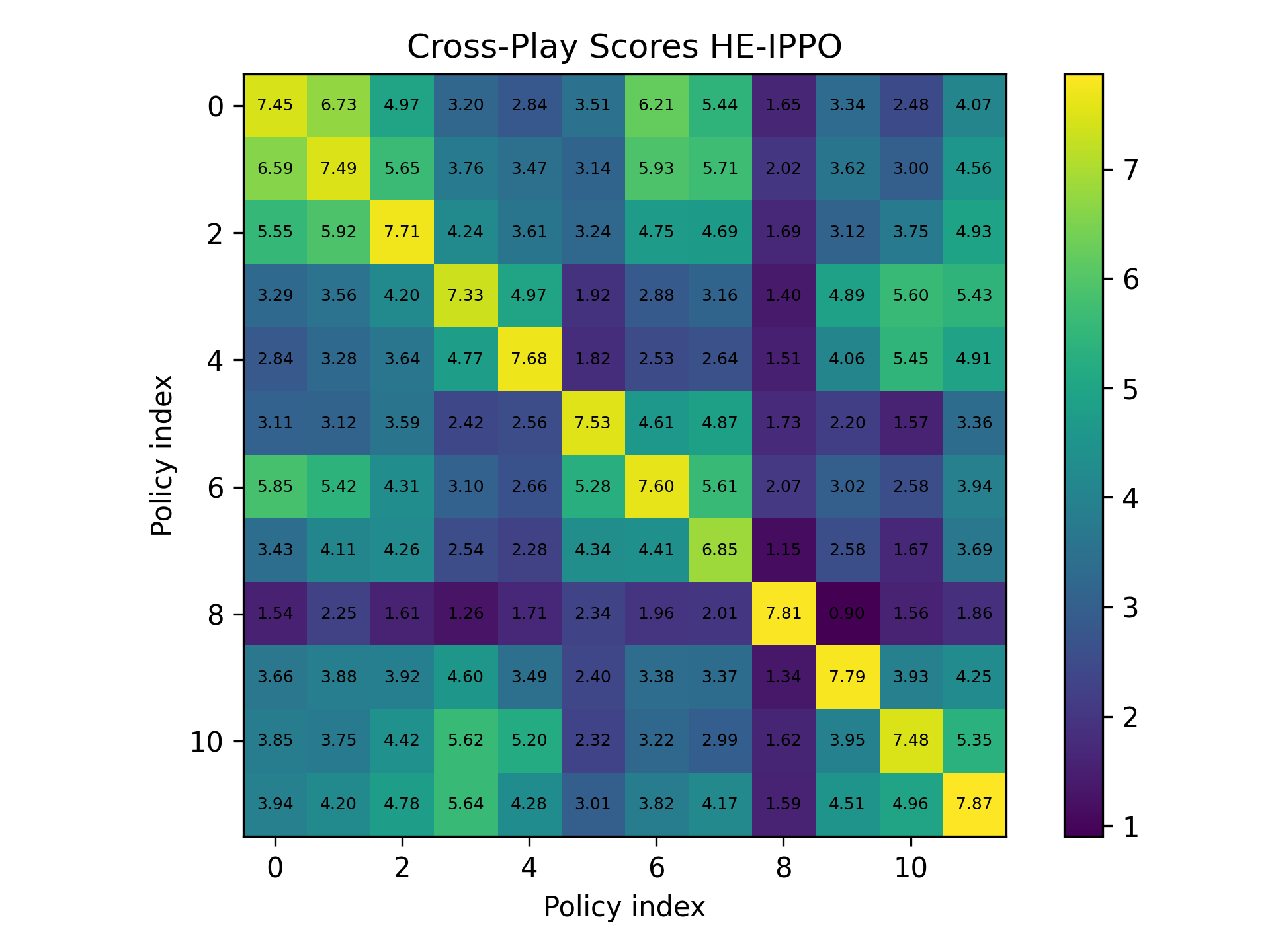}
    \caption{Cross-play scores for 2 player HE as featured in \autoref{tab:performance} ($\alpha = 0.07$, $\lambda_{\text{GAE}} = 0.85$).}
    \label{fig:HE-2p-cross-play}
\end{figure}
\begin{figure}[!h]
    \centering
    \includegraphics[width=\linewidth]{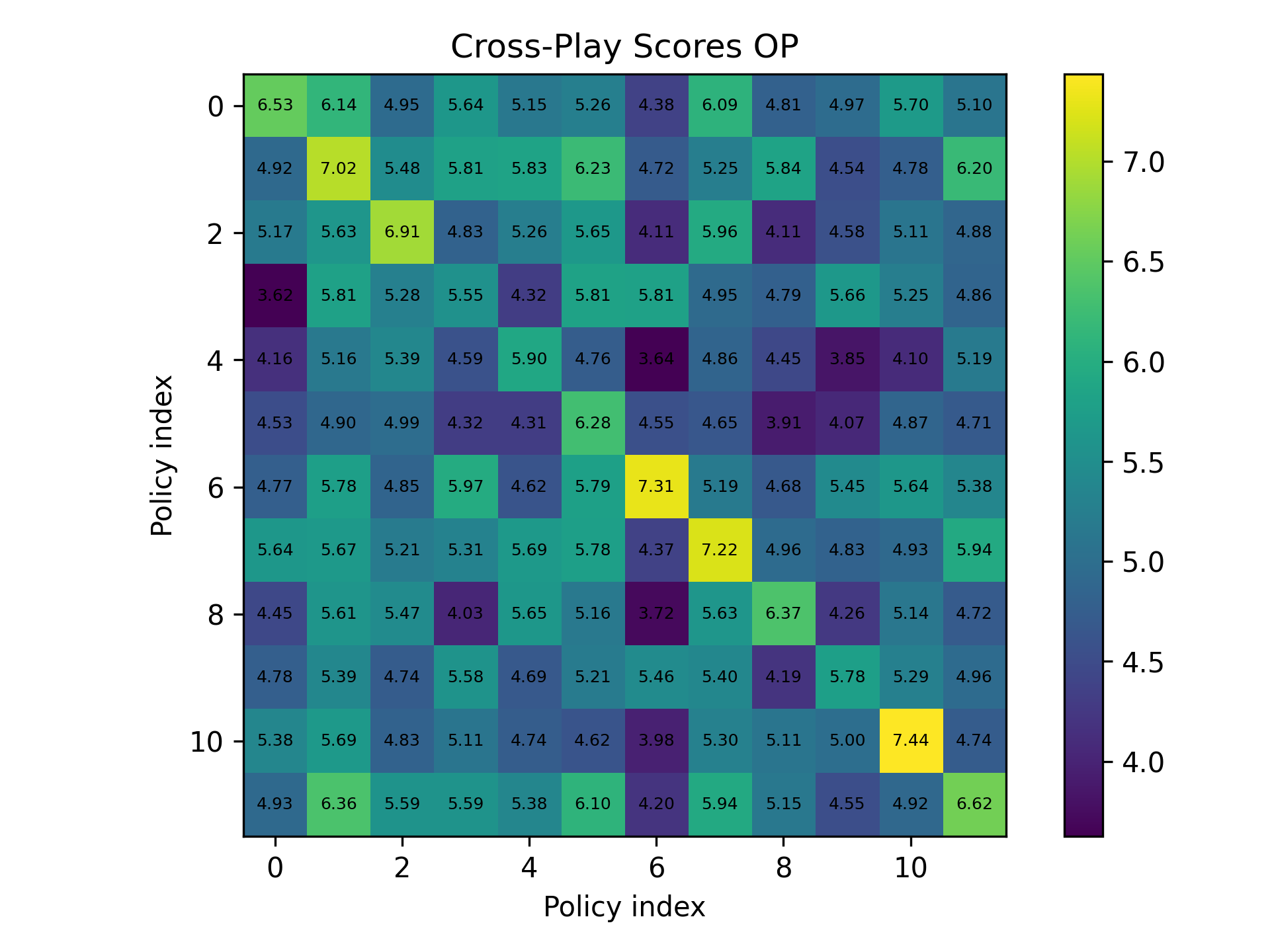}
    \caption{Cross-play scores for 2 player OP as featured in \autoref{tab:performance}.}
    \label{fig:op-2p-cross-play}
\end{figure}
\begin{figure}[!h]
    \centering
    \includegraphics[width=\linewidth]{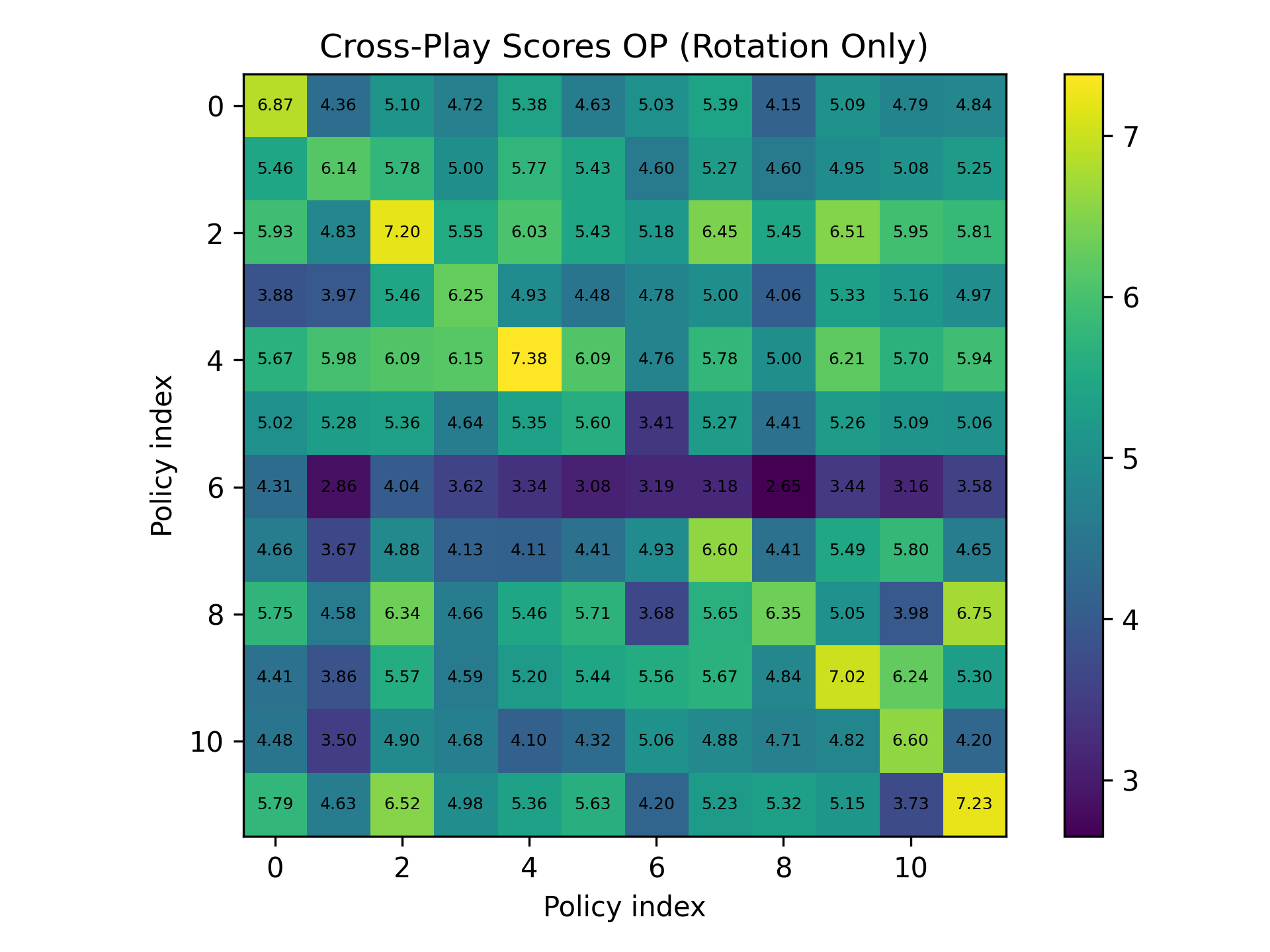}
    \caption{Cross-play scores for 2 player OP ($\Phi_{\text{rot}}$ only) as featured in \autoref{tab:performance}.}
    \label{fig:op-2p-cross-play-rotation}
\end{figure}
\begin{figure}[!h]
    \centering
    \includegraphics[width=\linewidth]{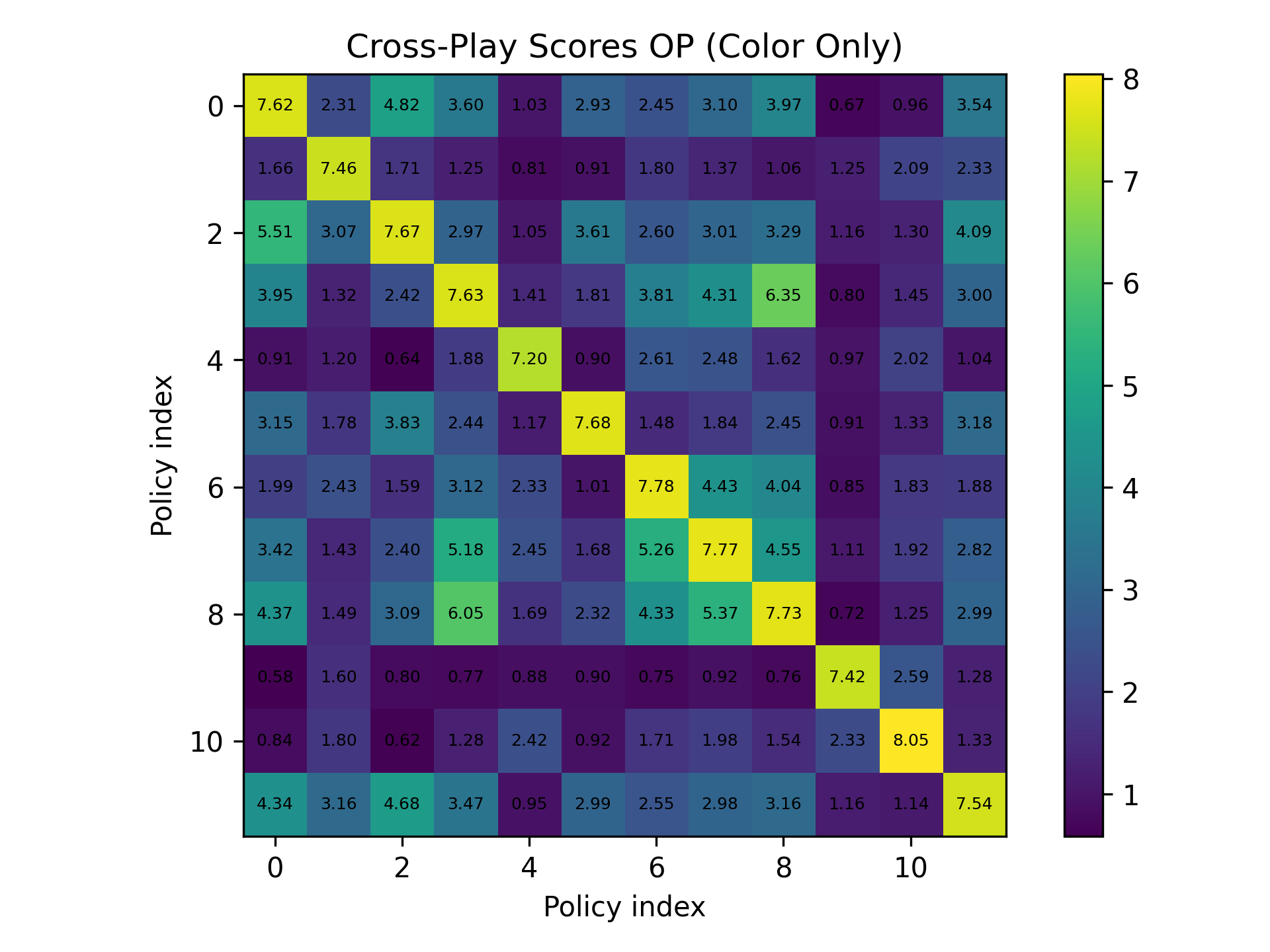}
    \caption{Cross-play scores for 2 player OP ($\Phi_{\text{colour}}$ only) as featured in \autoref{tab:performance}.}
    \label{fig:op-2p-cross-play-color}
\end{figure}
\begin{figure}[!h]
    \centering
    \includegraphics[width=\linewidth]{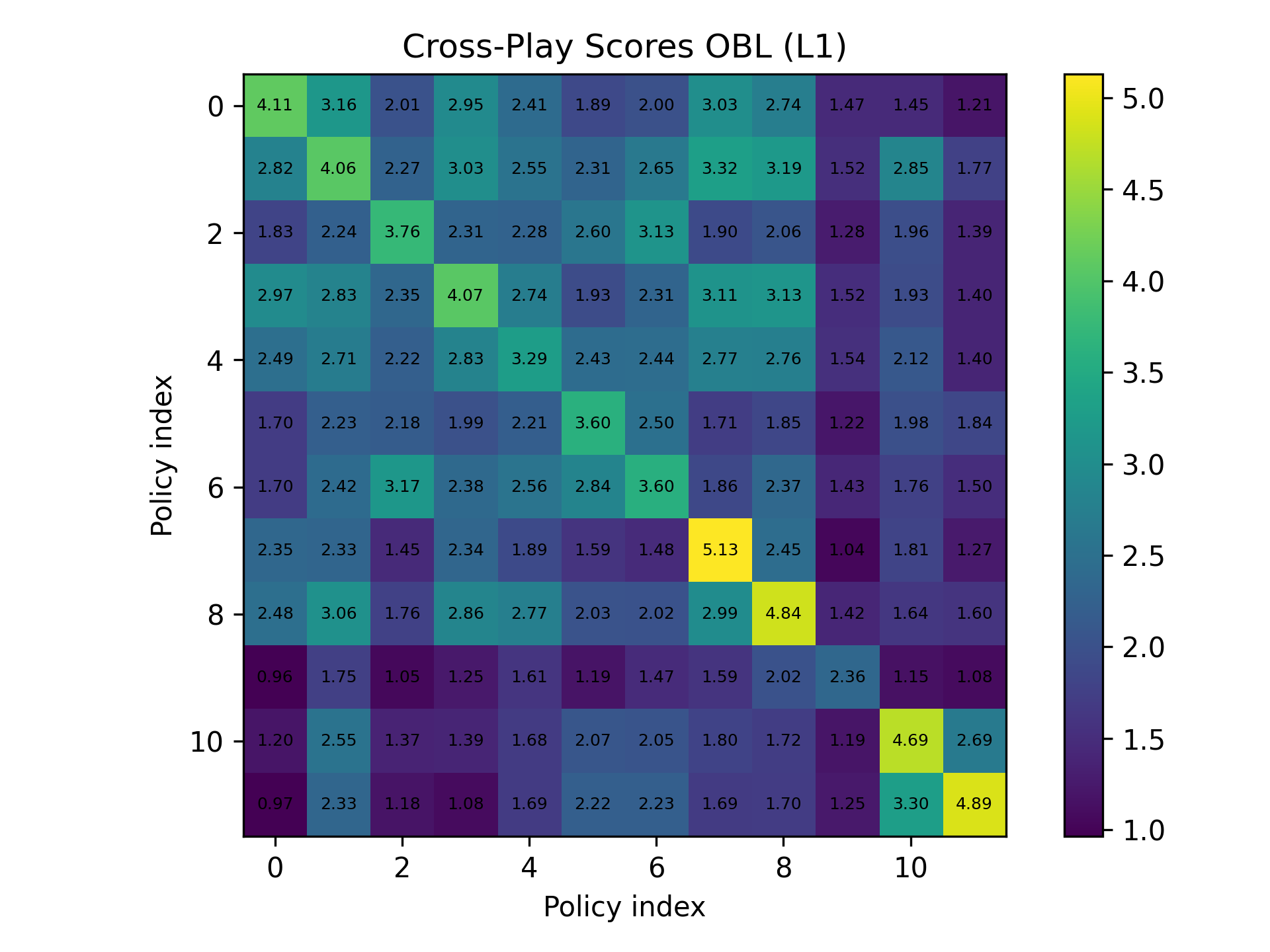}
    \caption{Cross-play scores for 2 player OBL level 1 as featured in \autoref{tab:performance}.}
    \label{fig:obl-l1-2p-cross-play}
\end{figure}
\begin{figure}[!h]
    \centering
    \includegraphics[width=\linewidth]{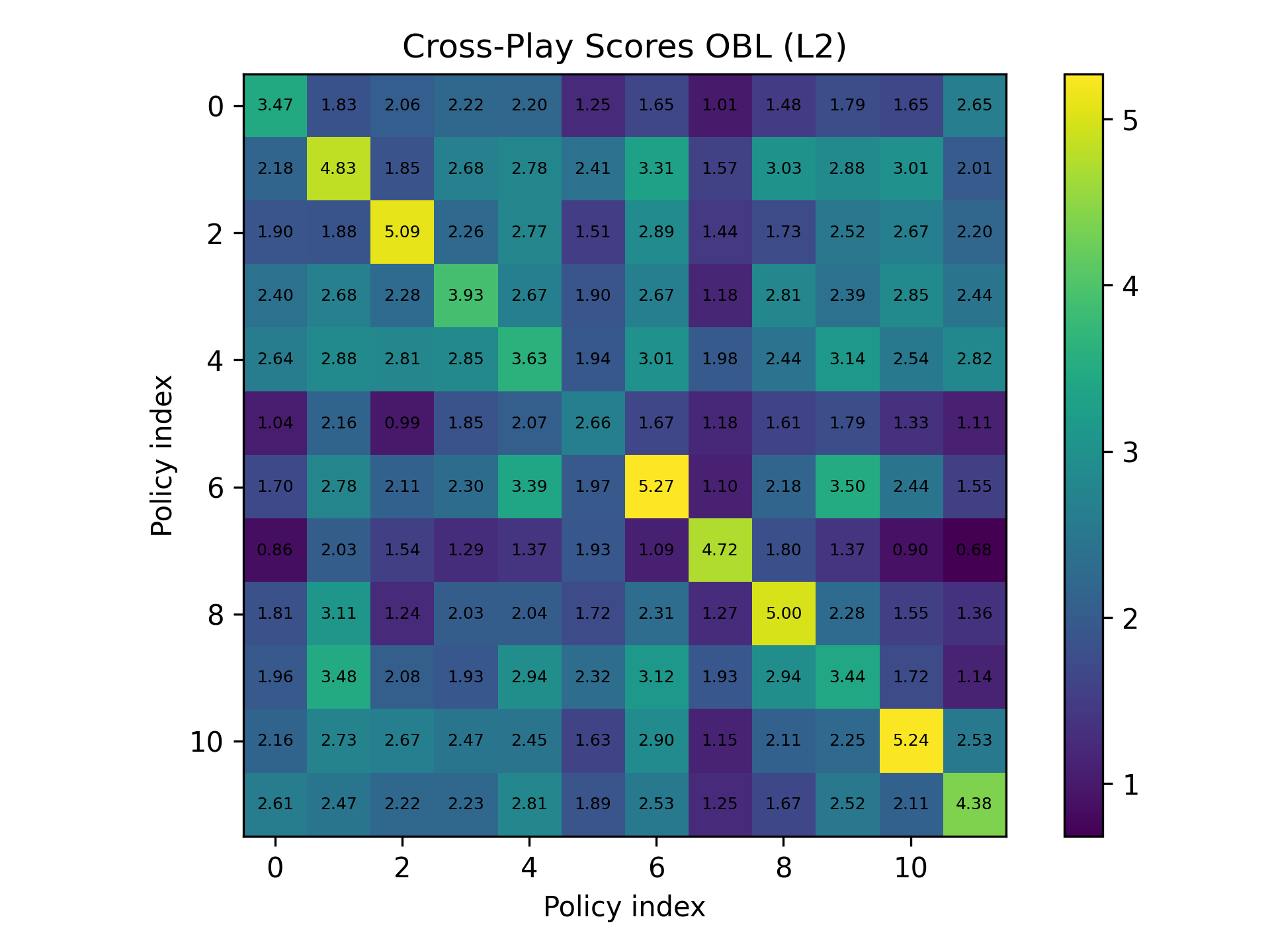}
    \caption{Cross-play scores for 2 player OBL level 2 as featured in \autoref{tab:performance}.}
    \label{fig:obl-l2-2p-cross-play}
\end{figure}
\begin{figure}[!h]
    \centering
    \includegraphics[width=\linewidth]{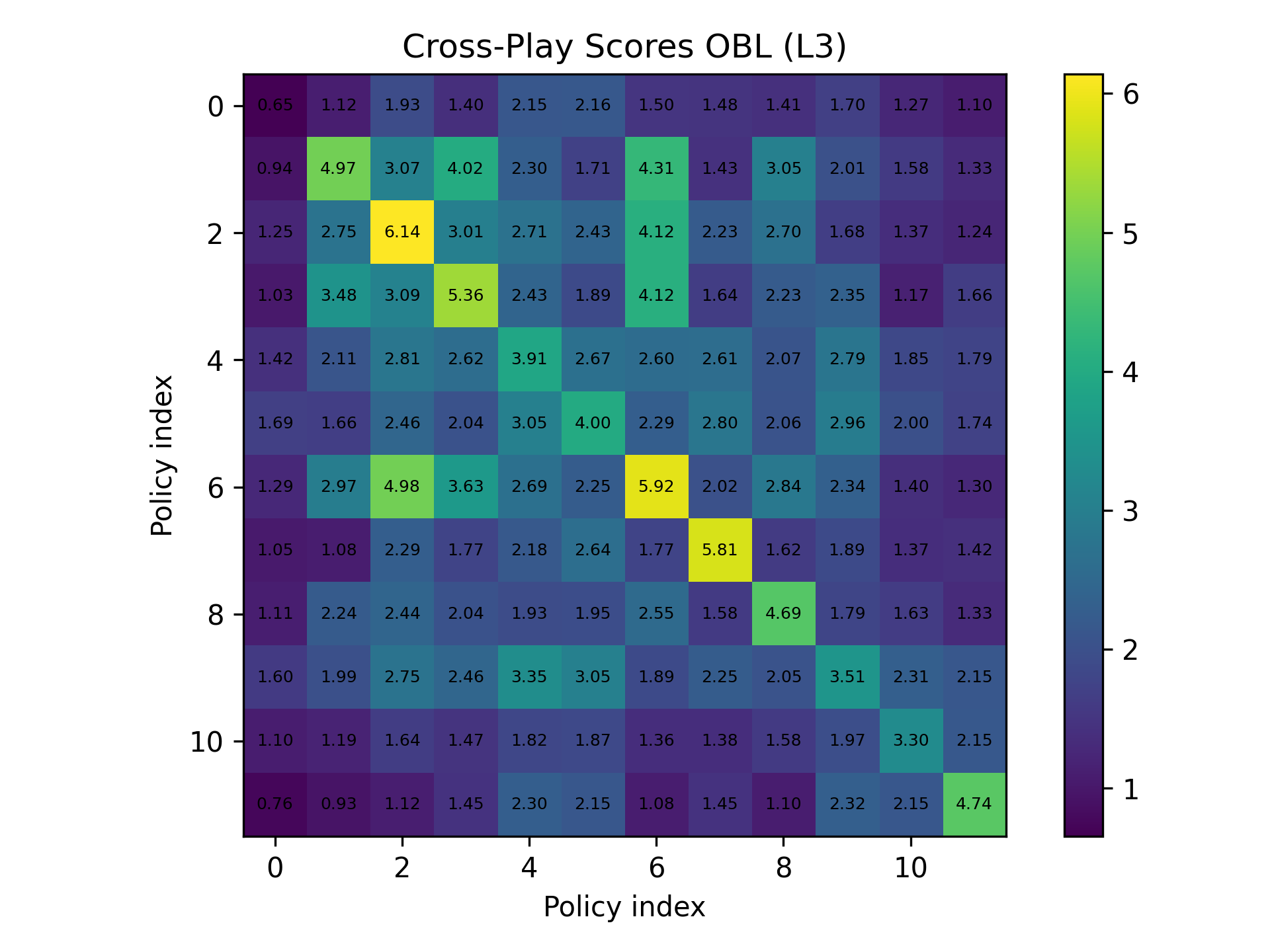}
    \caption{Cross-play scores for 2 player OBL level 3 as featured in \autoref{tab:performance}.}
    \label{fig:obl-l3-2p-cross-play}
\end{figure}
\begin{figure}[!h]
    \centering
    \includegraphics[width=\linewidth]{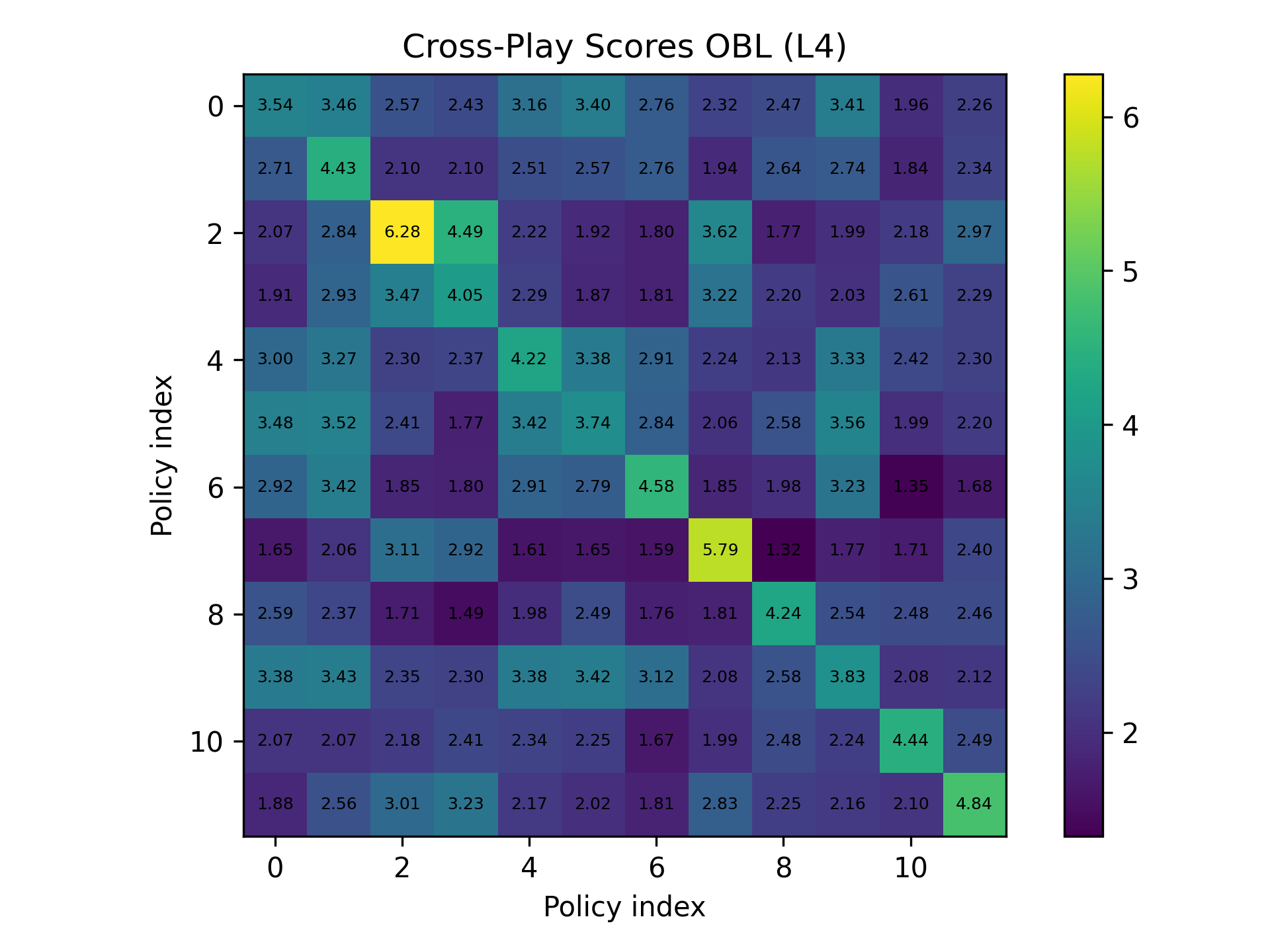}
    \caption{Cross-play scores for 2 player OBL level 4 as featured in \autoref{tab:performance}.}
    \label{fig:obl-l4-2p-cross-play}
\end{figure}
\begin{figure}[!h]
    \centering
    \includegraphics[width=\linewidth]{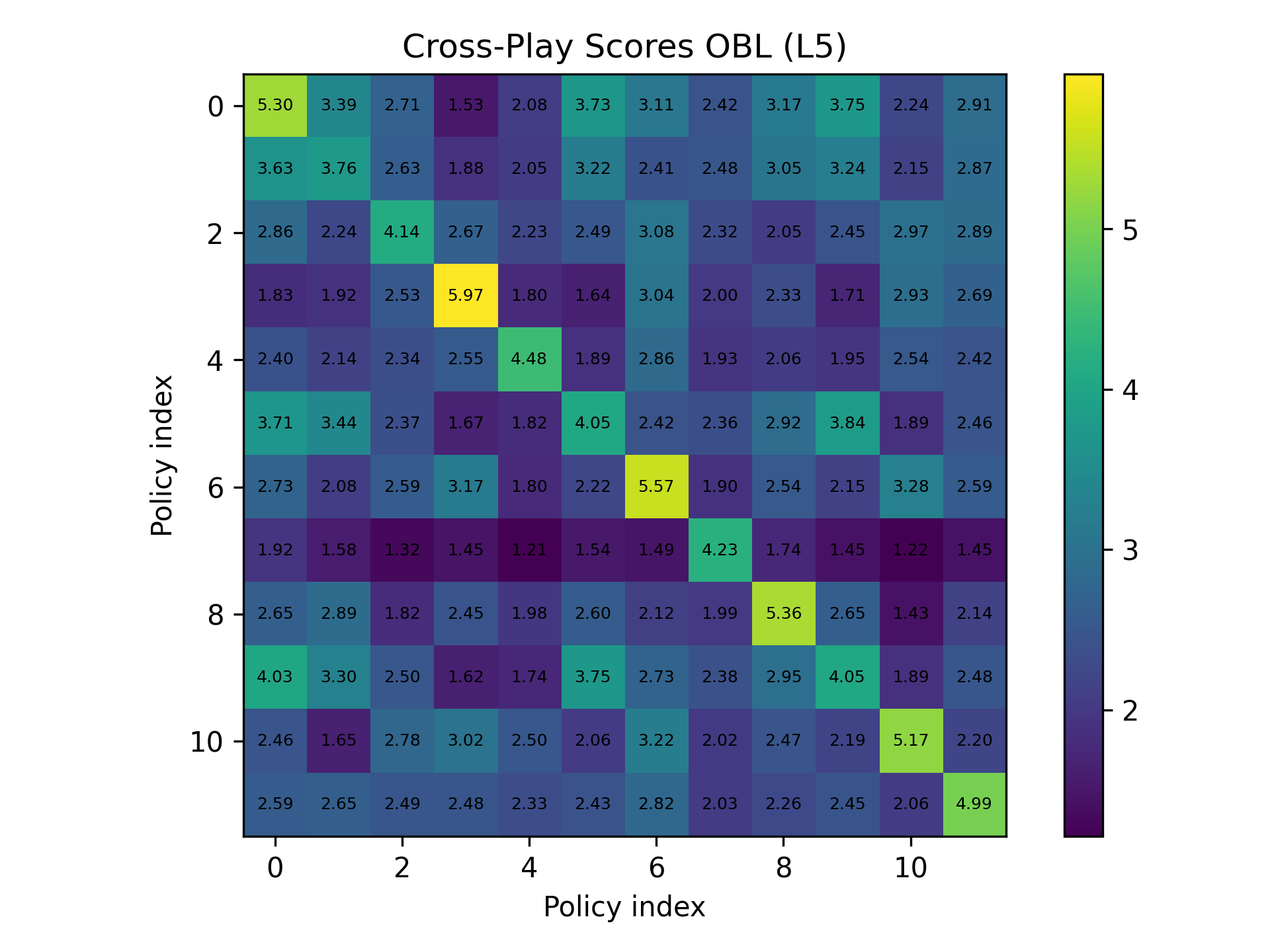}
    \caption{Cross-play scores for 2 player OBL level 5 as featured in \autoref{tab:performance}.}
    \label{fig:obl-l5-2p-cross-play}
\end{figure}
We show cross-play matrices for self-play (\autoref{fig:sp-2p-cross-play}, HE (\autoref{fig:HE-2p-cross-play}), Other-Play (for $\Phi$ in \autoref{fig:op-2p-cross-play}, for $\Phi_{\text{rot}}$ in \autoref{fig:op-2p-cross-play-rotation} and for $\Phi_{\text{colour}}$ in \autoref{fig:op-2p-cross-play-color}) and OBL (level 1 \autoref{fig:obl-l1-2p-cross-play}, level 2 \autoref{fig:obl-l2-2p-cross-play}, level 3 \autoref{fig:obl-l3-2p-cross-play}, level 4 \autoref{fig:obl-l4-2p-cross-play} and level 5 \autoref{fig:obl-l5-2p-cross-play}).

\newpage
\section{Training Curves}
\label{sec:training-curves}
\begin{figure}[!h]
    \centering
    \includegraphics[width=.5\linewidth]{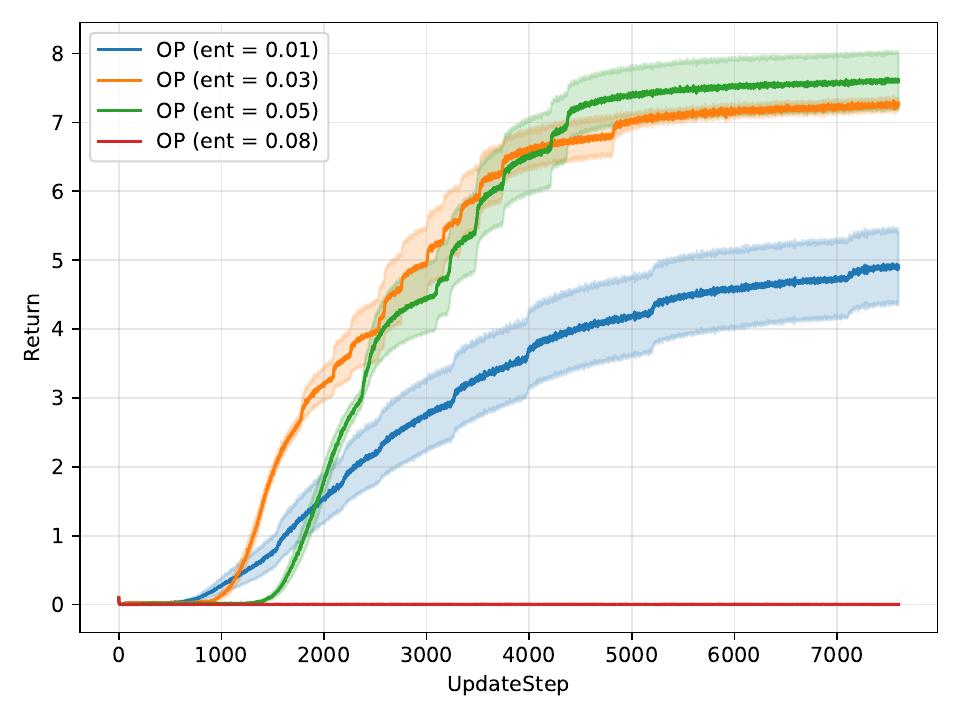}
    \caption{SP training return for all two player OP results as shown in \autoref{tab:ablating_entropy}.}
    \label{fig:op-training-curve-2p-entropy}
\end{figure}
\begin{figure}
    \centering
    \includegraphics[width=0.5\linewidth]{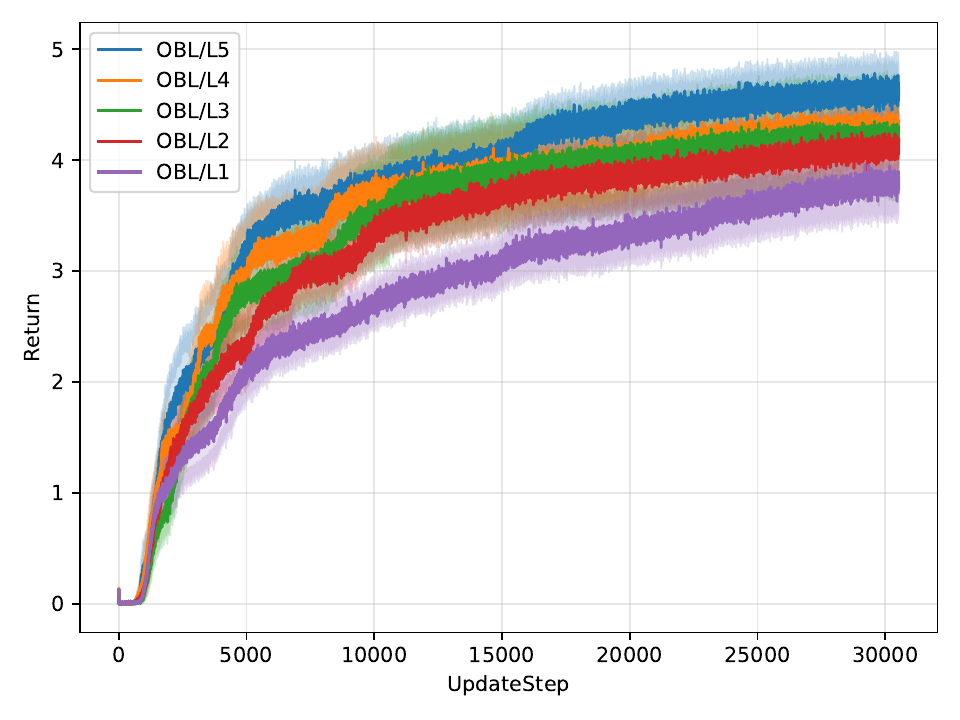}
    \caption{SP training curves for for two player OBL as shown in \autoref{tab:performance}.}
    \label{fig:obl-training-curve-2p}
\end{figure}
\begin{figure}
    \centering
    \includegraphics[width=0.5\linewidth]{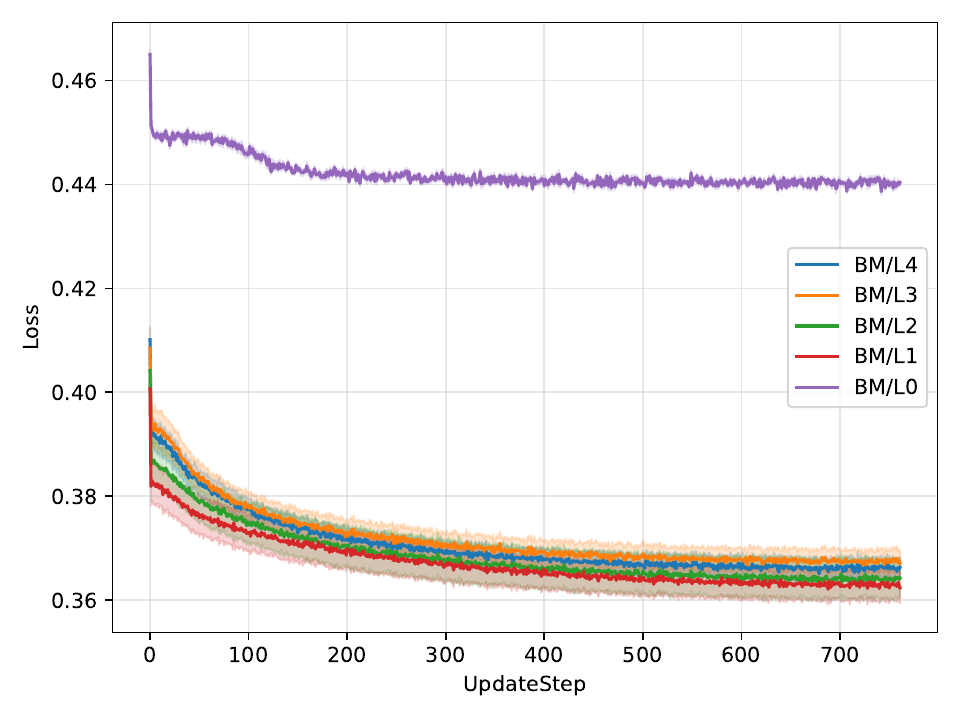}
    \caption{Training curves for all belief model levels for two player OBL as shown in \autoref{tab:performance}.}
    \label{fig:obl-training-curve-2p-bm}
\end{figure}
We show relevant training curves in \autoref{fig:op-training-curve-2p-entropy}, \autoref{fig:obl-training-curve-2p} and \autoref{fig:obl-training-curve-2p-bm}.
All curves show convergence.
As discussed in the main text, OP with $\alpha=0.08$ never learns any meaningful cooperation abilities and therefore shows a flat learning curve.

More insightful are the OBL learning curves.
We can see that the belief model improves hugely from level 0 to level 1 which is expected given that the belief model now is trained on competent self-play behaviour.
After this however, we surprisingly find that future belief models all level out at a final cross-entropy loss of about $0.36$.
Consequently, higher levels of OBL policies also start to saturate at similar final return levels in self-play.
Whether this behaviour is related to the limited cross-play improvements observed in the main paper remains an interesting open question that we leave to future work.

\end{document}